\RequirePackage{fix-cm}
\documentclass[twocolumn]{svjour3}          % twocolumn

\smartqed           % twocolumn
\journalname{International Journal of Computer Vision}

\usepackage{graphicx}
\usepackage{mathptmx}      % use Times fonts if available on your TeX system
\usepackage{latexsym}
\usepackage{times}
\usepackage{amsmath}
\usepackage{epsfig}
\usepackage{graphicx}
\usepackage{amssymb}
\usepackage{indentfirst}

\usepackage{dsfont}

\usepackage{enumerate}
\usepackage{enumitem}
\usepackage{xspace}
\usepackage{xcolor}

\usepackage[toc,page]{appendix}
\usepackage[font=small,labelfont=bf]{caption}

\usepackage{xcolor}

\usepackage{caption}
\usepackage{subcaption}
\usepackage{bm}
\usepackage{isomath}
\usepackage{booktabs} 
\usepackage[numbers]{natbib}
\usepackage{footmisc}

\usepackage[normalem]{ulem}
\usepackage{multirow}

\usepackage{colortbl}

\usepackage{array}

\usepackage{colortbl}
\usepackage{tabularx}
\usepackage{arydshln}
\usepackage{xspace}
\usepackage[pagebackref=true,breaklinks=true,colorlinks,bookmarks=false]{hyperref}

\renewcommand\vec[1]{\ensuremath\boldsymbol{#1}}
\renewcommand\cdots{...}

\newcommand{\vb}{\mathbf{b}}
\newcommand{\vy}{\mathbf{y}}

\newcommand{\tX}{\vec{\mathcal{X}}}

\newcommand{\vx}{\mathbf{x}}

\newcommand{\mbr}[1]{\mathbb{R}^{#1}}

\newcommand{\mvv}{\mathbf{v}}

\newcommand{\idx}[1]{\mathcal{I}_{#1}}

\newcommand{\vu}{\mathbf{u}}

\newcommand{\vzeta}{\boldsymbol{\zeta}}

\newcommand{\vphi}{\boldsymbol{\phi}}
\newcommand{\vpsi}{\boldsymbol{\psi}}

\newcommand{\vupsilon}{\boldsymbol{\upsilon}}
\newcommand{\vdelta}{\boldsymbol{\delta}}
\newcommand{\mupsilon}{\boldsymbol{\Upsilon}}

\newcommand{\iu}{{i\mkern1mu}}

\newcommand{\enorm}[1]{\left\|{#1}\right\|_2}

\DeclareMathOperator*{\argmin}{arg\,min}
\DeclareMathOperator*{\argmax}{arg\,max}

\DeclareMathOperator*{\kronstack}{\uparrow\!\otimes}

\DeclareMathOperator*{\avg}{avg}
\DeclareMathOperator*{\sgn}{Sgn}

\newcommand{\expl}[1]{\text{e}^{#1}}

\newcommand{\mI}{\mathbf{I}}

\newcommand{\mLambda}{\bm{\lambda}}
\newcommand{\mU}{\bm{U}}
\newcommand{\mV}{\bm{V}}

\newcommand{\mTheta}{\bm{\theta}}
\newcommand{\mLLa}{\bm{\Lambda}}

\newcommand{\vsss}{\boldsymbol{s}}
\newcommand{\vh}{\boldsymbol{h}}

\def\eg{\emph{e.g.}}

\newcommand{\mIdent}{\boldsymbol{\mathds{I}}}

\newcommand{\cov}{\boldsymbol{\Sigma}}

\newcommand{\mM}{\boldsymbol{M}}

\newcommand{\mW}{\boldsymbol{W}}

\newcommand{\vkappa}{\boldsymbol{\kappa}}
\newcommand{\vmu}{\boldsymbol{\mu}}

\newcommand{\mP}{\boldsymbol{\Theta}}
\newcommand{\mPP}{\boldsymbol{P}}

\newcommand{\stkout}[1]{{\ifmmode\text{\sout{\ensuremath{#1}}}\else\sout{#1}\fi}}

\DeclareMathOperator*{\arcsinh}{arcsinh}

\newcommand{\comment}[1]{}

\makeatletter
\@namedef{ver@everyshi.sty}{}
\makeatother

\def\lei#1{{\color{black}{#1}}}
\definecolor{pink}{HTML}{db5a6b}
\definecolor{lblue}{HTML}{2e4e7e}
\definecolor{tiffany}{HTML}{1bd1a5}

\makeatletter
\DeclareRobustCommand\onedot{\futurelet\@let@token\bmv@onedotaux}
\def\bmv@onedotaux{\ifx\@let@token.\else.\null\fi\xspace}
\def\eg{\emph{e.g}.} 
\def\ie{\emph{i.e}.} 
 
\def\etc{\emph{etc}.} \def\vs{\emph{vs}.}
\def\wrt{w.r.t.}

\makeatother

\begin{document}

\title{%Auxiliary Feature Hallucination for Action Recognition
Feature Hallucination for Self-supervised Action Recognition
}


\author{Lei Wang         \and
        Piotr Koniusz
}

\institute{$\cdot\;\!$L. Wang (corresponding author) is a Research Fellow (Grade 2) in the School of Engineering and Built Environment, Electrical and Electronic Engineering, at Griffith University, and a Visiting Scientist at Data61/CSIRO. 
           \email{l.wang4@griffith.edu.au}.\\
           \and
          P. Koniusz is a Principal Research Scientist at Data61/CSIRO, and an Honorary Associate Professor (Level D) at ANU\lei{, and an Adjunct Associate Professor (Level D) at the University of New South Wales (UNSW)}. % \email{piotr.koniusz@data61.csiro.au}.
}

\date{Received: 16.06.2024 / Revised: 24.11.2024 \lei{/ Revised: 05.03.2025} / Accepted: 25.06.2025
}

\maketitle

\begin{sloppypar}
\begin{abstract}

\lei{Understanding human actions in videos requires more than raw pixel analysis; it relies on high-level semantic reasoning and effective integration of multimodal features. 
We propose a deep translational action recognition framework that enhances recognition accuracy by jointly predicting action concepts and auxiliary features from RGB video frames. At test time, hallucination streams infer missing cues, enriching feature representations without increasing computational overhead.
To focus on action-relevant regions beyond raw pixels, we introduce two novel domain-specific descriptors. \textit{Object Detection Features} (ODF) aggregate outputs from multiple object detectors to capture contextual cues, while \textit{Saliency Detection Features} (SDF) highlight spatial and intensity patterns crucial for action recognition. Our framework seamlessly integrates these descriptors with auxiliary modalities such as optical flow, Improved Dense Trajectories, skeleton data, and audio cues. It remains compatible with state-of-the-art architectures, including I3D, AssembleNet, Video Transformer Network, FASTER, and recent models like VideoMAE V2 and InternVideo2.
To handle uncertainty in auxiliary features, we incorporate aleatoric uncertainty modeling in the hallucination step and introduce a robust loss function to mitigate feature noise. Our multimodal self-supervised action recognition framework achieves state-of-the-art performance on multiple benchmarks, including Kinetics-400, Kinetics-600, and Something-Something V2, demonstrating its effectiveness in capturing fine-grained action dynamics.}

\end{abstract}
\end{sloppypar}

%%%%%%%%% BODY TEXT

\section{Introduction}
\label{sec:intro}

\begin{figure}[tbp]%htbp % left bottom right top
\centering
\begin{subfigure}[b]{0.495\linewidth}
\centering\includegraphics[trim=0 0 0 0, clip=true,width=0.95\linewidth]{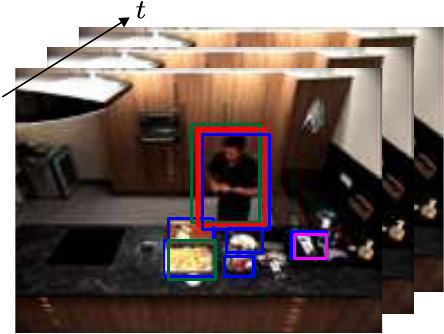}
\caption{\label{fig:det}}
\end{subfigure}
\begin{subfigure}[b]{0.495\linewidth}
\begin{subfigure}[b]{0.995\linewidth}
\centering\includegraphics[trim=0 0 0 0, clip=true,width=0.90\linewidth]{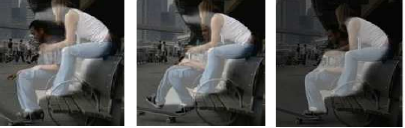}
\caption{\label{fig:sal-reg}}
\end{subfigure}
\begin{subfigure}[b]{0.995\linewidth}
\centering\includegraphics[trim=0 0 0 0, clip=true,width=0.90\linewidth]{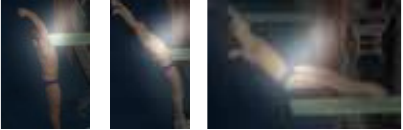}
\caption{\label{fig:sal-temp}}
\end{subfigure}
\end{subfigure}
\caption{\lei{We use object detectors and saliency maps to identify action-relevant regions within video frames.} Fig. \ref{fig:det} shows bounding boxes from four detectors. The faster R-CNN detector with ResNet101 is focused on human-centric actions, such as {\em stand}, {\em watch}, {\em talk}, \etc. The other three detectors identify objects, such as {\em oven}, {\em sink}, {\em clock}, \etc. Fig. \ref{fig:sal-reg} shows how the MNL saliency detector~\cite{Zhang_2018_CVPR} emphasizes spatial regions. Fig. \ref{fig:sal-temp} shows how the ACLNet saliency detector~\cite{Zhang_2019_CVPR} highlights motion regions.}
\label{fig:det-sal}
\end{figure}
\begin{sloppypar}

\begin{figure*}[tbp]
\centering\includegraphics[trim=0 0 0 0, clip=true,width=\linewidth]{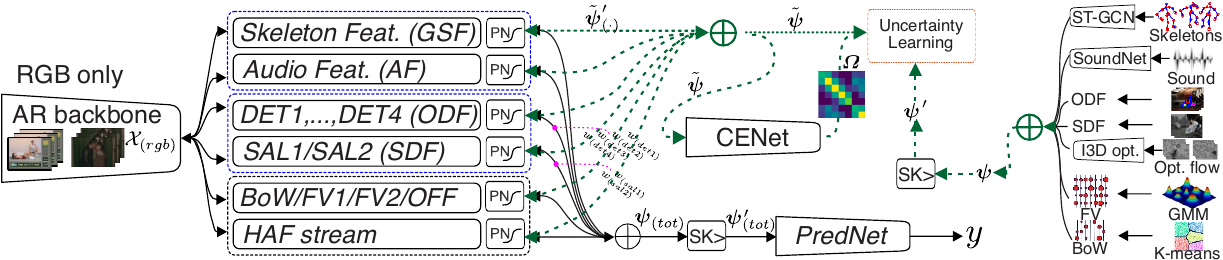}
\caption{We use six mainstream action recognition (AR) backbones: I3D, AssembleNet/AssembleNet++, Video Transformer Network (VTN), the lightweight FASTER framework, and recent models like VideoMAE V2 and InternVideo2. 
The prediction layer (\eg, the final 1D covolutional layer of I3D) is removed from each backbone, and the intermediate representation (\eg, the pooled spatiotemporal token embeddings of VideoMAE V2 / InternVideo2 encoder), $\mathcal{X}_{(rgb)}$, is passed into the following streams: {\em Bag-of-Words} ({\em BoW}), {\em Fisher Vector} ({\em FV}), {\em Optical Flow Features} ({\em OFF}), and {\em High Abstraction Features} ({\em HAF}), followed by the {\em Power Normalization} ({\em PN}) block (dashed black). The OFF stream is supervised by features extracted from the pre-trained I3D optical flow network (on Kinetics-400). Additionally, we introduce new feature streams: {\em Object Detection Features} ({\em ODF}) (from detector-based descriptors {\em DET1}$,\dots,${\em DET4}), {\em Saliency Detection Features} ({\em SDF}) (from saliency-based descriptors {\em SAL1} and {\em SAL2}), {\em spatio-temporal GCN-encoded Skeleton Features} ({\em GSF}), and {\em Audio Features} ({\em AF}) (dashed blue). 
The {\em GSF} stream is supervised by skeleton features from the pre-trained ST-GCN (on Kinetics-Skeletons), and the {\em AF} stream is supervised by audio features from SoundNet (pre-trained on 2-million unlabeled videos).
The resulting feature vectors, $\vec{\tilde{\psi}}_{(\cdot)}$, where ($(\cdot)$ denotes the stream name, are aggregated ($\oplus$), followed by {\em Sketching} ({\em SK}) block, and passed into the {\em Prediction Network} ({\em PredNet}). The {\em ODF} and {\em SDF} features are reweighted by corresponding weights $w_{(\cdot)}$ (magenta lines)). 
Green dashed arrows show the feature hallucination process.
During training, we use either MSE loss or our uncertainty learning loss (dashed red) for hallucination streams. 
The {\em Covariance Estimation Network} ({\em CENet}) takes the concatenated hallucinated features and produces a precision matrix $\boldsymbol{\Omega}$ (the inverse of the covariance matrix). 
This matrix, along with the hallucinated features $\vec{\tilde{\psi}}$ and ground truth features $\vpsi'$, is fed into the uncertainty learning module.
During testing, the hallucinated features $\vec{\tilde{\psi}}$ are input into PredNet to obtain the predicted labels $y$. 
}
\label{fig:pipe}
\end{figure*}

\lei{Action recognition (AR) has evolved significantly, transitioning from handcrafted feature-based methods \cite{hof,sift_3d,3D-HOG,dense_traj,dense_mot_boundary,improved_traj,sivic_vq,csurka04_bovw,perronnin_fisher,perronnin_fisherimpr,lei_thesis_2017,lei_tip_2019,lei_icip_2019} to deep learning-driven approaches \cite{two_stream,spattemp_filters,spat_temp_resnet,i3d_net,wang2023robust,chen2024motion,zhuadvancing,wangtaylor}. Early techniques such as Histogram of Gradients (HOG) \cite{hog2d,3D-HOG}, Histogram of Optical Flow (HOF) \cite{hof}, and Improved Dense Trajectories (IDT) \cite{dense_mot_boundary,improved_traj} effectively captured spatial and temporal patterns but suffered from high computational costs and poor scalability. The emergence of deep learning models, particularly two-stream networks \cite{two_stream} and Inflated 3D CNNs (I3D) \cite{i3d_net}, introduced significant improvements by learning end-to-end representations from raw RGB and optical flow. However, despite these advancements, action recognition remains an open challenge due to the absence of well-established models that can effectively address three fundamental issues: (i) incomplete and imbalanced multimodal data, (ii) inefficient feature fusion across modalities, and (iii) the lack of structured motion descriptors in deep learning models.}

\lei{A key limitation of existing robust AR models is their dependence on multimodal data, such as RGB, optical flow, and skeletons. While multimodal learning can enhance action recognition performance by using complementary cues \cite{ding2025learnable}, most benchmark datasets only provide RGB videos, leading to missing or imbalanced modalities. Many datasets do not provide all possible modalities. For example, some datasets only contain RGB videos, while others may include skeleton or depth data but lack optical flow. This inconsistency forces models to either rely exclusively on RGB-based representations, leading to suboptimal motion reasoning, or incorporate handcrafted features like IDT, which, despite their effectiveness, are computationally prohibitive and incompatible with modern deep learning frameworks \cite{basura_rankpool2,hok,anoop_rankpool_nonlin,anoop_advers,potion}. These constraints prevent existing models from fully exploiting multimodal cues in a scalable and efficient manner \cite{basura_rankpool2,hok,anoop_rankpool_nonlin,anoop_advers,potion,wang2024high,wang2023robust}.}

\lei{Beyond the challenge of missing modalities, current feature fusion strategies suffer from fundamental inefficiencies \cite{ding2025learnable}. Late fusion techniques process each modality separately before combining predictions, failing to model fine-grained cross-modal interactions. Early fusion, on the other hand, directly combines raw features, often introducing modality misalignment and redundant information. Existing hallucination-based approaches \cite{Wang_2019_ICCV,tencent_hall,lei_mm21} attempt to synthesize missing modalities (\eg, estimating optical flow), but they remain constrained to a fixed set of features and fail to generalize across datasets with diverse action categories (\eg, DEEP-HAL \cite{Wang_2019_ICCV}). Even when multiple modalities are present, they may not be equally available or useful. Some modalities might be noisy, sparse, or unreliable. Consequently, there is no well-established approach that efficiently integrates multimodal information while handling missing data in a robust and scalable manner.}

\lei{Moreover, modern deep learning models \cite{wang2024internvideo2} primarily focus on RGB and text-based semantics while neglecting motion-specific domain knowledge \cite{Wang_2019_ICCV,lei_mm21,chen2024motion,dingjourney,10.1145/3701716.3717744}. Unlike handcrafted methods that explicitly model motion dynamics, deep networks, especially 3D CNNs \cite{i3d_net,assemblenet,assemblenet_plus}, Vision Transformers \cite{10.1145/3341105.3373906,piergiovanni2023rethinking,yao2023side4video,10655590} and Masked Autoencoder (MAE) \cite{tong2022videomae,wang2023videomae}, learn representations implicitly, often discarding structured motion cues that are critical for distinguishing similar actions, such as trajectories or motion boundaries \cite{improved_traj}.
While handcrafted descriptors like IDT effectively capture motion, they rely on fixed heuristics and are computationally expensive. Deep learning models, in contrast, lack explicit mechanisms to track movement over time, making them susceptible to failures in fast or subtle motion scenarios \cite{chen2024motion}. Additionally, they require massive amounts of data to learn motion cues \cite{dingjourney,10.1145/3701716.3717744}, which is impractical for many real-world applications. This fundamental gap in motion reasoning limits model performance, particularly in challenging scenarios where appearance-based features alone are insufficient. The trade-off between efficiency and accuracy remains unresolved, as highly expressive models demand extensive feature engineering and computationally expensive training, hindering large-scale deployment\cite{9607406, Arnab_2021_ICCV, bulat2021spacetime, liu2021video,dosovitskiy2021an,yan2022multiview,assemblenet,assemblenet_plus,patrick2021keeping}.}

\lei{To address these limitations, we propose a self-supervised multimodal framework that enhances feature integration, reduces reliance on handcrafted descriptors, and enables robust action recognition even in incomplete multimodal settings. Our approach introduces two novel domain-specific descriptors. First, \textit{Object Detection Features} (ODF) capture action-relevant entities and contextual information using object detection outputs, such as those from Faster R-CNN \cite{faster-rcnn}, improving spatial awareness. Figure \ref{fig:det} shows examples of bounding boxes detected by four object detectors. Second, \textit{Saliency Detection Features} (SDF) extract salient motion regions, helping the model focus on task-relevant patterns and improving action recognition accuracy. Figures \ref{fig:sal-reg} and \ref{fig:sal-temp} show saliency maps from region-wise and temporal saliency detectors. These descriptors serve as semantic priors, guiding our model to attend to informative regions in video frames. Unlike existing methods that rely solely on RGB and optical flow, our approach dynamically integrates motion and structural cues into deep learning pipelines. More importantly, ODF and SDF introduce structured, learnable motion-aware features that enhance deep learning models with explicit spatial and motion cues while maintaining computational efficiency.}

\lei{Beyond feature enhancement, our framework incorporates a self-supervised hallucination mechanism, allowing the model to synthesize missing modalities (\eg, skeleton data \cite{stgcn2018aaai} and audio cues \cite{soundnet}, \etc) at test time. This enables robust performance even when certain modalities are absent, making the model scalable and practical. Additionally, we introduce aleatoric uncertainty modeling, which mitigates feature noise and ensures stable predictions when auxiliary data is unreliable.
By addressing the three fundamental challenges of multimodal action recognition, our framework offers a generalizable, efficient, and scalable solution. Figure \ref{fig:pipe} provides a conceptual overview of our approach. Our method achieves state-of-the-art results on benchmark datasets such as Kinetics-400, Kinetics-600, and Something-Something V2, demonstrating its ability to bridge the gap between handcrafted and deep learning-based approaches. 
Our contributions can be summarized as follows:
\renewcommand{\labelenumi}{\roman{enumi}.}
\begin{enumerate}[leftmargin=0.6cm]
    \item We introduce \textbf{a novel multimodal action recognition framework} that integrates diverse auxiliary features while reducing the reliance on computationally expensive handcrafted descriptors during inference.
    \item We propose Object Detection Features (ODF) and Saliency Detection Features (SDF) as \textbf{domain-specific descriptors} that guide the model toward action-relevant regions, improving motion reasoning and action recognition accuracy.
    \item We develop \textbf{a self-supervised hallucination mechanism} to synthesize missing cues at test time, addressing the challenge of incomplete multimodal data.
    \item We incorporate \textbf{aleatoric uncertainty modeling} and a robust loss function to mitigate feature noise, enhancing performance on fine-grained action recognition tasks.
\end{enumerate}}

\end{sloppypar}

\section{Related Work}
\label{sec:related}

\begin{sloppypar}

% Below, we first outline early video descriptors and their encoding strategies, followed by an overview of deep learning pipelines for Action Recognition (AR), covering optical flow, popular Graph Convolutional Networks (GCNs), Neural Architecture Search (NAS), recent transformers, and masked autoencoders. Next, we examine object and human detectors, spatial and temporal saliency detectors, and audio features used in AR. Finally, we discuss the role of uncertainty in computer vision and the use of Power Normalization (PN) to mitigate the burstiness of feature descriptors.

\lei{We review early video descriptors, deep learning pipelines, object and human detectors, saliency and audio features, as well as uncertainty in vision and Power Normalization (PN) for mitigating feature burstiness in Action Recognition (AR).}

\subsection{Early Video Descriptors and Encoding Schemes}

\noindent{\bf{Early video descriptors.}} Early approaches relied on spatio-temporal interest point detectors \cite{harris3d,cuboid,sstip,hes-stip,mv-stip,dense_traj} and spatio-temporal descriptors \cite{hof,sift_3d,hof2, dense_traj,dense_mot_boundary,improved_traj} which capture various appearance and motion statistics. 
However, spatio-temporal interest point detectors struggle to capture long-term motion patterns. To address this, the Dense Trajectory (DT) \cite{dense_traj} approach is developed, which densely samples feature points in each frame and tracks them across the video frames using optical flow. Multiple descriptors are then extracted along these trajectories to capture shape, appearance and motion cues. Despite its utility, DT cannot account for camera motion. The Improved Dense Trajectory (IDT) approach \cite{improved_traj,dense_mot_boundary} overcomes this limitation by estimating and removing global background motion caused by the camera. Additionally, IDT filters out inconsistent matches using a human detector. 
For spatio-temporal descriptors, IDT uses HOG \cite{hog2d}, HOF \cite{hof} and MBH \cite{dense_mot_boundary}. 
HOG \cite{hog2d} contains statistics of the amplitude of image gradients \wrt~the gradient orientation, thus it captures the static appearance cues. 
In contrast, HOF \cite{hof} captures histograms of optical flow while MBH \cite{dense_mot_boundary} captures derivatives of the optical flow, thus it is highly resilient to the global camera motion whose cues cancel out due to derivatives. Thus, HOF and MBH contain the zero- and first-order optical flow statistics. Other notable spatio-temporal descriptors include HOG-3D \cite{3D-HOG}, SIFT3D \cite{sift_3d}, SURF3D \cite{hes-stip} and LTP \cite{LTP}.

% \vspace{0.05cm}
\noindent{\bf{BoW/FV encoding.}} The Bag-of-Words (BoW) method \cite{sivic_vq,csurka04_bovw} 
creates a visual vocabulary using k-means clustering, where local descriptors are assigned to specific clusters. Variants include Soft Assignment (SA) \cite{soft_ass,me_SAO} and Localized Soft Assignment (LcSA) \cite{liu_sadefense,me_ATN}. 
Following DEEP-HAL \cite{Wang_2019_ICCV}, we use BoW encoding \cite{csurka04_bovw} with Power Normalization \cite{me_ATN}. Additionally, we use Fisher Vectors (FV) \cite{perronnin_fisher,perronnin_fisherimpr}, which capture first- and second-order statistics of local descriptors assigned to Gaussian Mixture Model (GMM) clusters.

\subsection{Deep Learning in Action Recognition}

\noindent{\bf{\lei{CNN-based}.}} 
Early AR models using CNNs relied on frame-wise features with average pooling 
\cite{cnn_basic_ar}, which discarded the temporal order. To address this, frame-wise CNN scores are input to LSTMs \cite{cnn_lstm_ar}, while two-stream networks \cite{two_stream} compute separate representations for RGB frames and 10 stacked optical flow frames. Spatio-temporal patterns are later modeled using 3D CNN filters \cite{cnn3d_ar,spattemp_filters,spat_temp_resnet,long_term_ar}.

While two-stream networks \cite{two_stream} overlook temporal order,  approaches like rank pooling \cite{basura_rankpool, basura_rankpool2, anoop_rankpool_nonlin,anoop_advers} and higher-order pooling \cite{hok,me_tensor_eccv16,me_tensor,Pengfei_ICCV19,kon_tpami2020b,wang2024high} gained popularity.
The I3D model \cite{i3d_net} introduces spatio-temporal `inflation', where 2D CNN filters pre-trained on ImageNet-1K~\cite{5206848} are adapted to 3D, incorporating temporal pooling. 
PAN~\cite{acmmm19_ZhangZCG19} proposes the Persistence of Appearance motion cue, which distills motion information directly from adjacent RGB frames.  
A bootstrapping approach \cite{acmmm19_LiuGQWL19} uses long-range temporal context attention, while another \cite{acmmm20_KumarKSXS20} introduces a graph attention model to explore semantic relationships.
Slow-I-Fast-P (SIFP) \cite{acmmm20_2020A} uses dual pathways for compressed AR, with sparse sampling on I-frames and dense sampling using pseudo optical flow clips.

\noindent{\bf{\lei{Optical flow}.}} Optical flow remains a cornerstone in AR \cite{two_stream,i3d_net,basura_rankpool,anoop_advers,wang2024flow}. 
Early methods tackle small displacements \cite{flow_def2,brox_accurate}, while new methods address larger displacements, such as Large Displacement Optical Flow (LDOF) \cite{brox_largedisp}. Recent methods involve non-rigid descriptor or segment matching \cite{deep_flow,seg_flow} and edge-preserving interpolation \cite{epic_flow}. 
We use LDOF \cite{brox_accurate} for optical flow estimation.

% \vspace{0.05cm}
\noindent{\bf{\lei{GCNs for skeletons}.}} \lei{Graph Convolutional Networks (GCNs) have shown great success in skeletal AR~\cite{Shotton2011,Cao_2017_CVPR,8953648,Cheng_2020_CVPR, Chen_Li_Yang_Li_Liu_2021,10.1145/3412841.3441974,wang20213d,wang2022temporal,wang2022uncertainty,wang2024meet, wang2023robust}}. These methods construct skeleton graphs, where joints are vertices, and bones are edges, enabling GCNs to model dependencies~\cite{kipf2017semi}. The spatio-temporal GCN (ST-GCN)~\cite{stgcn2018aaai} simultaneously learns spatial and temporal features. Subsequent advancements include Actional-Structural GCN (AS-GCN)~\cite{Li_2019_CVPR},
Context-Aware GCN (CA-GCN)~\cite{9156373}, Shift-GCN~\cite{9157077}, dynamic directed GCN~\cite{10.1007/978-3-030-58565-5_45}, part-level GCN~\cite{Huang_Huang_Ouyang_Wang_2020}, and other specialized GCNs~\cite{UHAR_BMVC2021,2sagcn2019cvpr,9420299,9895208}.
\lei{In this work, we use pre-trained ST-GCN on Kinetics-skeleton to extract skeleton features for feature hallucination}.

\noindent{\bf{\lei{NAS}.}} AssembleNet~\cite{assemblenet} uses Neural Architecture Search (NAS) to identify an optimal architecture for spatio-temporal feature interactions in AR. AssembleNet++~\cite{assemblenet_plus} extends this by dynamically learning attention weights to explore interactions between appearance, motion, and spatial object information. We use AssembleNet++ in this work.

\noindent{\bf{\lei{Transformer-based}.}} The AR field is increasingly adopting transformer-based models~\cite{dosovitskiy2021an}. Pure transformer architectures have achieved state-of-the-art accuracy on major video recognition benchmarks~\cite{Arnab_2021_ICCV, 9607406, bulat2021spacetime,liu2021video,kondratyuk2021movinets, wang20233mformer} by globally connecting spatial and temporal patches. Lightweight transformers, such as the Video Transformer Network (VTN)~\cite{10.1145/3341105.3373906}, enable real-time AR on low-power devices, including edge computing scenarios. Recent advancements include TubeViT~\cite{piergiovanni2023rethinking}, Side4Video~\cite{yao2023side4video}, and OmniVec2~\cite{10655590}, which enhance video understanding via multimodal and multitask learning.

\noindent{\bf{\lei{Lightweight models}.}} \lei{While recent methods~\cite{9607406, Arnab_2021_ICCV, bulat2021spacetime, liu2021video,dosovitskiy2021an,yan2022multiview}, such as AssembleNet~\cite{assemblenet,assemblenet_plus} and Motionformer~\cite{patrick2021keeping}, have achieved state-of-the-art performance in AR, their high computational makes them unsuitable for real-time or resource-constrained applications. To address this, lightweight models like SqueezeNet~\cite{squeezenet}, Xception~\cite{Chollet_2017_CVPR}, ShuffleNet~\cite{Ma_2018_ECCV}, EfficientNet~\cite{pmlr-v97-tan19a}, MobileNet~\cite{9008835}, as well as frameworks like FASTER~\cite{faster} and Video Transformer Networks (VTN)~\cite{10.1145/3341105.3373906}, have been proposed to mitigate these challenges.} Recent efforts have focused on optimizing architectures at the clip level~\cite{9008835} to further reduce computational overhead. Given the strong temporal structure and high redundancy in video data, the FASTER framework~\cite{faster} tackles these challenges by emphasizing the temporal aggregation stage. By using a lightweight model, it effectively captures scene changes over time while minimizing redundant computations.

\noindent{\bf{\lei{Masked vision modeling}.}} Self-supervised video pre-training methods like VideoMAE~\cite{tong2022videomae} and its successor VideoMAE V2~\cite{wang2023videomae} have significantly improved tasks like action classification and spatial-temporal detection. VideoMAE uses high-ratio video tube masking to enhance representation learning. VideoMAE V2 introduces dual masking to reduce decoder input length, improving both computational efficiency and learning performance. InternVideo2~\cite{wang2024internvideo2} further advances AR through multi-stage training on web and YouTube datasets, achieving state-of-the-art performance across 60+ video and audio tasks.

Building on \cite{Wang_2019_ICCV,lei_mm21}, we investigate the use of various backbones, including AssembleNet++, lightweight VTN, FASTER, and recent advancements such as VideoMAE V2 and InternVideo2. These backbones are used to hallucinate computationally expensive handcrafted features, such as optical flow and skeleton features, effectively reducing the need for explicit feature extraction during testing stage.

\subsection{Object and Saliency Detectors}

\vspace{0.05cm}
\noindent{\bf{Object detectors.}} 
Modern deep learning-based object detection methods include Region-based Convolutional Neural Networks (R-CNN) \cite{rcnn}, its faster variants \cite{fast_rcnn,faster-rcnn}, mask-based extensions \cite{mask_rnn}, and the YOLO family \cite{yolo}, including YOLO v2 and YOLO v3, which prioritize efficiency by using a single network architecture.

In this work, we use the faster R-CNN detector \cite{faster-rcnn} with several backbones: (i) Inception V2~\cite{Szegedy_2016_CVPR}, (ii) Inception ResNet V2~\cite{Szegedy_2017_AAAI}, (iii) ResNet101~\cite{He_2016_CVPR} and (iv) NASNet~\cite{Zoph_2018_CVPR}. Among these, Inception V2, Inception ResNet V2, and NASNet are pre-trained on the COCO dataset~\cite{Lin_eccv2014_coco}, enabling detection across 91 object classes. These models are particularly adept at summarizing environments, such as indoor settings, and associating scene context with actions. ResNet101, on the other hand, is pre-trained on the AVA v2.1 dataset~\cite{Gu_2018_CVPR}, which includes 80 human actions, making it  highly suited for human-centric AR tasks.
In addition to detection scores, each bounding box is further described using ImageNet-1K \cite{ILSVRC15} scores from a pre-trained Inception ResNet V2 model~\cite{Szegedy_2017_AAAI}.

\vspace{0.05cm}
\noindent{\bf{Saliency detectors.}} 
Saliency detectors identify image regions that correlate with human visual attention, typically represented as saliency maps. Deep learning-based saliency models \cite{RFCN,ChengCVPR17} outperform traditional methods \cite{Background-Detection:CVPR-2014} but often require  pixel-wise annotations. \lei{Recent advancements include MNL \cite{Zhang_2018_CVPR} (a weakly-supervised model), RFCN \cite{RFCN} (a fully-supervised model), and a Robust Background Detector (RBD) \cite{Background-Detection:CVPR-2014} (refer to \cite{SalObjBenchmark_Tip2015} for a detailed survey).}

For spatial saliency, we use MNL~\cite{Zhang_2018_CVPR}, which is trained on noisy labels derived from weak or unsupervised handcrafted saliency models.
For temporal saliency, we rely on ACLNet~\cite{Zhang_2019_CVPR}, a CNN-LSTM-based architecture designed to capture dynamic saliency patterns over time.

\subsection{Audio Modality in Action Recognition}
The visual and audio modalities are highly correlated yet contain distinct and complementary information. Numerous studies~\cite{4627014, 5692578, 6343802, 10.1145/3389189.3389196,Gao_2020_CVPR} have explored the integration of audio and visual cues for AR, as audio can provide strong complementary evidence for certain actions. This strong correlation enables accurate semantic predictions of one modality from the other. At the same time, their intrinsic differences make cross-modal prediction a valuable pretext task for self-supervised learning, offering advantages over within-modality learning. Building on this idea, Cross-Modal Deep Clustering (XDC)~\cite{alwassel_2020_xdc} uses both the semantic correlation and distinct characteristics of visual and audio modalities to enhance AR. Similarly, a fused multisensory representation~\cite{multisensory2018} has been introduced to jointly model visual and audio components, leading to a richer and more robust understanding of video content.

\lei{SoundNet~\cite{soundnet} transfers discriminative knowledge from visual recognition models to the sound modality. Using two million unlabeled videos, it bridges the gap between vision and audio, making it ideal for extracting audio features as ground truth in our hallucination process.
More recently, OmniVec2~\cite{10655590}, a multimodal multitask transformer-based model, has been introduced. It employs modality-specialized tokenizers, a shared transformer architecture, and cross-attention mechanisms to project different modalities, including audio, into a unified embedding space. Additionally, InternVideo2~\cite{wang2024internvideo2}, a new family of video foundation models, incorporates video-audio correspondence in its second-stage training, encouraging deeper semantic learning across modalities.}

\subsection{Uncertainty in Computer Vision}
Uncertainty in computer vision is generally categorized into \textit{aleatoric} and \textit{epistemic} uncertainty~\cite{uncertainty1, uncertainty2, uncertainty3, uncertainty4, uncertainty5}. Aleatoric uncertainty is typically modeled by a Gaussian distribution over the predictions, while epistemic uncertainty is represented by a distribution over the model weights, as seen in Bayesian Neural Networks. In simple terms, aleatoric (or statistical) uncertainty refers to randomness or inherent variability in the data, whereas epistemic (or systematic) uncertainty refers to uncertainty stemming from a lack of knowledge (\eg, uncertainty about the best model, essentially representing ignorance).

In this work, we primarily focus on heteroscedastic aleatoric uncertainty, which has become popular in many applications. Examples include uncertainty-weighted multi-task loss in depth regression and segmentation~\cite{Kendall_2018_CVPR}, bounding box regression with uncertainty in Faster R-CNN~\cite{klloss} and YOLOv3~\cite{Choi_2019_ICCV}, deep learning-assisted methods for measuring uncertainty in action recognition~\cite{8834505}, uncertainty-aware audio-visual action recognition~\cite{Subedar_2019_ICCV}, uncertainty quantification for deep context-aware mobile AR~\cite{aistats-uncert}, structured uncertainty prediction networks for face images~\cite{8578672}, and recent few-shot keypoint detection with uncertainty learning~\cite{DBLP:journals/corr/abs-2112-06183}.
However, many of these approaches treat multiple variables independently, while we model uncertainty with covariance to capture the underlying relationships between variables. Specifically, we model the aleatoric uncertainty of features during the hallucination step to further enhance the performance of action recognition.

\subsection{Power Normalization Family}
BoW, FV and even CNN-based descriptors must address the phenomenon of burstiness, which is defined as `{\em the property that a given visual element appears more times in an image than a statistically independent model would predict}' \cite{jegou_bursts}. This phenomenon is also present in video descriptors. Power Normalization~\cite{me_ATN,me_tensor_tech_rep} is known to mitigate burstiness and has been extensively studied in the context of BoW \cite{me_ATN,me_tensor_tech_rep,me_tensor,me_deeper}. Additionally, a connection to max-pooling was identified in \cite{me_ATN}, which demonstrates that the so-called MaxExp pooling is, in fact, a detector of `\emph{at least one particular visual word being present in an image}'. According to the studies \cite{me_ATN,me_deeper}, many Power Normalization functions are closely related. The Power Normalizations used in our work are outlined in Section \ref{sec:backgr}.
\end{sloppypar}

\section{Background}
\label{sec:backgr}

\lei{Below, we provide the necessary background information for our framework, beginning with an introduction to our notations.}

\vspace{+0.05cm}
\noindent{\textbf{Notations.}} We use boldface uppercase letters to represent matrices, \eg, $\mM, \mPP$; regular uppercase letters with a subscript to represent matrix elements, \eg, $P_{ij}$, which denotes the $(i,j)^{\text{th}}$ element of $\mPP$; boldface lowercase letters for vectors, \eg, $\vx, \vphi, \vpsi$; and regular lowercase letters for scalars. Vectors may be numbered, \eg, $\vx_n$, while regular lowercase letters with a subscript represent an element of a vector, \eg, $\vx_i$ is the $i^{\text{th}}$ element of $\vx$. The operators `$;$' and `$,$' are used to concatenate vectors along the first and second modes, respectively.  For example, $\circledcirc_{i\in\idx{K}}\mvv_i\!=[\mvv_1; \cdots; \mvv_K]$ and $\circledcirc^2_{i\in\idx{K}}\mvv_i\!=[\mvv_1, \cdots, \mvv_K]$ concatenate a group of vectors along the first and second modes, respectively. $\kronstack_r$ denotes the $r$-th Kronecker power.
The operator $\oplus$ denotes aggregation (sum), while $\idx{d}$ represents an index set of integers $\{1,\cdots,d\}$.

\subsection{Descriptor Encoding Schemes}
\label{sec:enc}

\vspace{+0.05cm}
\noindent{\textbf{Bag-of-Words}} \cite{sivic_vq,csurka04_bovw} assigns each local descriptor $\vec{x}$ to the closest visual word from $\vec{M}\!=\!\left[\vec{m}_1,\cdots,\vec{m}_K\right]$, which is built using k-means. To obtain the mid-level feature $\vec{\phi}$, we solve the following optimization problem:
\begin{equation}\label{eq:sp1}
\begin{array}{l}
\vec{\phi}=\argmin\limits_{\vec{\phi'}}\;\bigr \lVert{ \vec{x}-\vec{M}\vec{\phi'} }\bigr \rVert_2^2,\\
s.\;t.\;\;\vec{\phi'}\in\{0,1\}, \lei{\textbf{1}^T\vec{\phi'}\!=\!1}. % \vOnes
\end{array}
\end{equation}

\vspace{0.05cm}
\noindent{\textbf{Fisher Vector Encoding}} \cite{perronnin_fisher,perronnin_fisherimpr} uses a mixture of $K$ Gaussians from a GMM as a dictionary. It encodes descriptors with respect to the Gaussian components $G(w_k, \vec{m}_k, \vec{\sigma}_k)$, which are parametrized by the mixing probability, mean, and diagonal standard deviation. The first- and second-order features $\vec{\phi}_k, \vec{\phi}'_k\in\mbr{D}$ are given by:

\begin{equation}\label{eq:fisher1}
\vec{\phi}_k=(\vec{x}\!-\!\vec{m}_k)/\vec{\sigma}_{k},\;\;
\vec{\phi}'_k=\vec{\phi}_k^2\!-\!1.
\end{equation}
The concatenation of per-cluster features $\vec{\phi}^{*}_k\in\mbr{2D}$ forms the mid-level feature $\vec{\phi}\in\mbr{2KD}$:
\vspace{-0.2cm}
\begin{equation}\label{eq:fisher2}
\!\!\!\!\vec{\phi}=\left[\vec{\phi}_1^{*}; ...; \vec{\phi}_K^{*} \right],\;\;  \vec{\phi}^{*}_k=\frac{p\left(\vec{m}_k|\vec{x}, \theta\right)}{\sqrt{w_k}}
\left[\vec{\phi}_k; \vec{\phi}'_k/\sqrt{2}\right],
\!\!\!
\end{equation}
where $p$ and $\theta$ represent the component membership probabilities and the parameters of the GMM, respectively. For each descriptor $\vec{x}$ with dimensionality $D$ (after PCA), its encoding $\vec{\phi}$ has a dimensionality of $2KD$, as it captures both first- and second-order statistics.

\subsection{Pooling a.k.a. Aggregation}
\label{sec:aggr}

Traditionally, pooling is performed by averaging mid-level feature vectors $\vphi(\vx)$ corresponding to local descriptors $\vx\!\in\!\tX$ from a video sequence $\tX$, expressed as $\vpsi\!=\!\avg_{\vx\in\tX}\vphi(\vx)$, with optional $\ell_2$-normalization. In this paper, we apply this approach to both full sequences $\tX$ and subsequences.
\begin{proposition}
\label{pr:subseq}
For subsequence pooling, let $\tX_{s,t}\!=\!\tX_{0,t}\!\setminus\tX_{0,s-1}$, where $\tX_{s,t}$ denotes the set of descriptors in the sequence $\tX$ from frame $s$ to frame $t$, with $0\!\leq\!s\!\leq\!t\!\leq\tau$, $\tX_{0,-1}\!\equiv\!\emptyset$, and $\tau$ is the length of $\tX$. Let us compute an integral mid-level feature $\vphi'_t\!=\!\vphi'_{t-1}\!+\!\sum_{\vx\in\tX_{t,t}}\vphi(\vx)$, which aggregates mid-level feature vectors from frame $0$ to frame $t$, with $\vphi'_{-1}$ initialized as an all-zeros vector. The pooled subsequence is then given by:
\begin{equation}\label{eq:integr1}
\lei{\vpsi_{s,t}\!=\frac{\vphi'_t\!-\!\vphi'_{s-1}}{{\lVert\vphi'_t\!-\!\vphi'_{s-1}\rVert_2}+\epsilon}
= \frac{\sum_{\vx\in \tX_{s,t}} \vphi(\vx)}{\lVert \sum_{\vx \in \tX_{s,t}} \vphi(\vx) \rVert_2  + \epsilon}},
\end{equation}
where $0\!\leq\!s\!\leq\!t\!\leq\tau$ are the starting and ending frames of subsequence $\tX'_{s,t}\!\subseteq\!\tX$, and $\epsilon$ is a small constant. We normalize the pooled sequences/subsequences as described next.
\end{proposition}

\subsection{Power Normalization}
\label{sec:pns}
As discussed in Section \ref{sec:related}, we apply power normalization functions to each stream, such as ODF and SDF. We investigate three operators $g(\vpsi,\cdot)$, detailed in Remarks \ref{re:asinhe}--\ref{re:axmin}.

\begin{remark}
\label{re:asinhe}
The AsinhE function \cite{me_deeper} is an extension of the well-known power normalization (Gamma) \cite{me_deeper}, defined as $g(\vpsi, \gamma)\!=\!\sgn(\vpsi)|\vpsi|^\gamma$ for $0\!<\!\gamma\!\leq\!1$, with a smooth derivative and a parameter $\gamma'$. The AsinhE function is defined as the normalized Arcsin hyperbolic function:
\begin{align}
& \!\!\!\!\!g(\vpsi, \gamma')\!=\arcsinh(\gamma'\!\vpsi)/\arcsinh(\gamma').
\end{align}
\end{remark}

\begin{remark}
\label{re:sigme}
Sigmoid (SigmE), a max-pooling approximation \cite{me_deeper}, is an extension of the MaxExp operator defined as $g(\vpsi, \eta)\!=\!1\!-\!(1\!-\!\vpsi)^{\eta}$ for $\eta\!>\!1$.
This operator is extended to have a smooth derivative and a response defined for real-valued $\vpsi$ (rather than $\vpsi\!\geq\!0$), with a parameter $\eta'$ and a small constant $\epsilon'$:
\begin{align}
& \!\!\!\!\!g(\vpsi, \eta')\!=\!\frac{2}{1\!+\!\expl{{-\eta'\vpsi}/{(\lVert\vpsi\rVert_2+\epsilon')}}}\!-\!1.
\label{eq:sigmoid}
\end{align}
\end{remark}

\begin{remark}
\label{re:axmin}
AxMin, a piecewise linear form of SigmE \cite{me_deeper}, is given as $g(\vpsi, \eta'')\!=\!\sgn(\vpsi)\min(\eta''\vpsi/(\lVert\vpsi\rVert_2+\epsilon'), 1)$ for $\eta''\!>\!1$ and a small constant $\epsilon'$.
\end{remark}

Although these three pooling operators serve similar roles, we investigate each one because their interplay with end-to-end learning differs. 
Specifically, $\lim_{\vpsi\rightarrow\pm\infty}g(\vpsi,\cdot)$ for AsinhE and SigmE are $\pm\!\infty$ and $\pm\!1$, respectively, thus their asymptotic behaviors differ. Moreover, AxMin is non-smooth and relies on the same gradient re-projection properties as ReLU.

\subsection{Count Sketches}
\label{sec:sketch}

\begin{sloppypar}

Sketching vectors via the count sketch \cite{cormode_sketch,weinberger_sketch} is a technique for dimensionality reduction, which we apply in this paper.
\begin{proposition}
\label{pr:ten_sketch}
Let $d$ and $d'$ denote the dimensionality of the input and sketched output vectors, respectively. Let the vector $\vh\!\in\!\idx{d'}^d$ contain $d$ integers uniformly drawn from $\{1,\cdots,d'\}$, and let the vector $\vsss\!\in\!\{-1,1\}^{d}$ contain $d$ values uniformly drawn from $\{-1,1\}$. The sketch projection matrix $\mPP\!\in\!\{-1,0,1\}^{d'\times d}$ is given by:
\vspace{-0.1cm}
\begin{equation}\label{eq:sk1}
P_{ij}\!=\!
\begin{cases} s_i  & \text{if }h_i\!=\!j,
\\
0 &\text{otherwise},
\end{cases}
\end{equation}
where $s_i$ is the corresponding value from $\vsss$. The sketch projection $p: \mbr{d}\!\rightarrow\!\mbr{d'}$ is a linear operation, defined as $p(\vpsi)\!=\!\mPP\vpsi$ (or $p(\vpsi; \mPP)\!=\!\mPP\vpsi$ to explicitly highlight $\mPP$).
\begin{proof}
\vspace{-0.15cm}
This follows directly from the definition of the count sketch, as explained in Definition 1 of \cite{weinberger_sketch}.
\end{proof}
\end{proposition}

\begin{remark}
\label{re:ten_sketch}
Count sketches are unbiased estimators:\\ $\mathbb{E}_{\vh,\vsss}(p(\vpsi,\mPP(\vh,\vsss)),p(\vpsi',\mPP(\vh,\vsss)))\!=\!\left<\vpsi,\vpsi'\right>$. The variance is given by: $\mathbb{V}_{\vh,\vsss}(p(\vpsi),p(\vpsi'))\!\leq\!\frac{1}{d'}\left(\left<\vpsi,\vpsi'\right>^2 +\lVert\vpsi\rVert_2^2\lVert\vpsi'\rVert_2^2\right)$, so larger sketches reduce noisy. Thus, for each modality, we use a separate sketch matrix $\mPP$. 
\begin{proof}
\vspace{-0.15cm}
For the first and second properties, see Appendix A of \cite{weinberger_sketch} and Lemma 3 of \cite{pham_sketch}.
\end{proof}
\end{remark}

\end{sloppypar}

\subsection{Sketching the Power Normalized Vectors}

\begin{sloppypar}

\begin{proposition}
\label{prop:fact}
Sketching PN vectors increases the sketching variance (normalized by $\ell_2$ vector norms) by a factor of $1\!\leq\!\kappa\!\leq\!2$.
\begin{proof}
The variance $\mathbb{V}$ from Remark \ref{re:ten_sketch} is normalized by the norms $\lVert\vpsi\rVert_2^2\lVert\vpsi'\rVert_2^2$. Let $\mathbb{V}^{(\gamma)}$ denote the variance for $d$-dimensional vectors $\{(\vpsi^\gamma,\vpsi'^\gamma)\!:\!\vpsi\!\geq\!0,\vpsi'\!\geq\!0\}$, where the vectors are power-normalized by Gamma as described in Remark \ref{re:asinhe}. This variance is similarly normalized by $\lVert\vpsi^\gamma\rVert_2^2\lVert\vpsi'^\gamma\rVert_2^2$. 
For extreme PN ($\gamma\!\rightarrow\!0$), the variance simplifies as follows:
\begin{equation}\lim\limits_{\gamma\!\rightarrow\!0}\mathbb{V}^{(\gamma)}\!=\!\frac{1}{d'}\lim\limits_{\gamma\!\rightarrow\!0}\left(\frac{\left<\vpsi^\gamma,\vpsi'^\gamma\right>^2}{\lVert\vpsi^\gamma\rVert_2^2\lVert\vpsi'^\gamma\rVert_2^2}\!+\!1\right)\!=
\!\frac{2}{d'}.
\end{equation}
Now, assume the $d$-dimensional vectors $\vpsi$ and $\vpsi'$ are $\ell_2$-norm normalized. The ratio of variances can then be expressed as:
\begin{equation}
\kappa\!=\!\mathbb{V}/\mathbb{V}^{(\gamma)}=2/({\left<\vpsi,\vpsi'\right>^2\!+\!1}),
\end{equation}
The factor $\kappa$ depends on the pair $(\vpsi,\vpsi')$ and varies smoothly within the range $[1, 2]$ as $\gamma$ changes between 0 and 1. The Gamma is a monotonically increasing function, and for typical values such as $\gamma\!=\!0.5$, empirical data suggests $\kappa\!\approx\!1.3$.
\end{proof}
\end{proposition}

\end{sloppypar}

\subsection{Positional Embedding}
\label{sec:kernel_linearization}
Let $G_{\sigma}(\vx\!-\!\vx'\!)=\exp(-\enorm{\vx\!-\!\vx'\!}^2/{2\sigma^2})$ denote the standard Gaussian RBF kernel centered at $\vx'\!$ with bandwidth $\sigma$. Kernel linearization refers to rewriting $G_{\sigma}$ as the inner-product of two infinite-dimensional feature maps. To obtain these maps, we use a fast approximation method based on probability product kernels \cite{jebara_prodkers}. Specifically, we use the inner product of $d''$-dimensional isotropic Gaussians for $\vx,\vx'\!\!\in\!\mbr{d''}\!$. Thus, we have: 
\begin{align}
&\!\!G_{\sigma}\!\left(\vx\!-\!\vx'\!\right)\!\!=\!\!\left(\frac{2}{\pi\sigma^2}\right)^{\!\!\frac{d''}{2}}\!\!\!\!\!\!\int\limits_{\vzeta\in\mbr{d''}}\!\!\!\!G_{\sigma/\sqrt{2}}\!\!\left(\vx\!-\!\vzeta\right)G_{\sigma/\sqrt{2}}(\vx'\!\!-\!\vzeta)\,\mathrm{d}\vzeta.
\label{eq:gauss_lin}
\end{align}
Eq. \eqref{eq:gauss_lin} is approximated by replacing the integral with a sum over $Z$ pivots $\vzeta_1,\cdots,\vzeta_Z$, yielding the feature map $\vphi$ as:
\vspace{-0.1cm}
\begin{align}
&\!\!\!\!\!\!\!\!\!\!\!\!\vphi(\vx; \{\vzeta_i\}_{i\in\idx{Z}})=\left[{G}_{\sigma/\sqrt{2}}(\vx-\vzeta_1),\cdots,{G}_{\sigma/\sqrt{2}}(\vx-\vzeta_Z)\right]^T\!\!\!\!,\!\!\!\!\label{eq:gauss_lin2a}\\
\!\!\text{ and } & G_{\sigma}(\vx\!-\!\vx'\!)\approx\left<\sqrt{c}\vphi(\vx), \sqrt{c}\vphi(\vx'\!)\right>,
\label{eq:gauss_lin2}
\end{align}
where $c$ is a constant. Eq. \eqref{eq:gauss_lin2} is the linearization of the RBF kernel, and Eq. \eqref{eq:gauss_lin2a} defines the feature map. The pivots $\{\vzeta_i\}_{i\in\idx{Z}}$ are evenly spaced in the interval $[0;1]$, with $Z$ total pivots. For simplicity, we drop the explicit notation of the pivots $\{\vzeta_i\}_{i\in\idx{Z}}$ and write $\vphi(\vx)$, \etc.

\begin{sloppypar}

\subsection{\lei{Action Recognition Backbones}}
\label{sec:backbone}

\lei{Below, we present six mainstream models used as backbones for feature hallucination.}

\lei{{\bf I3D}~\cite{i3d_net} (Inflated 3D ConvNet) is a two-stream architecture that extends 2D ConvNet (pre-trained on ImageNet-1K~\cite{5206848}) by inflating their kernels to 3D, enabling spatiotemporal feature extraction. The model has become one of the most widely adopted frameworks for video processing tasks. Variants such as S3D~\cite{Xie_2018_ECCV} build on the I3D architecture by introducing modifications to its modules to enhance efficiency and performance. In our work, we adapt I3D by removing its final convolutional layer and classifier to construct the backbone network. The pre-trained weights from Kinetics-400 are used for feature hallucination, allowing us to leverage its robust video representation capabilities. }

\lei{{\bf AssembleNet}~\cite{assemblenet} is a family of learnable models designed to optimize the `connectivity' among features across input modalities, tailored for specific target tasks such as AR. Its extension, {\bf AssembleNet++}~\cite{assemblenet_plus}, introduces peer-attention mechanisms that enable the model to learn the interactions and relative importance of features, particularly between semantic object information and raw appearance and motion features. By removing the classification layer, we extract a 2048-dimensional output from the 3D average pooling layer as the intermediate representation for feature hallucination. For our backbone, we use AssembleNet++ pre-trained on Kinetics-400, using its robust capability to capture spatiotemporal and multimodal information.}

\lei{{\bf VTN}~\cite{10.1145/3341105.3373906} uses the latest advancements in Vision Transformer (ViT) for computer vision tasks, applying it to AR. VTN consists of two main components: (i) an encoder that processes each frame of the input sequence independently using a 2D CNN to generate frame embeddings (pre-trained models are used to maximize the benefits of transfer learning from image classification tasks), and (ii) a decoder that integrates intra-frame temporal information in a fully-attentional, feed-forward manner, ultimately producing classification scores for the video clip. 
For our backbone, we remove the final classification layer of VTN. We adopt the default model hyperparameters as specified in~\cite{10.1145/3341105.3373906}, such as 4 stacked decoder blocks, 8 attention heads, and frame embeddings of size 512. For the encoder, we use MobileNet V2~\cite{Sandler_2018_CVPR} and ResNet-34~\cite{7780459}, forming two different backbones: VTN-MobileNet and VTN-ResNet.}

\lei{{\bf FASTER}~\cite{faster} is a general framwork designed to aggregate both expensive and lightweight representations from different clips. It combines an expensive model, which captures detailed action information (\eg, subtle motion), and a lightweight model, which tracks scene changes over time to minimize redundant computation between neighboring clips. This approach ensures global coverage of the entire video at a low cost by using FAST-GRU to aggregrate representations from different clip models. For the clip-level backbone in FASTER, we follow the method outlined in ~\cite{faster}, selecting R(2+1)D-50~\cite{tran2018closer} as the expensive model and R2D-26~\cite{tran2018closer} as the lightweight model. We form our backbone by removing the final classification layer.
In our experiments, we choose a clip length of $L\!=\!8$ and use 8 clips.}

\lei{{\bf VideoMAE}~\cite{tong2022videomae} is a cutting-edge self-supervised video pre-training framework that introduces video tube masking with a high masking ratio. This design makes video reconstruction a more challenging and meaningful self-supervision task, encouraging the extraction of more effective video representations during pre-training. 
Its advanced version, \textbf{VideoMAE V2}~\cite{wang2023videomae}, scales VideoMAE further by introducing a dual masking strategy for improved efficiency. This approach applies a masking map to both the encoder and decoder, significantly reducing the decoder's input length while maintaining effectiveness. 
For feature hallucination, we use the VideoMAE V2 pre-trained encoder on UnlabeledHybrid~\cite{wang2023videomae}, with ViT-g as the backbone.}

\lei{{\bf InternVideo2}~\cite{wang2024internvideo2} is a state-of-the-art video foundation model excelling in video recognition, video-text tasks, and video-centric dialogue. It employs a progressive training approach that integrates masked video modeling, cross-modal contrastive learning, and next-token prediction, scaling the video encoder to an impressive 6 billion parameters. The training unfolds in three stages: first, video data is fed into the model to reconstruct unmasked video tokens, maximizing the capture of spatiotemporal visual concepts. Second, the model aligns video with audio, speech, and text through cross-modal contrastive learning, enriching it with semantic information. Finally, the model predicts the next token using video-centric inputs, embedding even deeper semantics. For our feature hallucination backbone, we use InternVideo2-1B, trained in the first stage, denoted as InternVideo2$_{s1}$.}

\end{sloppypar}

\newcommand\id{\ensuremath{\mathbbm{1}}}

\section{Approach}
\label{sec:approach}
\begin{sloppypar}

Our pipeline, illustrated in Fig. \ref{fig:pipe}, comprises the following components: (i) BoW/FV/OFF streams (dashed black), (ii) Object Detection Features (ODF) and Saliency Detection Features (SDF) streams, (iii) the GCN-encoded Skeleton Features (GSF) stream, (iv) the SoundNet-encoded Audio Features (AF) stream, (v) the High Abstraction Features (HAF) stream, (vi) the Prediction Network (abbreviated as PredNet), and (vii) the Covariance Estimation Network (CENet), which facilitates uncertainty-aware feature descriptor learning.

Each stream begins by processing the intermediate representations generated by the backbone from the RGB frames. These representations are refined through a hallucination process to approximate features using Mean Squared Error (MSE) loss between the ground-truth features and the outputs of the hallucinated streams. 
The HAF stream enhances the backbone representations before integrating them with the hallucinated streams. PredNet then combines the outputs from all streams, BoW, FV, OFF, HAF, ODF, SDF, GSF and AF, to learn action concepts for classification. 

The following sections detail our method: \lei{we start by describing the extraction of ODF and SDF descriptors}. Next, we explain the hallucination streams and the computation of ground-truth features, followed by a discussion on uncertainty-aware learning and the design of CENet. Finally, we introduce the objective function used for feature hallucination.

\subsection{Statistical Motivation}
\label{sec:statmot}
Before introducing our ODF and SDF descriptors, we highlight the importance of higher-order statistics in capturing the nuanced characteristics of video data~\cite{me_tensor_eccv16, kon_tpami2020a,chen2020homm, kon_tpami2020b}. 
Comparing videos requires more than simple feature matching; it necessitates robust representations of the underlying distribution of local features (\eg, detection scores) or descriptors. The characteristic function, $\varphi_\Upsilon(\boldsymbol{\omega})$, provides a comprehensive description of this distribution by representing the probability density $f_\Upsilon(\vupsilon)$ of the features $\vupsilon\!\sim\!\Upsilon$: $\varphi_\Upsilon(\boldsymbol{\omega})\!=\!\mathbb{E}_{\vupsilon\sim\Upsilon}\left(\exp(\iu\boldsymbol{\omega}^T\!\vupsilon)\right)$. Using a Taylor series expansion, the characteristic function can be expressed as:
\begin{align}
&\!\!\!\!\!\!\!\!\!\mathbb{E}_{\vupsilon\sim\Upsilon}\left(\sum\limits_{r=0}^\infty\frac{\iu^j}{r!}\left<\vupsilon,\boldsymbol{\omega}\right>^r\right)\!\approx\!\frac{1}{N}\sum\limits_{n=0}^N\sum\limits_{r=0}^\infty\frac{\iu^r}{r!}\left<{\kronstack}_r\vupsilon_n,{\kronstack}_r\boldsymbol{\omega}\right>\!=\!\!\!\!\\
&\!\!\!\!\!\sum\limits_{r=0}^\infty\frac{\iu^r}{r!}\bigg<\frac{1}{N}\sum\limits_{n=0}^N{\kronstack}_r\vupsilon_n,{\kronstack}_r\boldsymbol{\omega}\bigg>\!=\!\sum\limits_{r=0}^\infty\left<\tX^{(r)},\frac{\iu^r}{r!}{\kronstack}_r\boldsymbol{\omega}\right>\!,\nonumber
\end{align}
where $\iu$ is the imaginary unit, $\kronstack_r$ is the $r$-th Kronecker power, and $\tX^{(r)}$, defined as $\tX^{(r)}\!=\!\frac{1}{N}\!\sum\limits_{n=0}^N\!{\kronstack}_r\vupsilon_n$, is a tensor capturing the $r$-th order moment of the feature distribution.

In principle, with infinite data and infinite moments, one can fully capture  $f_\Upsilon(\vupsilon)$. In practice, first-, second- and third-order moments are typically sufficient, however, second- and third-order tensors grow quadratically and cubically with respect to the size of $\vupsilon$. Thus, in what follows, we represent second-order moments not by a covariance matrix but by the subspace corresponding to the top $n'$ leading eigenvectors. We also make use of the corresponding eigenvalues of the signal. Finally, it suffices to notice that $\vkappa^{(r)}\!=\!\text{diag}\!\left(\tX^{(r)}\right)$ corresponds to the notion of order $r$ cumulants used in calculations of skewness ($r\!=\!3$) and kurtosis ($r\!=\!4$) but it grows linearly with respect to the size of $\vupsilon$. Thus, in what follows, we use the $\ell_2$ norm normalized mean, leading eigenvectors (and trace-normalized eigenvalues), skewness and kurtosis (rather than coskewness and cokurtosis) to obtain compact representation of ODF and SDF.

\subsection{Object Detection Features}
\label{sec:odf}

Each object bounding box is described by the feature vector: 
\begin{align}
& \!\!\!\vupsilon\!=\!\left[\vdelta(y_{(det)}); \vy_{(inet)}; \vphi(\varsigma); \circledcirc_{i\in\idx{4}}\vphi(v_i); \vphi\!\left(\frac{\scriptstyle t\!-\!1}{\scriptstyle\tau\!-\!1}\right)\right]\!\in\!\mbr{d}\!\!,
\label{eq:det-feat}
\end{align}
where $\vdelta\!=[0,\cdots,1,\cdots,0]^T$ is a vector with all zeros except for a single $1$ at the location $y$. Since there are 91 object
classes for detectors trained on the COCO dataset and 80 classes for those trained on the AVA
v2.1 dataset, we assume $y_{(det)}\!\in\!\idx{91\!+\!80}$, where the labels $0,\cdots,90$ correspond to COCO classes, and classes $91,\cdots,79\!+\!91$ correspond to AVA v2.1 classes. Additionally, $\vy_{(inet)}\!\in\!\mbr{1001}$ represents an $\ell_1$-norm normalized ImageNet-1K classification score, $0\!\leq\!\varsigma\!\leq\!1$ is the detector confidence score, $v_0,\cdots,v_4$ are the normalized Cartesian coordinates (top-left and bottom-right) of a bounding box in the range $[0;1]$, and $(t\!-\!1)/(\tau\!-\!1)$ is the normalized frame index with respect to the total sequence length $\tau$. For feature maps $\vphi(\cdot)$ defined in Eq. \eqref{eq:gauss_lin2a}, we use $Z\!=\!7$ pivots and set the RBF $\sigma$ to $0.5$. 

For all detections per video from a given detector, we first compute the mean $\vmu([\vupsilon_1,\cdots,\vupsilon_N])\!\in\!\mbr{d}$ (denoted as $\vmu$), where $N$ is the total number of detections. We then form a matrix $\mupsilon\!\in\mbr{d\!\times\!N}$:
\begin{align}
& \!\!\!\mupsilon\!=\!\frac{\scriptstyle 1}{ \scriptstyle J}\!\left[\frac{\scriptstyle 1}{\scriptstyle K_1}\!\left[\circledcirc^2_{i\in\idx{K_1}}\!(\vupsilon_{i1}\!-\!\vmu)\right],\cdots,\frac{\scriptstyle 1}{\scriptstyle K_J}\!\left[\circledcirc^2_{i\in\idx{K_J}}\!(\vupsilon_{iJ}\!-\!\vmu)\right]\right]\!,\!\!
\label{eq:mat-m}
\end{align}
where $K_j$ denotes the number of detections per frame $j\!\in\!\idx{J}$. 
From this, we extract higher-order statistical moments as described below. Since $N$ is large and its size varies across videos, hallucinating $\mupsilon$ directly is infeasible (and lacks invariance  
properties). 

First, we compute $\mU\mLambda\mV\!=\!\text{svd}\left(\mupsilon\right)$, rather than $\mU\mLambda^2\mU^T\!=\!\text{eig}\left(\mupsilon\!\mupsilon^T\right)$ since $N\!\ll\!d$, where $\mU\!=\![\vu_1,\vu_2,\cdots]$. 
We take $\tX^{(r)}\!\left(\{\mvv\!\!-\!\!\vmu\}_{n\!=\!0}^N\right)$ (abbreviated as $\tX^{(r)}$) and define $\vkappa^{(r)}\!=\!\text{diag}\!\left(\tX^{(r)}\right)$ as described in Section \ref{sec:statmot}. We then form our multi-moment descriptor $\vpsi_{(det)}\!\in\!\mbr{d(4+n')}$, $n'\!\ge\!1$:
\vspace{-0.3cm}
{\fontsize{8.5}{9}\selectfont
\begin{align}
&\label{eq:moment1}\!\!\vpsi_{(det)}\!=\left[\frac{\vmu}{||\vmu||_2}; \circledcirc^2_{i\in\idx{n'}} \vu_i\left(\tX^{(2)}\right); 
\frac{\vkappa^{(3)}}{\left(\vkappa^{(2)}\right)^{3/2}}; \frac{\vkappa^{(4)}}{\left(\vkappa^{(2)}\right)^{2}};  \frac{\text{diag}(\mLambda^2)}{\sum_i\!\lambda^2_{ii}\!}\!\right]\!.
\end{align}
}

\vspace{-0.1cm}
\noindent
The composition of Eq. \eqref{eq:moment1} is explained in Section \ref{sec:statmot}. It is easy to verify that $\frac{\vkappa^{(3)}}{\left(\vkappa^{(2)}\right)^{3/2}}$ and $\frac{\vkappa^{(4)}}{\left(\vkappa^{(2)}\right)^{2}}$ are the empirical versions of skewness and kurtosis, given by $\frac{\mathbb{E}_{\vupsilon\sim\Upsilon}\!\left((\vupsilon\!-\!\vmu)^3\right)}{\mathbb{E}^{3/2}_{\vupsilon\sim\Upsilon}\!\left((\vupsilon\!-\!\vmu)^2\right)}$ and $\frac{\mathbb{E}_{\vupsilon\sim\Upsilon}\!\left((\vupsilon\!-\!\vmu)^4\right)}{\mathbb{E}^{2}_{\vupsilon\sim\Upsilon}\!\left((\vupsilon\!-\!\vmu)^2\right)}$, respectively.

\subsection{Saliency Detection Features}
\label{sec:sdf}

We extract directional gradients from saliency frames using  discretized gradient operators $[-1,0,1]$ and $[-1,0,1]^T$, obtaining gradient amplitude and orientation maps, $\mLLa$ and $\mTheta$, for each frame, encoded as follows:
\begin{align}
& \!\!\!\!\vupsilon'_{(sal)}=\!\!\!\!\!\!\!\!\!\sum\limits_{i\in\idx{W},j\in\idx{H}}\!\!\!\!\!\!\!\!\Lambda_{ij}\vphi(\theta_{ij}/(2\!\pi))\otimes\vphi\left(\frac{\scriptstyle i\!-\!1}{\scriptstyle  W\!-\!1}\right)\otimes\vphi\left(\frac{\scriptstyle  j\!-\!1}{\scriptstyle  H\!-\!1}\right)\!,
\label{eq:sal}
\end{align}
where $\otimes$ represents the Kronecker product, and $\vphi(\theta)$ follows Eq. \eqref{eq:gauss_lin2a}, except that the assignment to Gaussians is performed in the modulo ring to respect the periodic nature of $\theta$. 
We encode $\vphi(\theta)$ with 12 pivots to capture the orientation of gradients. 
The remaining maps $\vphi(\cdot)$ are encoded with 5 pivots each, corresponding to spatial binning. 
Note that $\vupsilon'_{(sal)}$ (denoted as $\vupsilon'$) is conceptually similar to a single CKN layer \cite{ckn}, but simpler: for one-dimensional variables, we sample pivots (similar to learning) for the maps $\vphi(\cdot)$.
Each saliency frame is represented as a feature vector: $\vupsilon^\dagger\!=\!\left[\scriptstyle\vupsilon'/\scriptstyle||\vupsilon'||_2;\; \scriptstyle\mI_{:}/\scriptstyle||\mI_{:}||_1 \right]\!\in\!\mbr{d^\dagger}\!$, 
where $\mI_{:}$ is a vectorized low-resolution saliency map. Thus, $\vupsilon^\dagger$ captures both the directional gradient statistics and the intensity-based gist of the saliency maps.

Next, we compute the mean $\vmu([\vupsilon^\dagger_1,\cdots,\vupsilon^\dagger_J])\!\in\!\mbr{d^\dagger}$ (denoted as $\vmu$), where $J$ is the total number of frames per video. We then obtain $\mupsilon^\dagger\!=\!\left[\vupsilon^\dagger_1,\cdots,\vupsilon^\dagger_J\right]\!/\!J\in\!\mbr{d^\dagger\!\times\!J}\!\!$, 
which is compactly described by the multi-moment expression in Eq. \eqref{eq:moment1}, resulting in the multi-moment descriptor $\vpsi_{(sal)}\!\in\!\mbr{d(4+n^\dagger)}$.

\begin{figure}[tbp]%htbp % left bottom right top
\centering%%%%
\begin{subfigure}[b]{0.98\linewidth}
\centering\includegraphics[trim=0 0 0 0, clip=true,width=0.95\linewidth]{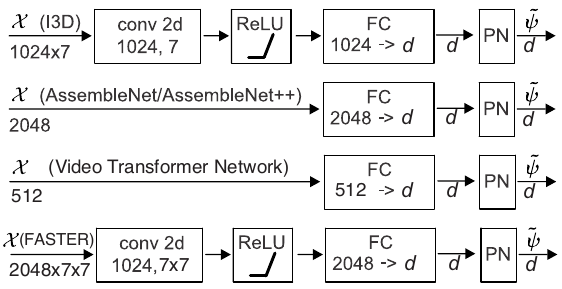}
\vspace{-0.2cm}
\caption{\label{fig:stra}}
\end{subfigure}
\\
\begin{subfigure}[b]{0.98\linewidth}
\centering\includegraphics[trim=0 0 0 0, clip=true,width=0.75\linewidth]{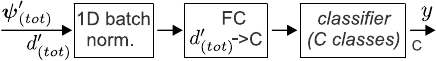}
\vspace{-0.2cm}
\caption{\label{fig:strc}}
\vspace{-0.2cm}
\end{subfigure}
\caption{Stream details. Figure \ref{fig:stra} shows the architecture we use for the BoW, FV, OFF, HAF, ODF, SDF, GSF, and AF streams for four different backbones. Figure \ref{fig:strc} shows the architecture of our PredNet. The operations and their parameters are specified within each block (\eg, {\em conv2d} with its filter count/size, and {\em Power Normalization} ({\em PN})). The input and output sizes are indicated under the corresponding arrows. For our experiments, we select $d\!=\!128$, $256$ or $512$, depending on the dataset.
}\vspace{-0.3cm}
\label{fig:streams}
\end{figure}

\subsection{Hallucinating Streams}
\label{sec:hal-stream}

Each stream processes the intermediate representation, $\mathcal{X}_{(rgb)}$, which is obtained either by removing the classifier and/or the final 1D convolutional layer of the backbone network, as detailed in Section~\ref{sec:backbone}, or by pooling spatiotemporal token embeddings, as used in VideoMAE V2 and InternVideo2. 
To implement the pipeline for each stream, we follow the approach in~\cite{Wang_2019_ICCV}, using a Fully Connected (FC) unit.
Furthermore, each stream is equipped with a PN module. For the PN implementation, we explore three variants: AsinhE, SigmE, and AxMin, described in Remarks \ref{re:asinhe}, \ref{re:sigme} and \ref{re:axmin}, respectively.
Below, we provide a detailed explanation of each stream and its corresponding ground truth. 
It is important to note that ground-truth features are used exclusively during the training phase to train the hallucination streams.

\noindent{\bf BoW/FV.} As FV captures both first- and second-order statistics, we use a separate stream for each type of statistic. 
For BoW, we follow the process outlined in Section \ref{sec:enc}. Specifically, we apply k-means to build a 1000-dimensional dictionary using the same descriptors employed for precomputing FV. The descriptors are then encoded according to Eq. \eqref{eq:sp1}, aggregated using the steps described in Section \ref{sec:aggr}, and normalized with PN as discussed in Section \ref{sec:pns}. Where applicable, we use 4000-dimensional dictionary for BoW and apply sketching to reduce the vector size to $d$ dimensions.
To train Fisher Vectors, we compute 256-dimensional GMM-based dictionaries on descriptors derived from IDT \cite{improved_traj}, following the steps detailed in Sections  \ref{sec:related} and \ref{sec:enc}. PCA is applied to the trajectory (30 dimensions), HOG (96 dimensions), HOF (108 dimensions), MBHx (96 dimensions) and MBHy (96 dimensions) features, yielding a final 213-dimensional local descriptor. 
The encoded first- and second-order FV representations, each of size $256\!\times\!213\!=\!54528$, are sketched to $d$ dimensions as described in Section \ref{sec:sketch}. For this purpose, we prepare matrices $\mPP_{(fv1)}$ and $\mPP_{(fv2)}$ as defined in Proposition \ref{pr:ten_sketch}.
The sketched first- and second-order representations,  $\vec{\psi}'_{(fv1)}\!=\!\mPP_{(fv1)}\vec{\psi}_{(fv1)}$ and $\vec{\psi}'_{(fv2)}\!=\!\mPP_{(fv2)}\vec{\psi}_{(fv2)}$, can then be seamlessly integrated with the loss functions detailed in Section \ref{sec:obj}.

\noindent{\bf I3D OFF.} We use the I3D optical flow pre-trained on Kinetics-400 as the feature extractor to obtain the OFF (Fig. \ref{fig:pipe}), denoted as $\vec{\psi}_{(off)}$. These features have a dimensionality of 1024, which we sketch to $d$ dimensions before using them as training ground truth for the OFF layer. LDOF~\cite{brox_largedisp} is used as it effectively handles large displacements.

\noindent{\textbf{ODF and SDF.}} 
The ODF ground-truth training representations are of size $1214\!\times\!N$, where $N$ is the total number of bounding boxes per video (ranging from 50 to 10,000). The 1214-dimensional features consist of: $80\!+\!91$ one-hot encoding for detection classes, $6\!\times\!7$ values representing $\vphi(\cdot)$-embedded confidence scores, bounding box coordinates, and frame numbers, and 1001-dimensional ImageNet scores. 
An alternative representation without the RBF embedding, $\vphi(\vx) = \vx$, results in features of size $1178 \times N$.
The SDF ground-truth training representations have a size of $556\!\times\!J$, where $J$ is the number of frames per video.  These features include: 300 dimensions ($12\!\times\!5\!\times\!5$) capturing spatio-angular gradient distributions, and 256 dimensions ($16\!\times\!16$) capturing the luminance of saliency maps. 
Both ODF and SDF are encoded per video using the multi-moment descriptor in Eq. \eqref{eq:moment1}, producing compact representations of size  $1178\!\times\!(4+n')$ and  $556\!\times\!(4+n^\dagger)$, respectively, where $n'$ and $n^\dagger$ are varied between 1 and 5. 
These representations are Power Normalized with SigmE and sketched to $d$ dimensions via $\vec{\psi}'_{(\cdot)}\!=\!\mPP_{(\cdot)}\vec{\psi}_{(\cdot)}$. The resulting features are fed into loss functions such as MSE. Ground-truth representations are used only during training.

\noindent{\bf Skeleton features.} We extract skeleton features (GSF) using ST-GCN pre-trained on large-scale Kinetics-skeletons. These 400-dimensional features are sketched to $d$ dimensions as $\vec{\psi}'_{(gsf)}\!=\!\mPP_{(gsf)}\vec{\psi}_{(gsf)}$ and used as ground truth for training. 
For datasets lacking skeleton data (\eg, MPII Cooking Activities, Charades, and EPIC-Kitchens), GSF training is disabled. However, during testing, this stream can leverage pre-trained models (\eg, from Toyota Smarthome) to generate hallucinated features.

\noindent{\bf Audio features.} Audio features (AF) are extracted using the 8-layer SoundNet model pre-trained on two-million unlabeled videos~\cite{soundnet}. For datasets with audio, the audio is extracted from video using FFmpeg~\cite{tomar2006converting} with a bit rate of 160 kbps, two audio channels, and a sample rate of 44100 Hz. If a video contains sound, a corresponding .wav file is generated. Follow~\cite{soundnet}, the audio is downsampled to 22 kHZ and converted to a single channel for efficiency, with a minor trade-off in sound quality. The waveform is scaled to the range [-256, 256] for feature extraction from the {\tt pool5} layer. 
These features are sketched to $d$ dimensions as ground truth. The AF stream is trained only if the dataset contains audio.

\noindent{\bf HAF.} Each stream applies PN to hallucinated features, aligning the hallucinated output $\vec{\tilde{\psi}_{(\cdot)}}$ (of $d$ dimensions) with the ground-truth features $\vec{\psi}'_{(\cdot)}$ described earlier. 
High Abstraction Features (HAF) follow the same steps, combining with other streams and passing the backbone features into PredNet (see Fig. \ref{fig:pipe}). While hallucinated streams co-supervise the backbone through external ground-truth tasks, HAF directly processes backbone features for PredNet.

\noindent{\textbf{PredNet.}} The final component of the pipeline, PredNet, is illustrated in Figure \ref{fig:strc}. Its input is $\vec{\psi}_{(tot)}$ (unsketching) or $\vec{\psi}'_{(tot)}$ (sketched). This input is passed through batch normalization, followed by a fully connected (FC) layer that produces a $C$-dimensional representation. This output is optimized using cross-entropy loss.

\subsection{Uncertainty Learning}
\label{sec:uncertainty}

We concatenate the ground truth features of BoW, FV, I3D OFF, ODF, SDF, GSF, and AF into a column feature vector: $\vpsi'\!=\![\vpsi'_{bow}, \vpsi'_{fv1}, \vpsi'_{fv2}, \vpsi'_{off}, \vpsi'_{odf}, \vpsi'_{sdf}, \vpsi'_{gsf}, \vpsi'_{af}]^T\!\in\!\mbr{d'}$. Similarly, the hallucinated features are concatenated into a column vector: $\tilde{\vpsi}\!=\![\tilde{\vpsi}_{bow}, \tilde{\vpsi}_{fv1}, \tilde{\vpsi}_{fv2}, \tilde{\vpsi}_{off}, \tilde{\vpsi}_{odf}, \tilde{\vpsi}_{sdf}, \tilde{\vpsi}_{gsf}, \tilde{\vpsi}_{af}]^T\!\in\!\mbr{d'}$.

Mean Squared Error (MSE) assumes the errors across all features are independent and identically distributed (i.i.d.) or, equivalently, that the covariance matrix is diagonal~\cite{DBLP:journals/corr/BurdaGS15,DBLP:journals/corr/KingmaW13}. While this approach enables local noise level estimation, it makes a limiting and often flawed assumption that residuals (errors) across features are uncorrelated. 
To address this, we extend the noise model to use a multivariate Gaussian likelihood with a full covariance matrix. This matrix captures feature correlations, allowing for structured residual sampling. A maximum likelihood approach is used to train the covariance prediction. 
Given the hallucinated features $\tilde{\vec{\psi}'}$ and the ground truth features $\vec{\psi}'$, we replace the MSE loss with the following uncertainty learning objective:
\begin{align}
& \quad\argmax \log\mathcal{N}(\tilde{\vpsi}; \vpsi', \cov) \nonumber\\
&=\argmax \log\frac{1}{(2\pi)^{\frac{d'}{2}}|\cov|^{\frac{1}{2}}}e^{\!-\frac{1}{2}(\tilde{\vpsi}\!-\!\vpsi')^\text{T}\cov^{-1}(\tilde{\vpsi}\!-\!\vpsi')} \nonumber\\
&=\argmin\!\frac{d'}{2}\!\log(2\pi)\!+\!\frac{1}{2}\log(|\cov|)\!+\!\frac{1}{2}(\tilde{\vpsi}\!-\!\vpsi')^\text{T}\cov^{-1}(\tilde{\vpsi}\!-\!\vpsi') \nonumber\\
&\approx\argmin \log(|\cov|)+(\tilde{\vpsi}\!-\!\vpsi')^\text{T}\cov^{-1}(\tilde{\vpsi}\!-\!\vpsi'),
\label{eq:uncer}
\end{align}
where $\cov$ is the covariance matrix, and $d'$ is the feature dimension. Note that the constant term is removed, as it does not influence the optimisation. To estimate the covariance matrix $\cov$, we use a Covariance Estimation Network (CENet), discribed in the following section.

\subsection{Covariance Estimation Network}
\label{sec:cenet}

A deep neural network is used to estimate the covariance matrix $\cov$, using the latent representation $\mathcal{X}_{(rgb)}$ as input. By definition, $\cov$ is symmetric and positive definite. For a feature vector $\vpsi'$ of dimensionality $d'$, the covariance matrix contains $(d'^2\!-\!d')/2\!+\!d'$ unique parameters. This matrix captures structured information about reconstruction uncertainty, allowing the hallucinated output to more closely approximate the ground truth features.

Since $\cov$ appears in its inverted form ($\boldsymbol{\Omega}\!=\!\cov^{-1}$) in the negative log-likelihood calculation (Eq.~\eqref{eq:uncer}), it is more practical to estimate the precision matrix $\boldsymbol{\Omega}$ directly. This also simplifies the log determinant computation, as $\log(|\cov|)\!=\!-\log(|\boldsymbol{\Omega}|)$. Rewriting Eq.~\eqref{eq:uncer}, the objective function for feature hallucination becomes:
\begin{equation}
\argmin(\tilde{\vpsi}\!-\!\vpsi')^\text{T}\boldsymbol{\Omega}(\tilde{\vpsi}\!-\!\vpsi')\!-\!\kappa\log(|\boldsymbol{\Omega}|),  
\end{equation}
where $\kappa\!\geq\!0$ (typically $\kappa\!\neq\!d'$) adjusts the penalty for large uncertainty.

Using Cholesky decomposition, the precision matrix can be represented as $\boldsymbol{\Omega}\!=\!\boldsymbol{\omega}\boldsymbol{\omega}^\text{T}$, where $\boldsymbol{\omega}$ is a lower triangular matrix. The covariance network estimates $\boldsymbol{\omega}$ explicitly. With this decomposition, it is trivial to evaluate two terms in Eq.~\eqref{eq:uncer}: (i) the reconstruction error: $\vy^\text{T}\vy$, where $\vy\!=\!\boldsymbol{\omega}^\text{T}(\tilde{\vpsi}\!-\!\vpsi')$, and (ii) the log determinant: $\log(|\cov|)\!=\!-2\sum_i^{d'}\log\omega_{i, i}$, where $\omega_{i, i}$ represents the $i$-th diagonal element of $\boldsymbol{\omega}$. 
To ensure $\boldsymbol{\Omega}$ is positive-definite, the diagonal elements are constrained to be strictly positive, for instance, by estimating $\log\omega_{i, i}$ using a network.

\noindent{\bf Computational complexity.} Sampling from $\cov$ involves solving a triangular system of equations via backward substitution, which requires $O{(d'^2)}$ operations. The number of parameters to estimate grows quadratically with $d'$, making direct estimation feasible only for features with small dimensionality.

\noindent{\bf Sparse Cholesky decomposition.} To scale to higher-dimensional feature vectors, we impose a fixed sparsity pattern on $\boldsymbol{\omega}$, estimating only the non-zero values. Specifically, $\omega_{i, j}$ is non-zero only if $i$ and $j$ are within the same batch of sampled feature indices.
This reduces the maximum number of non-zero elements in $\boldsymbol{\omega}$ to $(d^{*2}\!-\!d^*)/2\!+\!d'$, where $d^* \ll d'$ represnts the batch size. 
We set 
$d^*$ to be the number of feature types, ensuring each sampled index corresponds to a feature within a specific type. 
With the sparsity, the uncertainty model for each feature representation resembles a Gaussian Random Field for residuals. A zero value in the precision matrix $\boldsymbol{\Omega}$ at position $(i, j)$ indicates conditional independence between feature $i$ and $j$.

\noindent{\bf Parallel computing.} This sparse representation allows efficient evaluation of the uncertainty model without constructing a full dense matrix. Similarly, sampling can be performed by solving a sparse system of equations. The method is GPU-parallelizable, as each batch can be evaluated independently.

\noindent{\bf CENet architecture.} The Covariance Estimation Network (CENet) is composed of several key components designed to estimate the precision matrix $\boldsymbol{\Omega}$ efficiently and ensure it is symmetric and positive-definite. First, the network begins with a simple Multi-Layer Perceptron (MLP) that includes two fully connected (FC) layers, a batch normalization layer, and a ReLU activation function applied between them. This MLP processes the input feature vector to generate a refined representation.

Next, a zero padding layer expands the output dimensions from $(d^{*2}\!-\!d^*)/2\!+\!d'$ to $d'\!\cdot\!d'$, aligning it with the dimensionality required for the precision matrix. The expanded vector is then reshaped into a square matrix of size $d'\!\times\!d'$. To enforce the positive-definiteness of the precision matrix, an exponential block is applied to the diagonal elements of the reshaped matrix. This ensures that the diagonal entries are strictly positive by transforming them to remove the logarithmic scale introduced during estimation.

Finally, the precision matrix $\boldsymbol{\Omega}$ is derived using the Cholesky decomposition, $\boldsymbol{\Omega}\!=\!\boldsymbol{\omega}\boldsymbol{\omega}^\text{T}$, where $\boldsymbol{\omega}$ is a lower triangular matrix estimated by the network. This decomposition guarantees the symmetry and positive-definiteness of $\boldsymbol{\Omega}$.

The input to the CENet can be either (i) the intermediate representation $\mathcal{X}_{(rgb)}$, or (ii) the concatenated hallucinated feature $\tilde{\vpsi}$, as described earlier. Both variants are evaluated to determine the optimal input data for CENet's operation.

\subsection{Objective and its Optimization}
\label{sec:obj}

\noindent\textbf{Objective function.} During training, we optimize a combined loss function that incorporates an uncertainty learning term for training hallucination streams and a classification loss:

\vspace{-0.4cm}
\begin{align}
&\ell^*(\tX, \vy; \bar{\mP})\!=\alpha\left((\tilde{\vpsi}\!-\!\vpsi')^\text{T}\boldsymbol{\Omega}(\tilde{\vpsi}\!-\!\vpsi')\!-\!\kappa\log(|\boldsymbol{\Omega}|)\right) \!\nonumber\\
& \qquad\qquad\qquad\qquad+\! \ell\!\left(f\!(\vpsi'_{(tot)}; \mP_{(pr)}),\vy; \mP_{(\ell)}\right), \nonumber\\
&\qquad\text{ where: } \forall i\!\in\!\mathcal{H}, \vec{\tilde{\psi}}_i\!=\!g\!\left(\hslash(\tX, \mP_{i}),\eta\right), \vpsi'_i\!=\!\mPP_i\vpsi_i,\nonumber\\
&\qquad\qquad\qquad\qquad\;\tilde{\vpsi}\!=\!\oplus_{i\in\mathcal{H}}\vec{\tilde{\psi}}_i, \vpsi'\!=\!\oplus_{i\in\mathcal{H}}\vpsi'_i, \nonumber\\
&\qquad\qquad\qquad\qquad\;\vpsi_{(haf)}\!=\!g\!\left(\hslash(\tX,\mP_{(haf)}),\eta\right),\nonumber\\
&\qquad\qquad\qquad\qquad\;\vpsi'_{(tot)}\!=\!\mPP_{(tot)}\left[\tilde{\vpsi}; \vpsi_{(haf)}\right],\nonumber\\
&\qquad\qquad\qquad\qquad\;\boldsymbol{\Omega}\!=\!\boldsymbol{\omega}\boldsymbol{\omega}^\text{T}, \boldsymbol{\omega}\!=\!c(\tX,\mP_{(cov)})  \nonumber \\
&\qquad\qquad\qquad\qquad\qquad\qquad\;\text{or }\boldsymbol{\omega}\!=\!c(\vec{\tilde{\psi}},\mP_{(cov)}).
\label{eq:loss}
\end{align}
The equation above represents a trade-off between the uncertainty learning term, $(\tilde{\vpsi}\!-\!\vpsi')^\text{T}\boldsymbol{\Omega}(\tilde{\vpsi}\!-\!\vpsi')\!-\!\kappa\log(|\boldsymbol{\Omega}|)$, and the classification loss, $\ell(\cdot,\vy; \mP_{(\ell)})$, with labels $\vy\!\in\mathcal{Y}$ and parameters $\mP_{(\ell)}\!\equiv\!\{\mW,\vb\}$. 
The trade-off is controlled by the constant $\alpha\!\geq\!0$. Uncertainty is computed over hallucination streams $i\!\in\!\mathcal{H}$, where $\mathcal{H}\!\equiv\!\left\{(bow),(fv1),(fv2),(off),(odf),(sdf),(gsf),(af)\right\}$, a set of hallucination streams that can be adjusted based on the task at hand. 
Additionally, $g(\cdot,\eta)$ is the Power Normalization function described in Section \ref{sec:pns}, and 
$c(\cdot, \mP_{(cov)})$ is the CENet module with parameters $\mP_{(cov)}$. 
The PredNet module, $f(\cdot; \mP_{(pr)})$, has learnable parameters $\mP_{(pr)}$.
The hallucination streams $\{\hslash(\cdot, \mP_{i}), i\!\in\!\mathcal{H}\}$ produce the corresponding hallucinated BoW/FV/OFF/ODF/SDF/GSF/AF representations $\{\vec{\tilde{\psi}}_i, i\!\in\!\mathcal{H}\}$.
The HAF stream is denoted by $\vpsi_{(haf)}$, which is generated by $\hslash(\cdot, \mP_{(haf)})$. 

The parameters $\{\mP_{i}, i\!\in\!\mathcal{H}\}$ are learned for the hallucination streams, while $\mP_{{(haf)}}$ is learned for the HAF stream. The complete set of parameters is denoted as $\bar{\mP}\!\equiv\!(\{{\mP_{i}, i\!\in\!\mathcal{H}}\}, \mP_{(haf)}, \mP_{(pr)},\mP_{(cov)}, \mP_{(\ell)})$. 
Furthermore, the projection matrices $\{\mPP_{i}, i\!\in\!\mathcal{H}\}$ are used for count sketching of the ground-truth BoW/FV/OFF/ODF/SDF/GSF/AF feature vectors $\{\vpsi_{i}, i\!\in\!\mathcal{H}\}$, and the corresponding sketched/compressed representations are $\{\vpsi'_i, i\!\in\!\mathcal{H}\}$. The projection matrix $\mPP_{(tot)}$ handles the concatenation of the hallucinated BoW/FV/OFF/ODF/SDF/GSF/AF representations with HAF. This results in $\vpsi_{(tot)}\!=\!\left[\tilde{\vpsi}; \vpsi_{(haf)}\right]$, and its sketched counterpart $\vpsi'_{(tot)}$ is fed into the PredNet module $f$. Section \ref{sec:sketch} explains how to select the matrices $\mPP$. 
If sketching is not needed, we simply set $\mPP$ to be the identity matrix, $\mPP\!=\!\mIdent$. In our experiments, we set $\alpha\!=\!1$.

We also explore a variant in which we use a weighted average of several streams fed into the PredNet module $f$:
\begin{align}
&\qquad\;\vpsi'_{(tot)}\!=\!\frac{1}{|\mathcal{H^*}\!|\!+\!1}\Big(w_{(haf)}\vpsi_{(haf)}\!\!+\!\!\!\sum_{i\in\mathcal{H^*}}\!w_i\vec{\tilde{\psi}}_i\Big),\nonumber\\
&\qquad\;\tilde{\vpsi}_{(det)}\!=\!\frac{1}{|\mathcal{D}|}\!\sum_{i\in\mathcal{D}}\!w_i\vec{\tilde{\psi}}_i, \text{ and }%
\tilde{\vpsi}_{(sal)}\!=\!\frac{1}{|\mathcal{S}|}\!\sum_{i\in\mathcal{S}}\!w_i\vec{\tilde{\psi}}_i,
\label{eq:odf-sdf-wei}    
\end{align}
where $\mathcal{H^*}\!\equiv\!\left\{(fv1),(fv2),(bow),(off),(odf),(sdf)\right\}$, $\mathcal{D}\!\equiv\!\left\{(det1),\cdots,(det4)\right\}$, and $\mathcal{S}\!\equiv\!\left\{(sal1),(sal2)\right\}$.
Let $\mathcal{T}$ be set to $\mathcal{H^*}$, $\mathcal{D}$ or $\mathcal{S}$, and the weights are defined as:
\vspace{-0.2cm}
\begin{align}
& w_{i}\!=\!\frac{1}{|\mathcal{T}|} \frac{\max(w'^{\beta}_{i}\!\!\!,\rho)}{\sum_{j\in\mathcal{T}}\!\max(w'^{\beta}_j\!\!\!,\rho)}.
\label{eq:wei}
\end{align}

\vspace{0.05cm}
\noindent{\textbf{Optimization.}} Before training the CNN, we first train an SVM for each ground-truth stream separately (using a manageable subset of the data). The weights $w'$ are set proportionally to the accuracies obtained on the validation set. 
For the HAF stream, we set $w'_{(haf)}\!=\!\frac{1}{|\mathcal{H^*}\!|\!+\!1}$ and $\rho\!=\!0.1$. 
In the first few epochs (\ie, 10), we set $\beta\!=\!0$, ensuring that all streams receive equal weights. Subsequently, we perform a Golden-section search to determine the optimal $\beta\!\geq\!0$ in each epoch. 
We start with boundary values $\beta\!\in\!\{0,50\}$, train an SVM on a manageable subset of training data, evaluate $\beta$ on the validation set,  
and update the boundary values for the next epoch.

Eq. \eqref{eq:wei} has an interesting property: for $\beta\!=\!0$, all weights are equal, $w_i\!=\!1/|\mathcal{T}|$. As $\beta\!\rightarrow\!\infty$, the weights become binary: $w_i\!=\!1$ if $w_i\!=\!\max(\{w_i\}_{i\in\mathcal{T}})$, and $w_i\!=\!0$ otherwise. Thus, $\beta$ interpolates between equal weighting and a winner-takes-all approach.

 We minimize $\ell^*(\tX, \vy; \bar{\mP})$ with respect to the parameters of each stream: $\{\mP_{i}, i\!\in\!\mathcal{H}\}$ for hallucination streams, $\mP_{(haf)}$ for the HAF stream, $\mP_{(cov)}$ for CENet, $\mP_{(pr)}$ for PredNet, and $\mP_{(\ell)}$ for the classification loss. 
 In practice, we alternate between two minimization steps:  one forward and backward pass to update the parameters $\{\mP_{i}, i\!\in\!\mathcal{H}\}$ and $\mP_{(cov)}$ for uncertainty learning, followed by another forward and backward pass for the classification loss $\ell$. 
 This can be viewed as a multi-task learning process, where we simultaneously learn BoW/FV/OFF/ODF/SDF/GSF/AF and label tasks. We use the Adam minimizer with an initial learning rate of $10^{-4}$, halved every 10 epochs, and train for 50--100 epochs, depending on the dataset.

\end{sloppypar}

\section{Experiment}
\label{sec:exper}
\begin{sloppypar}
Below, we demonstrate the effectiveness of our method. For smaller datasets, such as HMDB-51 and YUP++, we use the I3D, VTN, and FASTER backbones. For Charades, EPIC-KITCHENS-55, and Toyota Smarthome, we also investigate the AssembleNet++ backbones. For large-scale Kinetics-400, Kinetics-600, and Something-Something V2, we use the recent, popular VideoMAE V2 and InternVideo2.

\subsection{Datasets and Evaluation Protocols}
\label{sec:data}

\noindent\textbf{HMDB-51}~\cite{kuehne2011hmdb} consists of 6766 internet videos across 51 classes, with each video containing approximately 20 to 1000 frames. We report the mean accuracy across three splits.

\noindent\textbf{YUP++}~\cite{yuppp} contains 20 scene classes of video textures, with 60 videos per class. The splits include scenes captured by either static or moving cameras. We use the standard splits (1/9 of the dataset for training) for evaluation.

\noindent\textbf{MPII Cooking Activities}~\cite{rohrbach2012database} includes high-resolution videos of people cooking various dishes. 
The 64 activities span 3748 clips, including coarse actions such as \emph{opening refrigerator}, and fine-grained actions like \emph{peel}, \emph{slice}, and \emph{cut apart}. We report the mean Average Precision (mAP) over 7-fold cross validation. For the human-centric protocol \cite{anoop_generalized, anoop_rankpool_nonlin}, we use Faster R-CNN \cite{faster-rcnn} to crop the video around human subjects.

\noindent\textbf{Charades}~\cite{sigurdsson2016hollywood} consists of of 9848 videos of daily indoor activities, 66,500 clip annotations, and 157 classes.

\noindent\textbf{EPIC-KITCHENS-55}~\cite{Damen_2018_ECCV} is a multi-class, egocentric dataset with approximately 28K training videos associated with 331 noun and 125 verb classes. The dataset contains of 39,594 segments across 432 videos.  
We follow the protocol in~\cite{Baradel_2018_ECCV} and evaluate our model on the validation set, as well as the standard seen (S1: 8,047 videos), and unseen (S2: 2,929 videos) test sets.

\noindent\textbf{Toyota Smarthome}~\cite{Das_2019_ICCV} consists of 16,115 RGB+D video clips spanning 31 activity classes. This dataset poses several challenges, such as high intra-class variation, high class imbalance, simple and composite activities, and activities with similar motion and variable duration. Activities are annotated with both coarse and fine-grained labels. There are two evaluation protocols for activity classification: cross-subject and cross-view. Follow~\cite{Das_2019_ICCV}, we report the mean per-class accuracy.

\noindent\textbf{Kinetics-400}~\cite{kay2017kinetics} contains 400 human action classes with over 300K video clips, each containing 180 frames. The dataset covers a wide range of actions, including sports, everyday tasks, and human-object interactions, and is split into training, validation, and test sets.

\noindent\textbf{Kinetics-600}~\cite{carreira2018short} extends the Kinetics-400 dataset, with over 500K video clips. It includes a more diverse set of actions compared to Kinetics-400, covering a broad spectrum of human activities from both indoor and outdoor environments.

\noindent\textbf{Something-Something V2}~\cite{goyal2017something} is a large-scale action recognition dataset consisting of 220K videos across 174 action classes. The dataset focuses on human-object interactions, where actions are typically described by phrases like \textit{putting something in something} or \textit{picking something up}.

\subsection{Evaluations of Various Design Components}
\label{sec:evals}
\vspace{0.05cm}
\noindent{\textbf{Sketching and Power Normalization.}} Since PredNet uses a fully connected (FC) layer (see Figure \ref{fig:strc}), we expect that limiting the input size to this layer through count sketching, as described in Section \ref{sec:sketch}, should improve performance. Additionally, given that visual and video representations often suffer from burstiness, we investigate the AsinhE, SigmE, and AxMin methods, as outlined in Remarks \ref{re:asinhe}, \ref{re:sigme} and \ref{re:axmin}.

\begin{figure}[t]%htbp % left bottom right top
% \vspace{-0.5cm}
% \hspace{-0.3cm}
\centering%%%%
\begin{subfigure}[b]{0.495\linewidth}
\centering\includegraphics[trim=0 0 0 0, clip=true,width=\linewidth]{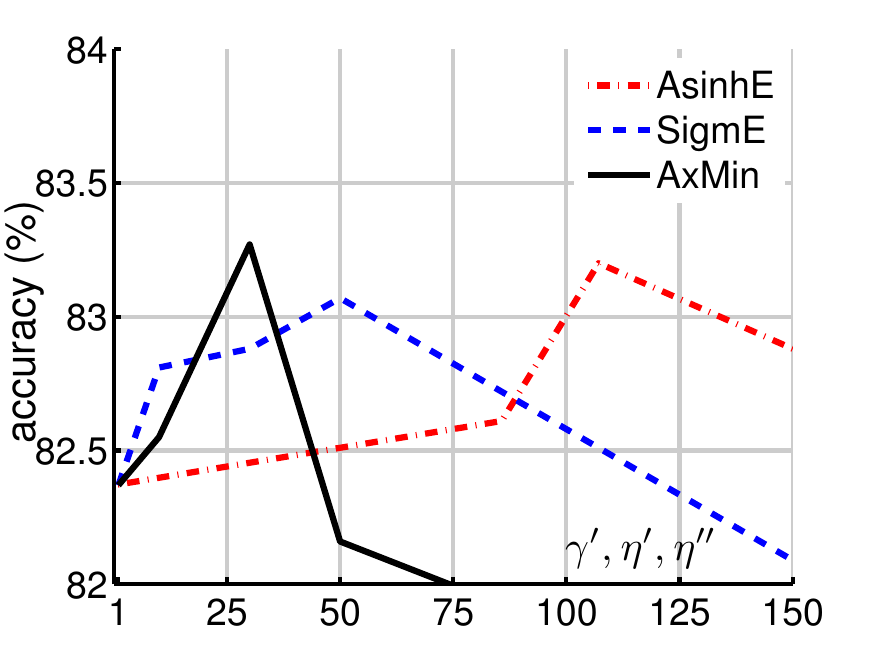}
\vspace{-0.6cm}
\caption{\label{fig:pn}}
\vspace{-0.2cm}
\end{subfigure}
%
% \hspace{0.1cm}
\begin{subfigure}[b]{0.495\linewidth}
\centering\includegraphics[trim=0 0 0 0, clip=true,width=\linewidth]{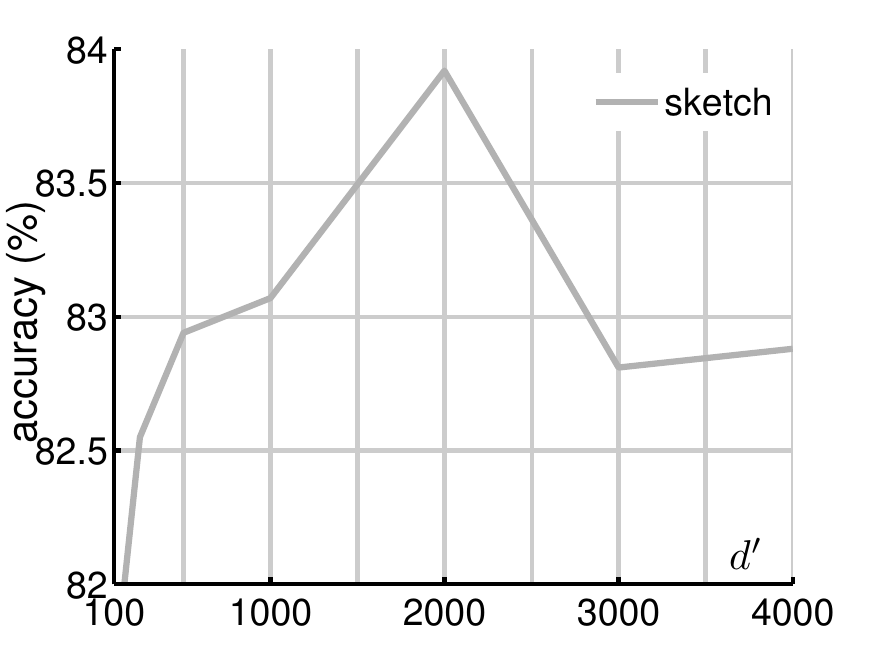}
\vspace{-0.6cm}
\caption{\label{fig:sketch}}
\vspace{-0.2cm}
\end{subfigure}
%
%\vspace{-0.7cm}
\caption{Evaluations of (fig. \ref{fig:pn}) Power Normalization and (fig. \ref{fig:sketch}) sketching on the HMDB-51 dataset (split 1 only). 
}\vspace{-0.3cm}
\label{fig:pnsk}
\end{figure}

Figure \ref{fig:pn} investigates the classification accuracy on the HMDB-51 dataset (split 1) when our HAF and BoW/FV feature vectors $\{\vec{\tilde{\psi}}_i, i\!\in\!\mathcal{H}\}$ and $\vpsi_{(haf)}$ (described in Sections \ref{sec:hal-stream} and \ref{sec:obj}) are processed through Power Normalizing functions: AsinhE, SigmE and AxMin, prior to concatenation (see Figure \ref{fig:pipe}). Our experiments show that all PN functions perform similarly, improving results from a baseline $82.29$\% to approximately $83.20$\% accuracy. A similar improvement is observed on YUP++ ({\em static}), with accuracy rising from $93.15$\% to $94.44$\%. For simplicity, we use AsinhE for PN in the following experiments.

Figure \ref{fig:sketch} illustrates the effect of applying count sketching to the concatenated HAF and BoW/FV feature vectors $\vpsi_{(tot)}$, resulting in $\vpsi'_{(tot)}$ (see Section \ref{sec:obj} for reference to symbols), on the HMDB-51 dataset (split 1). This approach improves the accuracy from $82.88$\% to $83.92$\% for $d'\!=\!2000$. This improvement is expected, as reduced size of $\vpsi'_{(tot)}$ results in fewer parameters for the FC layer in PredNet, reducing overfitting. Similarly,  on the YUP++ dataset (split {\em static}), the accuracy increases from $93.15$\% to $94.81$\%.

\vspace{0.05cm}
\noindent{\textbf{ODF+SVM.}} Firstly, we evaluate our ODF on SVM using the HMDB-51 dataset. We set $n'\!=3$ for Eq. \eqref{eq:moment1} and compare various detector backbones and pooling strategies. Table \ref{tab:det1234} shows that all detectors perform similarly, with ({\em det3}) slightly outperforming the others. Additionally, max-pooling on ODFs from all four detectors marginally outperforms average-pooling. However, only the weighted mean ({\em all+wei}), as defined in Eq. \eqref{eq:wei}, outperforms ({\em det3}), highlighting the importance of robust aggregation of ODFs. Similarly, when combining pre-trained DEEP-HAL with all detectors, the weighted mean ({\em DEEP-HAL+all+wei}) achieves the best performance. Table \ref{tab:yup-pool} shows a similar trend on YUP++.

\begin{figure}[t]%htbp % left bottom right top
%\hspace{0.5cm}%
\centering%%%%
% \vspace{-0.1cm}
%
\begin{subfigure}[b]{0.495\linewidth}
\includegraphics[trim=0 0 0 0, clip=true,width=0.99\linewidth]{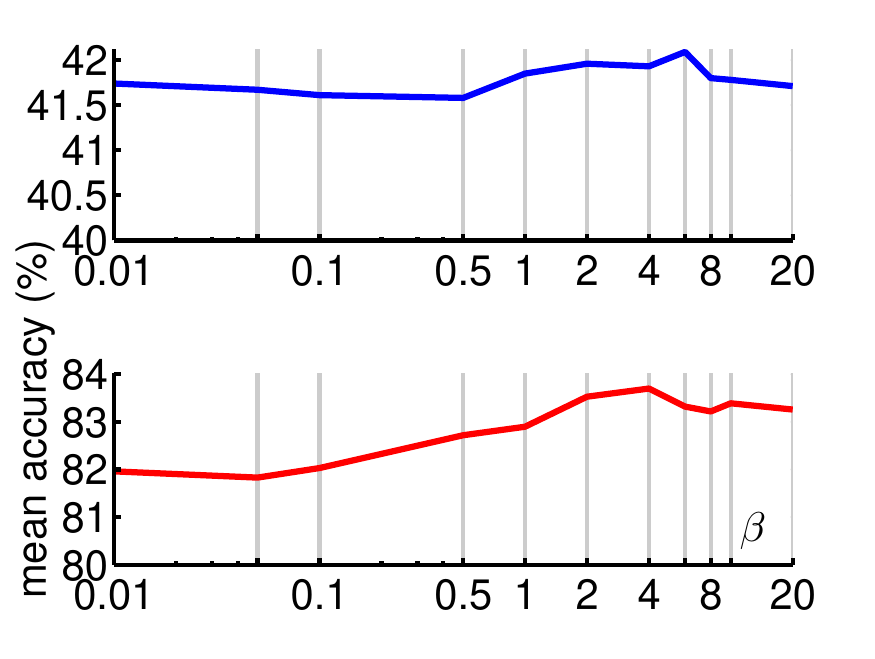}\vspace{-0.2cm}
\caption{\label{fig:beta-h}}
\vspace{-0.2cm}
\end{subfigure}
\begin{subfigure}[b]{0.495\linewidth}
\includegraphics[trim=0 0 0 0, clip=true,width=0.99\linewidth]{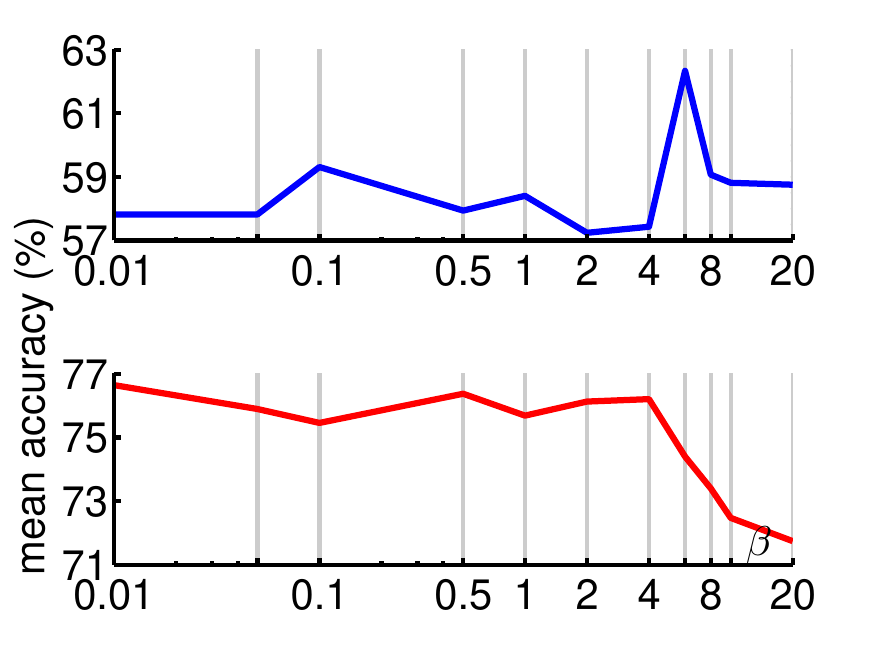}\vspace{-0.2cm}
\caption{\label{fig:beta-y}}
\vspace{-0.2cm}
\end{subfigure}
\caption{Impact of $\beta$ in the weighted mean on classification performance. Figure \ref{fig:beta-h} presents results for HMDB-51 with ({\em top}) four combined detectors + SVM and ({\em bottom}) DEEP-HAL with four combined detectors + SVM. Figure \ref{fig:beta-y} shows results for YUP++.}
\vspace{-0.3cm}
\label{fig:beta}
\end{figure}

We train SVM only on videos where at least one detection occurred, so the resulting accuracy of $75.74\%$ is lower than the main results reported on the full pipeline. 
Finally, Figure \ref{fig:beta} demonstrates that  $\beta\!\neq\!1$ has a positive impact on reweighting.

\begin{table}[t]%htbp % left bottom right top
% \vspace{-0.3cm}
% \setlength{\tabcolsep}{0.12em}
% \renewcommand{\arraystretch}{0.70}
%\fontsize{9}{9}\selectfont
\centering
\resizebox{0.9\linewidth}{!}{\begin{tabular}{ l c c c c }
\toprule
 & {\em sp1} & {\em sp2} & {\em sp3} & mean acc. \\
\hline
%\hdashline
{\em det1} & $42.00$ & $39.74$ & $40.39$ & $40.72$\\
{\em det1} & $40.49$ & $40.13$ & $39.67$ & $40.09$\\
{\em det3} & $43.78$ & $44.05$ & $41.97$ & $\mathbf{43.26}$\\
{\em det4} & $41.08$ & $39.22$ & $40.39$ & $40.23$\\
% \hline
\hdashline
{\em all+avg} & $42.50$ & $41.05$ & $41.01$ & $41.52$\\
{\em all+max} & $43.25$ & $42.32$ & $42.09$ & $42.55$\\
{\em all+wei} & $45.80$ & $44.52$ & $44.09$ & $\mathbf{44.80}$\\
% \hline
\hdashline
{\em DEEP-HAL+all+avg} & $83.25$ & $82.24$ & $82.84$ & $82.77$\\
{\em DEEP-HAL+all+max} & $83.18$ & $81.86$ & $82.84$ & $82.62$\\
{\em DEEP-HAL+all+wei} & $84.01$ & $83.25$ & $83.10$ & $\mathbf{83.45}$\\
% \hline
\bottomrule
\end{tabular}
%\vspace{0.02cm}
}% }
\caption{Evaluations of ODF on HMDB-51. (Top) Performance is evaluated using backbones: ({\em det1}) Inception V2, ({\em det2}) Inception ResNet V2, ({\em det3}) ResNet101, and ({\em det4}) NASNet. (Middle) Results for average pooling, max pooling, and weighted mean combinations of all detectors are reported as ({\em all+avg}), ({\em all+max}), and ({\em all+wei}), respectively. (Bottom) Pre-trained DEEP-HAL combined with all four detectors is evaluated using average pooling, max pooling, and weighted mean.
}
\vspace{-0.3cm}
\label{tab:det1234}
\end{table}
\begin{table}[t]%htbp % left bottom right top
%\vspace{-0.1cm}
% \vspace{-0.3cm}
% \setlength{\tabcolsep}{0.12em}
% \renewcommand{\arraystretch}{0.70}
%\fontsize{9}{9}\selectfont
\centering
\resizebox{0.65\linewidth}{!}{\begin{tabular}{ l c c c }
\toprule
 & {\em avg} & {\em max} & {\em wei}  \\
\hline
{\em all}          & $55.12$ & $42.34$ & $\mathbf{60.52}$ \\
{\em DEEP-HAL+all} & $74.22$ & $71.85$ & $\mathbf{75.74}$ \\
% \hline
\bottomrule
\end{tabular}
%\vspace{0.02cm}
}
\caption{Pooling on YUP++. Results are shown for average pooling ({\em avg}), max pooling ({\em max}), and weighted mean ({\em wei}) of all detectors ({\em all}) compared to pre-trained DEEP-HAL combined with all detectors using average pooling, max pooling, and weighted mean.
}
\vspace{-0.5cm}
\label{tab:yup-pool}
\end{table}

\vspace{0.05cm}
\noindent{\textbf{SDF.}} The SDF achieves accuracies of $24.35\%$ on HMDB-51 and $32.68\%$ on YUP++. This is expected, as SDF does not capture discriminative information directly, but instead identify salient spatial and temporal regions to focus the main network's attention on.

\begin{figure}[t]%htbp % left bottom right top
%\hspace{0.5cm}%
\centering%%%%
% \vspace{-0.1cm}
%
\begin{subfigure}[b]{0.495\linewidth}
\includegraphics[trim=0 0 0 0, clip=true,width=0.99\linewidth]{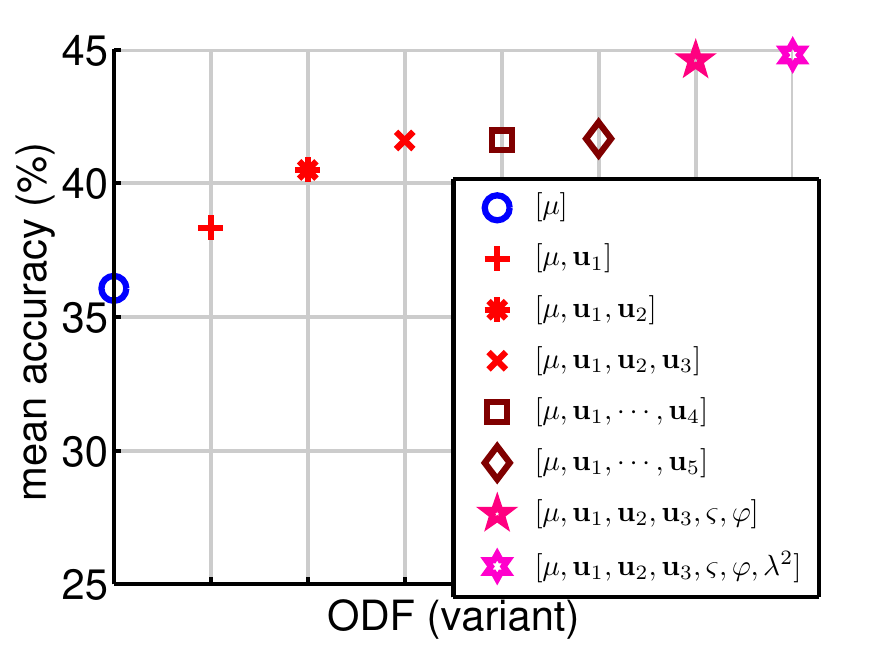}\vspace{-0.2cm}
\caption{\label{fig:odf-h}}
\vspace{-0.2cm}
\end{subfigure}
\begin{subfigure}[b]{0.495\linewidth}
\includegraphics[trim=0 0 0 0, clip=true,width=0.99\linewidth]{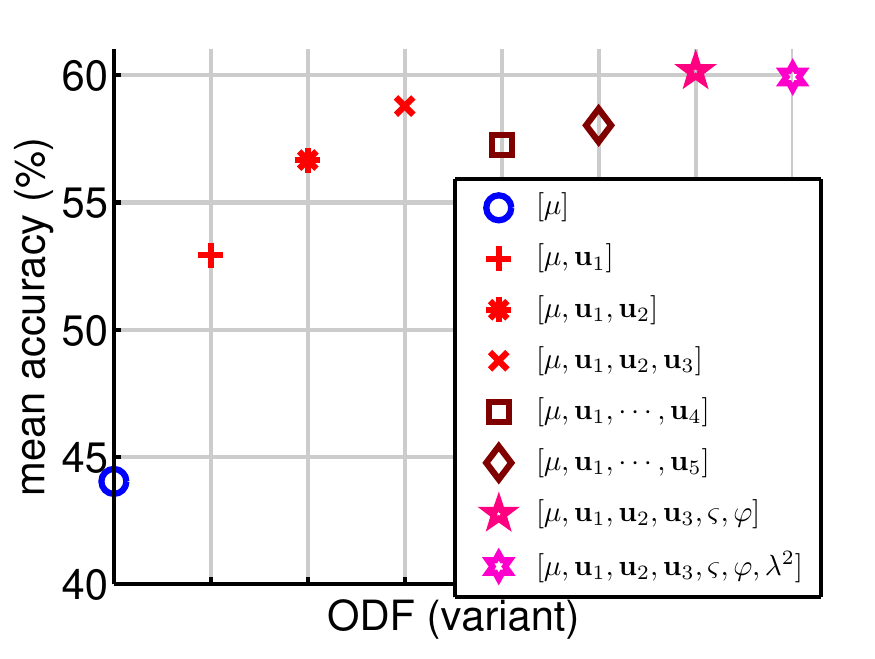}\vspace{-0.2cm}
\caption{\label{fig:odf-y}}
\vspace{-0.2cm}
\end{subfigure}
\caption{Evaluation of ODF with SVM using the weighted mean across four detectors. Figure \ref{fig:odf-h} and Figure \ref{fig:odf-y} show results on HMDB-51 and YUP++, respectively. The parameters $\vmu$, $\vu_1, \ldots, \vu_i$, $\boldsymbol{\varsigma}$, $\boldsymbol{\varphi}$, and $\boldsymbol{\lambda}^2$ correspond to the entries in Eq. \eqref{eq:moment1}.
}
%\vspace{-0.3cm}
\label{fig:odf}
\end{figure}

\vspace{0.05cm}
\noindent{\textbf{Multi-moment descriptor.}} Figure \ref{fig:odf} shows that concatenating the mean and three eigenvectors, as defined in Eq. \eqref{eq:moment1}, yields good results. However, adding additional vectors leads to a deterioration in performance. Adding skewness and kurtosis ($\boldsymbol{\varsigma}$ and $\boldsymbol{\varphi}$) further improves the results, while the inclusion of eigenvalues has a limited impact.

\noindent{\textbf{ImageNet (global score) \vs object detectors.}} 
Various scores from the object and saliency detectors we use cannot be directly integrated into DEEP-HAL due to the varying number of detected objects and frames. Therefore, we propose and use ODF and SDF descriptors. Additionally, We note that using a simplified version of ODF, which stacks ImageNet-1K scores per frame into a matrix (without detectors) and applies our multi-moment descriptor, yields approximately $4\%$ worse results on Charades compared to our DEEP-HAL+ODF (detectors-based approach), which achieves $48.0\%$ mAP. This is expected, as ImageNet-1K is trained in a multi-class setting (one object per image), while detectors allow us to robustly model the distribution of object classes and locations per frame.

\vspace{0.05cm}
\noindent{\textbf{Reweighting mechanism.}}
In this experiment, we employ the DEEP-HAL pipeline and hallucinate ODF and SDF ($d\!=\!512$). Typically, we use three levels of weighting mean pooling, applied to: (i) four object detectors constituting the ODF, (ii) two saliency detectors constituting the SDF, and (iii) the final combination of HAF/BoW/FV/OFF/ODF/SDF. Below, we investigate the performance of a single weighted mean pooling step, applied simultaneously to four object detectors, two saliency detectors, and the remaining streams. 

Table \ref{tab:weilevel} shows that using a flat, single-level weighted mean pooling yields 86.1\% accuracy on the HMDB-51, which is approximately $1.4\%$ lower than using  three levels of weighted mean pooling. 
We also observe that the improvement from using {\em wei+3 levels} is very minimal (around 0.2\%). We expect that applying a single weighted mean pooling per modality is a reasonable strategy, as object category detectors, for example, may yield similar responses. Therefore, they should first be reweighted for the best `combined detector' performance before being combined with more complementary modalities. 

\begin{figure}[t]%htbp % left bottom right top
\hspace{-0.2cm}%
\centering%%%%
% \vspace{-0.3cm}
%
\includegraphics[trim=0 0 0 0, clip=true,width=0.7\linewidth]{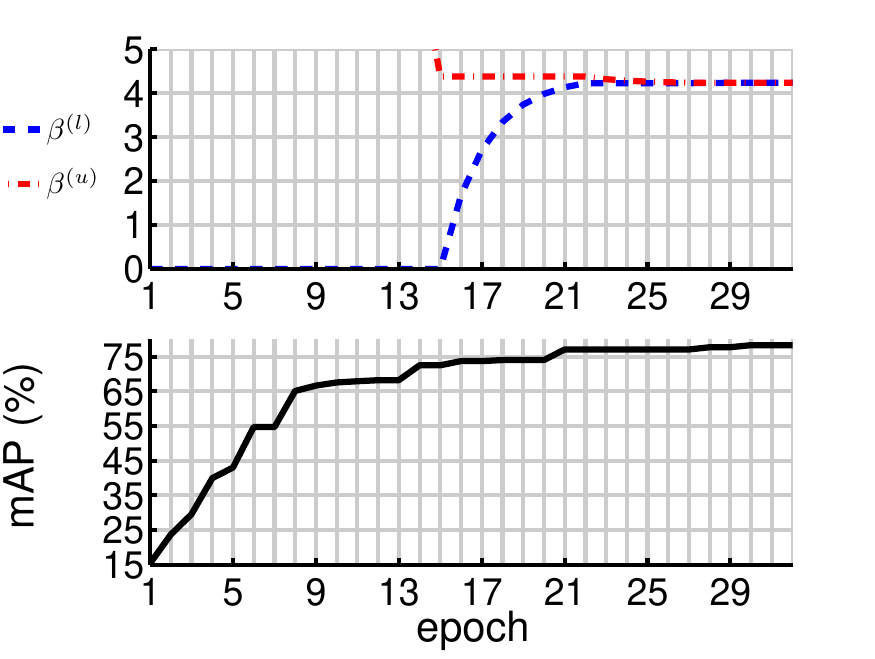}%\vspace{-0.2cm}
\caption{Visualization of the Golden-section search for the weighting mechanism at the final level. ({\em top}) Depicts the convergence of the lower and upper estimates, $\beta^{(l)}$ and $\beta^{(u)}$, over the course of training epochs. ({\em bottom}) For each epoch, $\beta$ is set as the midpoint $\beta\!=\!0.5(\beta^{(l)}\!+\!\beta^{(u)})$, and the corresponding validation score (mAP) is evaluated on MPII ({\em split1}). As training progresses, mAP steadily improves and stabilizes, reflecting the convergence of the Golden-section search algorithm.
}
\vspace{-0.3cm}
\label{fig:gold}
\end{figure}

Finally, Figure \ref{fig:gold} (top) demonstrates how our Golden-search selects the optimal $\beta$ on the MPII validation set ({\em split1}).  Figure \ref{fig:gold} (bottom) shows the corresponding validation mAP (note that this is not the mAP score on the test set). For the first 10 epochs we use $\beta\!=\!0$, and we start the Golden-search from epoch 11.

\begin{table}[t]%htbp % left bottom right top
% \vspace{0.3cm}
% \setlength{\tabcolsep}{0.12em}
% \renewcommand{\arraystretch}{0.70}
%\fontsize{9}{9}\selectfont
\centering
\resizebox{0.8\linewidth}{!}{\begin{tabular}{ l c c c c }
\toprule
 & {\em sp1} & {\em sp2} & {\em sp3} & mean acc. \\
\hline
{\em wei+flat} & $86.47$ & $85.56$ & $86.27$ & $86.10$\\
{\em wei+1 level} & $88.00$ & $86.33$ & $87.20$ & $87.18$\\
{\em wei+2 levels} & $88.20$ & $86.50$ & $87.33$ & $87.34$\\
{\em wei+3 levels} & $88.37$ & $86.80$ & $87.52$ & $\mathbf{87.56}$\\
% \hline
\bottomrule
\end{tabular}
%\vspace{0.02cm}
}
\caption{Evaluation of the flat single-level weighted mean ({\em wei+flat}) versus different hierarchical levels of weighted mean pooling on HMDB-51. {\em wei+1 level} indicates that weighted mean pooling is applied solely to the four object detectors in ODF. {\em wei+2 levels} extends this by applying weighted pooling to both ODF and SDF. {\em wei+3 levels} adds a final weighted combination of HAF, BoW, FV, OFF, ODF, and SDF.
}
\vspace{-0.3cm}
\label{tab:weilevel}
\end{table}

% \vspace{0.05cm}
% \noindent{\textbf{Hallucination loss.}} We now evaluate the performance of using MSE and uncertainty learning for the feature hallucination. We notice that the use of uncertainty learning help improve the AR performance on all datasets. On HMDB-51 and YUP++ (Table~\ref{tab:hmdb51f} and~\ref{tab:yupf}), uncertainty loss improves the performance by $\sim$ 1\% for all backbones on average. On MPII (Table~\ref{tab:mpiif}), hallucinating additional ODF/SDF with uncertainty outperforms the MSE loss by 0.8\% with I3D backbone, and hallucinating all features with uncertainty learning further improves the performance by 0.5\%.

% Videos generally contain background sounds that are irrelevant to the observed actions captured by wearable sensors. For example, in the recorded videos for EPIC-KITCHENS, there are sounds such as music/TV playing, ongoing coffee machine and/or washing machine in the background when actions take place. In Table~\ref{epic-kitchens}, the additional audio features boost the AR by more than 3\% on all backbones. The biggest performance gain ($\sim$ 5\%) is when AssembleNet++ backbone is used. Our experimental results show that these irrelevant sounds are not the source of confusion, which demonstrates the robustness of our model to noisy and unconstrained audio sources.

\vspace{0.05cm}
\noindent{\textbf{Input to CENet.}} The input to our CENet can either be the intermediate representation $\mathcal{X}_{(rgb)}$ from the backbone or the concatenated feature representations $\tilde{\vpsi}$ from hallucination (see Sec.~\ref{sec:cenet}), we compare the performance of these two variants on Charades. 
As shown in Table~\ref{tab:charades}, using concatenated feature representations as input to CENet improves the performance of AR by more than 0.5\% compared to the use of intermediate representation. Both variants use the same experimental settings, \ie, the same number of multiple sketches (MSK) and the same feature dimensionality $d$ through sketching, across all four backbones. The reasons for this improvement are: (i) the intermediate representations are `raw' and not sufficiently representative for CENet to identify the relationships between each feature descriptor and its corresponding uncertainty, and (ii) the hallucinated features are more informative than the intermediate representations. 
Therefore, we use the hallucinated feature representations as input to CENet (see Fig.~\ref{fig:pipe}).

\vspace{0.05cm}
\noindent{\textbf{Sketched feature dimension per modality.}} We evaluate the effects of feature dimension $d$ through sketching on both Charades and EPIC-KITCHENS-55. On Charades, $d\!=\!128$ achieves almost identical performance to $d\!=\!256$ (within 0.3\% difference, see Table~\ref{tab:charades}) across all backbones. Considering both the performance gain and the computational cost, we choose $d\!=\!128$ for subsequent experiments on Charades. 
On the validation set of EPIC-KITCHENS-55, with AssembleNet++ and VTN backbones, $d\!=\!128$ yields very similar results to $d\!=\!256$. However, with I3D and FASTER backbones, $d\!=\!128$ decreases performance by more than 0.5\%. Thus, for the seen and unseen test protocols, we choose $d\!=\!128$ for AssembleNet++ and VTN backbones and $d\!=\!256$ for I3D and FASTER backbones to achieve better performance.

\vspace{0.05cm}
\noindent{\textbf{Multiple sketches.}} On Charades, we observe that using more multiple sketches (MSK) helps improve AR performance. Specifically, 4$\times$ MSK outperforms 2$\times$ MSK by 0.2\%, and 8$\times$ MSK outperforms 4$\times$ MSK by 0.9\%. However, the improvement becomes smaller when 12$\times$ MSK is used compared to 8$\times$ MSK across all backbones (approximately 0.1\% improvement, see Table~\ref{tab:charades}), except the FASTER framework. Considering both the performance gain and computational cost, we set the number of MSK to 8 for the experiments on EPIC-KITCHENS-55 and Toyota Smarthome.

\begin{table}[t]%htbp % left bottom right top
%\vspace{-0.3cm}
\parbox{.99\linewidth}{
% \setlength{\tabcolsep}{0.12em}
% \renewcommand{\arraystretch}{0.70}
% \fontsize{9}{9}\selectfont
\centering
\resizebox{0.48\textwidth}{!}{\begin{tabular}{ l c c c c c c c}
\toprule
\multirow{2}{*}{Backbone} & \multicolumn{5}{c}{Modality}  &  {Hal.}  & Mean\\
\cline{2-6}
& \tiny{BoW/FV/OFF} & ODF & SDF  & GSF & AF  & loss &  acc.(\%)\\
\midrule
\multirow{10}{*}{\parbox{1.0cm}{I3D}}
&  & & & & & - &  $74.8$\\
& \checkmark & & & & & MSE &  $83.3$\\
& \checkmark & \checkmark & & & & MSE  &  $84.3$\\
& \checkmark &  & \checkmark & & & MSE  &  $83.9$\\
& \checkmark & \checkmark & \checkmark & & & MSE  &  $85.2$\\
& \checkmark& \checkmark & \checkmark & & & MSE &   $87.0$\\
& \checkmark& \checkmark & \checkmark & & & MSE &   $87.6$\\
& \checkmark& \checkmark & \checkmark & \checkmark & & MSE &   $88.0$\\
& \checkmark& \checkmark & \checkmark & \checkmark & \checkmark & MSE  &  $88.2$\\
& \checkmark& \checkmark & \checkmark & \checkmark & \checkmark & Uncert.  &  $\mathbf{88.6}$\\
\hline
\multirow{7}{*}{VTN-MobileNet}
&  & & & & & - & 59.2\\
& \checkmark & & & & & MSE & 67.3\\
& \checkmark & & & & & Uncert. & 68.1\\
& \checkmark & \checkmark & \checkmark & & & Uncert. & 70.9 \\
& \checkmark & \checkmark & \checkmark & \checkmark& & Uncert. & 72.2\\
& \checkmark & \checkmark & \checkmark & \checkmark& \checkmark& MSE & 71.9\\
& \checkmark & \checkmark & \checkmark & \checkmark& \checkmark& Uncert. & {\bf 72.5}\\
\hdashline
\multirow{7}{*}{VTN-ResNet}
&  & & & & & - & 63.1\\
& \checkmark & & & & & MSE & 73.3\\
& \checkmark & & & & & Uncert. & 75.2\\
& \checkmark & \checkmark & \checkmark & & & Uncert. & 77.0\\
& \checkmark & \checkmark & \checkmark & \checkmark& & Uncert. & 78.3\\
& \checkmark & \checkmark & \checkmark & \checkmark& \checkmark & MSE & 77.9\\
& \checkmark & \checkmark & \checkmark & \checkmark& \checkmark & Uncert. & {\bf 78.6}\\
\hline
\multirow{7}{*}{FASTER}
&  &  &  & & & - & 74.9\\
& \checkmark &  &  & & & MSE & 80.3\\
& \checkmark &  &  & & & Uncert. & 81.9\\
& \checkmark & \checkmark & \checkmark & & & Uncert. & 83.7\\
& \checkmark & \checkmark & \checkmark & \checkmark & & Uncert. & 85.8\\
& \checkmark & \checkmark & \checkmark & \checkmark & \checkmark & MSE & 85.6\\
& \checkmark & \checkmark & \checkmark & \checkmark & \checkmark & Uncert. & {\bf 86.2}\\
% \hline
\midrule
\end{tabular}
%\vspace{0.02cm}\\
}}
\parbox{.99\linewidth}{
% \setlength{\tabcolsep}{0.12em}
% \renewcommand{\arraystretch}{0.70}
% \fontsize{9}{9}\selectfont
\centering
\resizebox{0.48\textwidth}{!}{\begin{tabular}{ c c }
\kern-0.5em ADL+I3D $81.5\%$ \cite{anoop_advers} &  Full-FT I3D $81.3\%$ \cite{i3d_net}\\
\kern-0.5em EvaNet (Ensemble) $82.3\%$ \cite{Piergiovanni_2019_ICCV} & PA3D + I3D $82.1\%$ \cite{Yan_2019_CVPR}\\
 DEEP-HAL (exact) $82.50\%$ \cite{Wang_2019_ICCV} & DEEP-HAL $82.48\%$ \cite{Wang_2019_ICCV}\\
% \hline
\bottomrule
\end{tabular}
}}
\caption{Evaluation results for ({\em top}) our proposed methods and ({\em bottom}) comparisons with state-of-the-art approaches on HMDB-51.}
%\vspace{-0.2cm}
\vspace{-0.3cm}
\label{tab:hmdb51f}
\end{table}

% \begin{table}[t]%htbp % left bottom right top
% \parbox{.99\linewidth}{
% \setlength{\tabcolsep}{0.12em}
% \renewcommand{\arraystretch}{0.70}
% \centering
% \resizebox{0.48\textwidth}{!}{\begin{tabular}{ l c c c c }
% \toprule
%  & {\em sp1} & {\em sp2} & {\em sp3} & mean acc. \\
% \hline
% {\em DEEP-HAL+W} & $83.94\%$ & $82.50\%$ & $83.34\%$ & $83.26\%$\\
% {\em DEEP-HAL+ODF} & $85.03\%$ & $83.59\%$ & $84.25\%$ & $84.29\%$\\
% {\em DEEP-HAL+SDF} & $84.64\%$ & $83.20\%$ & $83.82\%$ & $83.88\%$\\
% {\em DEEP-HAL+ODF+SDF} & $86.14\%$ & $83.66\%$ & $85.81\%$ & $85.20\%$\\
% {\em DEEP-HAL+W+ODF+SDF} & $87.78\%$ & $86.27\%$ & $87.06\%$ & $87.04\%$\\
% {\em DEEP-HAL+W+G+ODF+SDF}  & $88.37\%$ & $86.80\%$ & $87.52\%$ & $\mathbf{87.56}\%$\\
% % \hline
% \midrule
% \end{tabular}
% %\vspace{0.02cm}\\
% }}
% %
% \parbox{.99\linewidth}{
% \setlength{\tabcolsep}{0.12em}
% \renewcommand{\arraystretch}{0.70}
% \fontsize{9}{9}\selectfont
% \centering
% \resizebox{0.48\textwidth}{!}{\begin{tabular}{ c c }
% \kern-0.5em ADL+I3D $81.5\%$ \cite{anoop_advers} &  Full-FT I3D $81.3\%$ \cite{i3d_net}\\
% \kern-0.5em EvaNet (Ensemble) $82.3\%$ \cite{Piergiovanni_2019_ICCV} & PA3D + I3D $82.1\%$ \cite{Yan_2019_CVPR}\\
%  HAF/BoW/FV exact $82.50\%$ \cite{Wang_2019_ICCV} & HAF/BoW/FV hal. $82.48\%$ \cite{Wang_2019_ICCV}\\
% % \hline
% \bottomrule
% \end{tabular}
% }}
% \caption{Evaluations of ({\em top}) our methods and ({\em bottom}) comparisons to the state of the art on  HMDB-51.}
% \vspace{-0.3cm}
% \label{tab:hmdb51f}
% \end{table}

\begin{table}[t]%htbp % left bottom right top
% \vspace{-0.3cm}
% \hspace{-0.45cm}
\parbox{.99\linewidth}{
% \setlength{\tabcolsep}{0.12em}
% \renewcommand{\arraystretch}{0.70}
%\fontsize{9}{9}\selectfont
\centering
\resizebox{0.48\textwidth}{!}{\begin{tabular}{ l c c c c c c }
\toprule
\multirow{2}{*}{Backbone} & \multicolumn{4}{c}{Modality}& Hal. & Mean \\
\cline{2-5}
& \tiny{BoW/FV/OFF} & ODF & SDF &  AF & loss   & acc. (\%) \\
\midrule
\multirow{8}{*}{I3D}
&  & & & & - & $89.9$ \\
& \checkmark & & & & MSE & $92.6$ \\
& \checkmark & \checkmark &  & &  MSE & $93.2$ \\
& \checkmark  & & \checkmark & &  MSE & $92.8$ \\
% &  \checkmark  & \checkmark & \checkmark & & MSE & $93.3$ \\
% &  \checkmark  & \checkmark & \checkmark  & & MSE & $94.2$ \\
&  \checkmark  & \checkmark & \checkmark  & &  MSE & $94.4$ \\
&  \checkmark  & \checkmark & \checkmark  & &  Uncert. & $94.9$ \\
&  \checkmark  & \checkmark & \checkmark  & \checkmark &  MSE & $95.5$ \\
&  \checkmark  & \checkmark & \checkmark  & \checkmark &  Uncert. & {\bf 96.0} \\
\hline
\multirow{6}{*}{VTN-ResNet}
&  & & & & - & $ 85.0 $ \\
& \checkmark & & & & MSE & $ 89.7 $ \\
&  \checkmark  & \checkmark & \checkmark  & &  MSE & $90.1$ \\
&  \checkmark  & \checkmark & \checkmark  & &  Uncert. & $90.3$ \\
&  \checkmark  & \checkmark & \checkmark  & \checkmark &  MSE & $92.0$ \\
&  \checkmark  & \checkmark & \checkmark  & \checkmark &  Uncert. & {\bf 92.8} \\
\hline
\multirow{6}{*}{FASTER}
&  & & & & - & $ 90.1 $ \\
& \checkmark & & & & MSE & $ 93.0 $ \\
&  \checkmark  & \checkmark & \checkmark  & &  MSE & $93.2$ \\
&  \checkmark  & \checkmark & \checkmark  & &  Uncert. & $93.8$ \\
&  \checkmark  & \checkmark & \checkmark  & \checkmark &  MSE & $94.9$ \\
&  \checkmark  & \checkmark & \checkmark  & \checkmark &  Uncert. & {\bf 95.3} \\
\midrule
\end{tabular}
%\vspace{0.2cm}\\
}}
\parbox{.99\linewidth}{
\fontsize{9}{9}\selectfont
\centering
\resizebox{0.48\textwidth}{!}{\begin{tabular}{ c c }
\kern-0.5em T-ResNet $87.6\%$ \cite{yuppp} &  ADL I3D $91.7\%$ \cite{anoop_advers}\\
\kern-0.5em DEEP-HAL $92.6\%$ \cite{Wang_2019_ICCV} & MSOE-two-stream $91.9\%$ \cite{Hadji_2018_ECCV}\\
% \hline
\bottomrule
\end{tabular}
}}
\caption{Evaluation results for ({\em top}) our proposed methods and ({\em bottom}) comparisons with state-of-the-art approaches on YUP++. Note that GSF is not applicable to this dataset.}
%\vspace{-0.3cm}
\vspace{-0.3cm}
\label{tab:yupf}
\end{table}

\begin{table}[!tb]%htbp % left bottom right top
\parbox{.99\linewidth}{
% \setlength{\tabcolsep}{0.12em}
% \renewcommand{\arraystretch}{0.70}
%\fontsize{9}{9}\selectfont
% \hspace{-0.3cm}
% \vspace{-0.3cm}
%\centering
\resizebox{0.48\textwidth}{!}{\begin{tabular}{ l c c c c c c c}
\toprule
\multirow{2}{*}{Backbone}  & \multicolumn{5}{c}{Modality} & Hal. & \multirow{2}{*}{mAP(\%)} \\
\cline{2-6}
 & \tiny{BoW/FV/OFF} & ODF & SDF & GSF & AF & loss & \\
\midrule
\multirow{7}{*}{I3D}
 &  &  &  &  &  & - & 74.8\\
 & \checkmark &  &  &  &  & MSE & 80.4\\
 & \checkmark & \checkmark & \checkmark &  &  & MSE & 84.8\\
 & \checkmark & \checkmark & \checkmark &  &  & Uncert. & 85.6\\
 & \checkmark & \checkmark & \checkmark & \checkmark &  & Uncert. & 86.2\\
 & \checkmark & \checkmark & \checkmark & \checkmark & \checkmark & MSE & 86.0\\
 & \checkmark & \checkmark & \checkmark & \checkmark & \checkmark & Uncert. & {\bf 86.5}\\
\hline
\multirow{4}{*}{VTN-ResNet}
 &  &  &  &  &  & - & 66.3\\
 & \checkmark & \checkmark & \checkmark &  &  & Uncert. & 75.5\\
  & \checkmark & \checkmark & \checkmark & \checkmark &  & Uncert. & 76.1\\
   & \checkmark & \checkmark & \checkmark & \checkmark& \checkmark  & Uncert. & {\bf 76.7}\\
 \hline
\multirow{4}{*}{FASTER}
 &  &  &  &  &  & - & 75.5\\
 & \checkmark & \checkmark & \checkmark &  &  & Uncert. & 82.9\\
 & \checkmark & \checkmark & \checkmark & \checkmark &  & Uncert. & 83.7\\
 & \checkmark & \checkmark & \checkmark & \checkmark& \checkmark  & Uncert. & {\bf 84.3}\\
\midrule
\end{tabular}
%\vspace{0.02cm}%\\
}}
\parbox{.99\linewidth}{
\centering
% \setlength{\tabcolsep}{0.12em}
% \renewcommand{\arraystretch}{0.70}
% \hspace{-0.45cm}
\fontsize{8}{9}\selectfont
%\centering
\resizebox{0.48\textwidth}{!}{\begin{tabular}{ c c }
%\hline
%\kern-0.5em KRP-FS $70.0\%$ \cite{anoop_rankpool_nonlin} & KRP-FS+IDT $76.1\%$ \cite{anoop_rankpool_nonlin}\kern-0.5em\\
%\kern-0.5em GRP $68.4\%$ \cite{anoop_generalized} & GRP+IDT $75.5\%$ \cite{anoop_generalized}\kern-0.5em\\
\kern-0.5em KRP-FS+IDT $76.1\%$ \cite{anoop_rankpool_nonlin} & GRP+IDT $75.5\%$ \cite{anoop_generalized}\kern-0.5em\\
\kern-0.5em I3D+BoW/OFF MTL $79.1\%$ \cite{Wang_2019_ICCV} & DEEP-HAL $81.8\%$ \cite{Wang_2019_ICCV}\\
% \hline
\bottomrule
\end{tabular}
}}
\caption{Evaluation results for ({\em top}) our proposed methods and ({\em bottom}) comparisons with state-of-the-art approaches on the MPII dataset.}
%\vspace{-0.3cm}
\vspace{-0.5cm}
\label{tab:mpiif}
\end{table}

\begin{table}[!tb]%htbp % left bottom right top
\parbox{.99\linewidth}{
\setlength{\tabcolsep}{0.12em}
\renewcommand{\arraystretch}{0.70}
%\fontsize{9}{9}\selectfont
% \hspace{-0.3cm}
% \vspace{-0.3cm}
%\centering
\resizebox{0.48\textwidth}{!}{\begin{tabular}{ l c c c c c c c c c c}
\toprule
\multirow{2}{*}{Backbone}  & \multicolumn{6}{c}{Modality} & SK & num. of & input to & mAP \\
\cline{2-7}
 & \tiny{BoW/FV} & OFF & ODF & SDF & GSF & AF & dim. $d$ & MSK & CENet & (\%) \\
\midrule
\multirow{21}{*}{I3D}
 &  &  &  &  &  & & - & - & - & 37.2\\
 & \checkmark &  &  &  &  & & 1000 & - & - & 41.9\\
 & \checkmark & \checkmark &  &  &  & & 1000 & 2 & - & 42.0\\
 & \checkmark & \checkmark &  &  &  & & 1000 & 4 & - & 42.2\\
 & \checkmark & \checkmark &  &  &  & & 1000 & 8 & - & 43.1\\
& \checkmark & \checkmark & \checkmark &  &  & & 512 & 8 & - & 47.2\\
& \checkmark & \checkmark &  & \checkmark &  & & 512 & 8 & - & 45.3\\
& \checkmark & \checkmark & \checkmark & \checkmark &  & & 512 & 8 & - & 49.1\\
& \checkmark & \checkmark & \checkmark & \checkmark &  & & 1000 & 8 & - & 50.1\\
& \checkmark & \checkmark & \checkmark & \checkmark & \checkmark & & 512 & 8 & - & 50.9\\
& \checkmark & \checkmark & \checkmark & \checkmark & \checkmark & & 256 & 8 & - & 50.5\\
& \checkmark & \checkmark & \checkmark & \checkmark & \checkmark & & 128 & 8 & - & 50.3\\
& \checkmark & \checkmark & \checkmark & \checkmark & \checkmark & & 128 & 8 & $\mathcal{X}_{(rgb)}$ & 50.5\\
& \checkmark & \checkmark & \checkmark & \checkmark & \checkmark & & 128 & 8 & $\tilde{\vpsi}$ & 51.0\\
& \checkmark & \checkmark & \checkmark & \checkmark & \checkmark & & 128 & 12 & $\tilde{\vpsi}$ & 51.0\\
& \checkmark & \checkmark & \checkmark & \checkmark & \checkmark & \checkmark & 512 & 8 & - & 51.6\\
& \checkmark & \checkmark & \checkmark & \checkmark & \checkmark & \checkmark & 256 & 8 & - & 51.5\\
& \checkmark & \checkmark & \checkmark & \checkmark & \checkmark & \checkmark & 128 & 8 & - & 51.2\\
& \checkmark & \checkmark & \checkmark & \checkmark & \checkmark & \checkmark & 128 & 8 & $\mathcal{X}_{(rgb)}$ & 51.3\\
& \checkmark & \checkmark & \checkmark & \checkmark & \checkmark & \checkmark & 128 & 8 & $\tilde{\vpsi}$ & 52.0\\
& \checkmark & \checkmark & \checkmark & \checkmark & \checkmark & \checkmark & 128 & 12 & $\tilde{\vpsi}$ & {\bf 52.1}\\
\hline
\multirow{15}{*}{AssembleNet++}
 &  &  &  &  &  & & - & - & - & 53.8\\
  &  &  &  &  &  & & - & - & - & 56.7$^*$\\
 & \checkmark & \checkmark & \checkmark & \checkmark &  & & 512 & 8 & - & 55.8\\
  & \checkmark & \checkmark & \checkmark & \checkmark &  & & 512 & 8 & - & 60.7$^*$\\
& \checkmark & \checkmark & \checkmark & \checkmark &  & & 1000 & 8 & - & 56.9\\
& \checkmark & \checkmark & \checkmark & \checkmark &  & & 1000 & 8 & - & 62.0$^*$\\
& \checkmark & \checkmark & \checkmark & \checkmark & \checkmark & & 512 & 8 & - & 58.0\\
& \checkmark & \checkmark & \checkmark & \checkmark & \checkmark & \checkmark & 512 & 8 & - & 58.5\\
& \checkmark & \checkmark & \checkmark & \checkmark & \checkmark & \checkmark & 256 & 8 & - & 58.2\\
& \checkmark & \checkmark & \checkmark & \checkmark & \checkmark & \checkmark & 128 & 8 & - & 58.0\\
& \checkmark & \checkmark & \checkmark & \checkmark & \checkmark & \checkmark & 128 & 8 & $\mathcal{X}_{(rgb)}$ & 58.3\\
& \checkmark & \checkmark & \checkmark & \checkmark & \checkmark & \checkmark & 128 & 8 & $\mathcal{X}_{(rgb)}$ & 64.7$^*$\\
& \checkmark & \checkmark & \checkmark & \checkmark & \checkmark & \checkmark & 128 & 8 & $\tilde{\vpsi}$ & 58.8\\
& \checkmark & \checkmark & \checkmark & \checkmark & \checkmark & \checkmark & 128 & 12 & $\tilde{\vpsi}$ & {\bf 59.0}\\
& \checkmark & \checkmark & \checkmark & \checkmark & \checkmark & \checkmark & 128 & 12 & $\tilde{\vpsi}$ & {\bf 65.3}$^*$\\
%  &  &  &  &  &  & & - & - & - & \\
\hline
\multirow{9}{*}{VTN}
 &  &  &  &  &  & & - & - & - & 43.5\\
 & \checkmark & \checkmark & \checkmark & \checkmark &  & & 512 & 8 & - & 48.3\\
  & \checkmark & \checkmark & \checkmark & \checkmark &  & & 256 & 8 & - & 48.0\\
  & \checkmark & \checkmark & \checkmark & \checkmark &  & & 128 & 8 & - & 47.8\\
  & \checkmark & \checkmark & \checkmark & \checkmark & \checkmark & & 128 & 8 & - & 50.2\\
  & \checkmark & \checkmark & \checkmark & \checkmark & \checkmark & \checkmark & 128 & 8 & - & 53.2\\
  & \checkmark & \checkmark & \checkmark & \checkmark & \checkmark & \checkmark & 128 & 8 & $\mathcal{X}_{(rgb)}$ & 53.7\\
  & \checkmark & \checkmark & \checkmark & \checkmark & \checkmark & \checkmark & 128 & 8 & $\tilde{\vpsi}$ & 54.3\\
  & \checkmark & \checkmark & \checkmark & \checkmark & \checkmark & \checkmark & 128 & 12 & $\tilde{\vpsi}$ & {\bf 54.5}\\
  % &  &  &  &  &  & & - & - & - & \\
  \hline
\multirow{9}{*}{FASTER}
 &  &  &  &  &  & & - & - & - & 40.7\\
 & \checkmark & \checkmark & \checkmark & \checkmark &  & & 512 & 8 & - & 46.0\\
  & \checkmark & \checkmark & \checkmark & \checkmark &  & & 256 & 8 & - & 45.2\\
  & \checkmark & \checkmark & \checkmark & \checkmark &  & & 128 & 8 & - & 45.0\\
  & \checkmark & \checkmark & \checkmark & \checkmark & \checkmark & & 128 & 8 & - & 47.8\\
  & \checkmark & \checkmark & \checkmark & \checkmark & \checkmark & \checkmark & 128 & 8 & - & 49.7\\
  & \checkmark & \checkmark & \checkmark & \checkmark & \checkmark & \checkmark & 128 & 8 & $\mathcal{X}_{(rgb)}$ & 51.2\\
  & \checkmark & \checkmark & \checkmark & \checkmark & \checkmark & \checkmark & 128 & 8 & $\tilde{\vpsi}$ & 52.7\\
  & \checkmark & \checkmark & \checkmark & \checkmark & \checkmark & \checkmark & 128 & 12 & $\tilde{\vpsi}$ & {\bf 53.3}\\
\midrule
\end{tabular}
%\vspace{0.02cm}%\\
}}
\parbox{.99\linewidth}{
\fontsize{9}{9}\selectfont
\centering
\resizebox{0.48\textwidth}{!}{\begin{tabular}{ c c }
\kern-0.5em LFB $42.5\%$ \cite{Wu_2019_CVPR} &  ActionCLIP  $44.3\%$ \cite{DBLP:journals/corr/abs-2109-08472}\\
\kern-0.5em En-VidTr-L $47.3\%$ \cite{Zhang_2021_ICCV} &  MoViNet-A4  $48.5\%$ \cite{kondratyuk2021movinets}\\
\kern-0.5em SlowFast $45.2\%$ \cite{slowfast} &  AssembleNet  $51.6\%$ \cite{assemblenet}\\
\kern-0.5em AssembleNet-101 $58.6\%$ \cite{assemblenet} & AssembleNet++ 50  $59.8\%$ \cite{assemblenet_plus}\\
% \hline
\bottomrule
\end{tabular}
}}
{\textsuperscript{*}\footnotesize{These results are obtained without pre-training on Kinetics-400.}}
\caption{Evaluation results for ({\em top}) our methods and ({\em bottom}) comparisons with state-of-the-art approaches on the Charades dataset.}
%\vspace{-0.3cm}
%\vspace{-0.5cm}
\label{tab:charades}
\end{table}

\newcommand{\fsnine}[0]{\fontsize{9}{9}\selectfont}
\newcommand{\fsninee}[0]{\fontsize{9}{9}\selectfont}

\begin{table}[!ht]
\begin{center}
% \vspace{-0.3cm}
% \vspace{-0.1cm}
\resizebox{0.48\textwidth}{!}{\begin{tabular}{ l c  c c  c c  c c}
\toprule
% \hline
% Backbone  & \multicolumn{3}{c}{Modality} & SK & \multicolumn{2}{c}{Verbs} & \multicolumn{2}{c}{Nouns} & \multicolumn{2}{c}{Actions}\\
% % \hline
% \cline{2-4} \cline{6-11}
% (+\tiny{BoW/FV/OFF}) & \tiny{ODF/SDF}& GSF & AF & dim. $d$ & top-1 & top-5 & top-1 & top-5 & top-1 & top-5\\
Backbone  & \multicolumn{3}{c}{Modality} & SK & \multirow{2}{*}{Verbs} & \multirow{2}{*}{Nouns} & \multirow{2}{*}{Actions}\\
% \hline
\cline{2-4}
+\tiny{BoW/FV/OFF} & \tiny{ODF/SDF}& GSF & AF & dim. $d$ & & & \\
\midrule
& & & & & \multicolumn{3}{c}{\bf Validation}\\
\cline{6-8}
\multirow{7}{*}{I3D}
& \checkmark & & & 1000 & 55.4 & 33.3 & 21.5 \\
& \checkmark & \checkmark & & 512 & 59.2  & 38.1 & 26.4 \\
& \checkmark & \checkmark & & 256 & 59.0 & 37.7 & 26.0 \\
& \checkmark & \checkmark & & 128 & 58.6 & 37.1 & 25.6 \\
& \checkmark & \checkmark & \checkmark & 512 & {\bf 62.0} & {\bf 40.9} & {\bf 30.5}\\
& \checkmark & \checkmark & \checkmark & 256 & {\bf 61.7} & {\bf 40.4} & {\bf 30.3}\\
& \checkmark & \checkmark & \checkmark & 128 & 60.9 & 39.0 & 29.7\\
% & & & & & & & & & & \\
% \hline
\hdashline
\multirow{5}{*}{AssembleNet++}
& \checkmark & & & 512 & 57.2 & 34.8 & 23.2\\
& \checkmark & & & 256 & 56.8 & 34.6 &22.9\\
& \checkmark & & & 128 & 56.6 & 34.0 &22.5\\
& \checkmark & \checkmark & & 128 & 63.5 & 41.2 &40.9\\
& \checkmark & \checkmark & \checkmark & 128 & {\bf 68.2} & {\bf 46.1}& {\bf 36.3}\\
% & & & & & & & & & & \\
% & & & & & & & & & & \\
% & & & & & & & & & & \\
\hdashline
\multirow{6}{*}{VTN}
& \checkmark  & & & 256 & 54.3 & 32.7 & 19.8\\
& \checkmark  & & & 128 & 54.0 & 32.6 & 19.7\\
& \checkmark  & \checkmark  & & 256 & 58.0 & 36.2 & 25.2 \\
& \checkmark  & \checkmark  & & 128 & 57.6 & 36.0 & 24.7\\
& \checkmark  & \checkmark  &  \checkmark  & 256 & {\bf 60.0} & {\bf 38.8} & {\bf 29.4}\\
& \checkmark  & \checkmark  &  \checkmark  & 128 & {\bf 59.6} & {\bf 38.2} & {\bf 29.3}\\
\hdashline
\multirow{6}{*}{FASTER}
&\checkmark & & & 256& 55.0 & 33.1 & 20.9 \\
&\checkmark & & & 128& 54.2 & 32.7 & 19.5 \\
&\checkmark& \checkmark & & 256 & 58.5 & 36.7 & 24.3\\
&\checkmark & \checkmark& & 128& 58.1 & 36.0 & 23.9\\
&\checkmark &\checkmark & \checkmark& 256 & {\bf 62.0} & {\bf 41.8} & {\bf 31.5}\\
&\checkmark & \checkmark& \checkmark& 128 & 61.3 & 41.6 & 31.0\\
\hdashline
& \multicolumn{4}{c}{LFB Max~\cite{Wu_2019_CVPR}} & 52.6 & 31.8 & 22.8\\
& \multicolumn{4}{c}{WeakLargeScale~\cite{Ghadiyaram_2019_CVPR}} & 58.4 & 36.9 & 26.1\\
\midrule
& & & & & \multicolumn{3}{c}{\bf Test s1 (seen)}\\
\cline{6-8}
I3D&\checkmark & \checkmark& \checkmark& 256 & 70.0 & 50.7 & 38.7\\
\hdashline
AssembleNet++&\checkmark & \checkmark& \checkmark& 128 & 76.5 & 57.2 & 47.9\\
\hdashline
VTN &\checkmark & \checkmark& \checkmark& 128 & 67.7 & 49.2 & 38.0\\
\hdashline
FASTER &\checkmark & \checkmark& \checkmark& 256 & 71.0 & 51.1 & 39.2\\
\hdashline
& \multicolumn{4}{c}{TSN Fusion~\cite{Damen_2018_ECCV}} & 48.2 & 36.7 & 20.5\\
& \multicolumn{4}{c}{LFB Max~\cite{Wu_2019_CVPR}} & 60.0 & 45.0 & 32.7\\
& \multicolumn{4}{c}{WeakLargeScale~\cite{Ghadiyaram_2019_CVPR}} & 65.2 & 45.1 & 34.5\\
\midrule
& & & & & \multicolumn{3}{c}{\bf Test s2 (unseen)}\\
\cline{6-8}
I3D&\checkmark & \checkmark& \checkmark& 256 & 61.7 & 40.0 & 30.3\\
\hdashline
AssembleNet++&\checkmark & \checkmark& \checkmark& 128 & 68.9 & 48.2 & 39.1\\
\hdashline
VTN &\checkmark & \checkmark& \checkmark& 128 & 58.8 & 39.6 & 29.0\\
\hdashline
FASTER &\checkmark & \checkmark& \checkmark& 256 & 63.6 & 40.9 & 30.9\\
\hdashline
& \multicolumn{4}{c}{TSN Fusion~\cite{Damen_2018_ECCV}} & 39.4 & 22.7 & 10.9\\
& \multicolumn{4}{c}{LFB Max~\cite{Wu_2019_CVPR}} & 50.9 & 31.5 & 21.2\\
& \multicolumn{4}{c}{WeakLargeScale~\cite{Ghadiyaram_2019_CVPR}} & 57.3 & 35.7 & 25.6\\
\bottomrule
\end{tabular}}
\end{center}
\caption{Experimental results on the EPIC-KITCHENS-55 dataset.
}
\label{epic-kitchens}
% \vspace{-0.5cm}
\end{table}

\begin{table}[!ht]
\begin{center}
% \vspace{-0.3cm}
% \vspace{-0.1cm}
\resizebox{0.48\textwidth}{!}{\begin{tabular}{ l c  c c  c c  c c}
\toprule
Backbone  & \multicolumn{3}{c}{Modality} & Hal. & \multirow{2}{*}{CS} & \multirow{2}{*}{CV$_1$} & \multirow{2}{*}{CV$_2$}\\
\cline{2-4}
+\tiny{BoW/FV/OFF} & \tiny{ODF/SDF}& GSF & AF & loss & & & \\
\hline
\multirow{5}{*}{I3D}
 &  & & & - & 53.4 & 34.9 & 45.1\\
 & \checkmark & & & MSE & 57.6 & 38.2 & 49.3\\
 & \checkmark & & & Uncert. & 58.2 & 39.0 & 50.1\\
 & \checkmark & \checkmark & & Uncert. & 62.3 & 43.1 & 55.2\\
 & \checkmark & \checkmark & \checkmark & Uncert. & {\bf 65.1} & {\bf 44.3} & {\bf 56.3}\\
\hline
\multirow{5}{*}{AssembleNet++}
& & & & - & 63.6 & 45.2 & 55.8 \\
& \checkmark & & & MSE & 65.5 & 48.0 & 59.0 \\
& \checkmark & & & Uncert. & 66.0 & 48.7 & 59.9 \\
& \checkmark & \checkmark & & Uncert. & 70.8 & 52.7 & 65.5 \\
& \checkmark & \checkmark & \checkmark & Uncert. & {\bf 72.3} & {\bf 54.1} & {\bf 68.8} \\
\hline
\multirow{5}{*}{VTN}
& & & & - & 53.0 & 33.2 & 43.7 \\
& \checkmark & & & MSE & 57.0 & 38.2 & 48.7 \\
& \checkmark & & & Uncert. & 57.8 & 39.3 & 49.2 \\
& \checkmark & \checkmark & & Uncert. & 61.3 & 43.0 & 54.1 \\
& \checkmark & \checkmark & \checkmark & Uncert. & {\bf 62.6} & {\bf 43.9} & {\bf 55.7} \\
\hline
\multirow{5}{*}{FASTER}
& & & & - & 53.7 & 35.0 & 46.7\\
& \checkmark & & & MSE & 59.1 & 41.2 & 53.2 \\
& \checkmark & & & Uncert. & 60.0 & 42.1 & 54.0 \\
& \checkmark & \checkmark & & Uncert. & 63.9 & 46.5 & 57.7 \\
& \checkmark & \checkmark & \checkmark & Uncert. & {\bf 65.5} & {\bf 48.8} & {\bf 59.4} \\
\hline
\multicolumn{5}{c}{Separable STA~\cite{Das_2019_ICCV}} & 54.2 & 35.2 & 50.3\\
\multicolumn{5}{c}{NPL~\cite{Piergiovanni_2021_CVPR}} & - & 39.6 & 54.6\\
\multicolumn{5}{c}{VPN~\cite{das2020vpn}} & 60.8 & 43.8 & 53.5\\
\multicolumn{5}{c}{UNIK~\cite{yang2021unik}} &  63.1 &  22.9& 61.2\\
\bottomrule
\end{tabular}}

\end{center}
\caption{Experimental results on the Toyota Smarthome dataset.
}
\label{toyota-smarthome}
% \vspace{-0.5cm}
\end{table}

\subsection{Discussion on Fine-grained Action Recognition}

In this section, we choose the I3D backbone to discuss fine-grained AR.

% \vspace{0.05cm}
% \noindent{\textbf{HMDB-51.}} Table \ref{tab:hmdb51f} shows the results on HMDB-51.  I3D backbone with our ODF and SDF descriptors  (ODF only) and (SDF only) outperform ({\em DEEP-HAL}) by $\sim\!1.8\%$ and $\sim\!1.4\%$, resp. This shows that both ODF and SDF are effective. 
% Combining I3D backbone, ODF and SDF  outperform DEEP-HAL by $\sim\!2.7\%$ demonstrating the complementary nature of ODF and SDF.

% \vspace{0.05cm}
% \noindent{\textbf{YUP++.}} Table \ref{tab:yupf} shows that ODF is better than SDF (with I3D backbone), that is (ODF only) and (SDF only) outperform ({\em DEEP-HAL}) by $\sim\!0.6\%$ and $\sim\!0.2\%$, resp. This is expected as YUP++ contains dynamic scenes without objects/specific saliency regions correlating with class concepts. 

\vspace{0.05cm}
\noindent{\textbf{MPII.}} Table \ref{tab:mpiif} shows a $\sim\!3.0\%$ mAP improvement over the ({\em DEEP-HAL}) baseline due to the detectors capturing human-object interaction, which helps model fine-grained AR. Additionally, incorporating skeleton and audio modalities further improves the performance by approximately 1\%, even though these modalities are not available in the MPII dataset.

\vspace{0.05cm}
\noindent{\textbf{Charades.}} Table \ref{tab:charades} (top) presents the relative gains of our hallucination pipeline using the I3D backbone. We evaluate both ({\em ODF}) and ({\em SDF}) with 512-dimensional sketching ({\em SK512})  
and observe that ODF outperforms SDF, with both methods surpassing the baseline ({\em DEEP-HAL}) \cite{Wang_2019_ICCV}. 

We also observe that combining ODF and SDF (with SK512) achieves a $49.1\%$ mAP, representing a $\sim\!\mathbf{6}\%$ gain over the baseline. This demonstrates the high complementarity of ODF and SDF. Using a larger sketch (ODF/SDF, SK1000) results in a $50.1\%$ mAP, which closely matches the performance of ({\em DEEP-HAL with ODF/SDF (exact)}, 50.16\%), where `exact' denotes late fusion by concatenation of ODF and SDF streams with the DEEP-HAL stream fed into PredNet. 
The matching results between ({\em DEEP-HAL with ODF/SDF (SK1000)}) and  ({\em DEEP-HAL+ODF/SDF (exact)}) show that we can hallucinate ODF and SDF at test time while maintaining full performance, thus saving computational time and boosting reults on Charades by $\sim\!\mathbf{6}\%$ over the baseline. Moreover, with additional skeleton (GSF) and audio information (AF), our model achieves 52.1\%, further improving performance by approximately 2\%. In contrast, SlowFast networks \cite{slowfast} and AssembleNet \cite{assemblenet} achieve 45.2\% and 51.6\% on Charades, respectively. 

\vspace{0.05cm}
\noindent{\textbf{EPIC-KITCHENS-55.}}Table~\ref{epic-kitchens} shows the experimental results. Our model learns human-like semantic features due to ODF/SDF, and even skeleton and audio information can be synthesized. There is no evidence suggesting that a backbone alone can discover these features without guidance. Comparing MPII (3748 clips) with the larger EPIC-KITCHENS-55 (39594 clips), both related to cooking tasks, SDF+ODF boosts MPII from 81.8 to 84.8\%, and boosts EPIC-KITCHENS-55 from 32.51\% (DEEP-HAL) to 35.88\% (on the seen classes protocol), and from 22.33\% (DEEP-HAL) to 27.32\% (on the unseen classes protocol). This demonstrates a $\sim$\textbf{3}\% improvement on both MPII and EPIC-KITCHENS-55, with EPIC-KITCHENS-55 containing nearly 10$\times$ more clips than MPII. With the addition of GSF and AF, we achieve an extra performance gain of around 8\%.

\vspace{0.05cm}
\noindent{\textbf{Toyota Smarthome.}} Table~\ref{toyota-smarthome} presents results for Toyota Smarthome. With ODF/SDF, our model outperforms I3D baseline by 4\% on average. Since this dataset provides reliable skeleton information, adding GSF improves the performance by an additional 4\%. This suggests that robust skeletons enhance AR.
Moreover, hallucinating AF further boosts performance by 1--3\%. Even with the I3D backbone (using only RGB), our model still achieves very competitive results, outperforming recent methods such as VPN~\cite{das2020vpn} and UNIK~\cite{yang2021unik} by 3\% and 6\% (on average), respectively.

\subsection{Discussion on the Uncertainty Learning}

We now evaluate the performance of using MSE and uncertainty learning for feature hallucination. We observe that uncertainty learning helps improve the AR performance across all datasets. On HMDB-51 and YUP++ (Table~\ref{tab:hmdb51f} and~\ref{tab:yupf}), uncertainty learning loss improves performance by approximately 1\% on average for all backbones. On MPII (Table~\ref{tab:mpiif}), hallucinating additional ODF/SDF with uncertainty learning outperforms using MSE by 0.8\% with the I3D backbone. Furthermore, hallucinating all features with uncertainty learning improves performance by an additional 0.5\%.

As mentioned earlier, skeleton data often contains noise, which affects the GSF, we evaluate the impact of noisy GSF on AR. On Toyota Smarthome, hallucinating GSF improves AR by 4--5\% across all backbones, as this dataset provides skeleton data. Using GSF on Charades improves results by 2--3\% on average, despite Charades not having skeleton data; in this case, we use the GSF stream pre-trained on skeleton data from Toyota Smarthome for hallucination. The use of skeleton information improves AR performance by approximately 1\% and 4\% on MPII and EPIC-KITCHENS-55, respectively. Note that these two cooking datasets are fine-grained AR datasets focusing on specific regions of actions (\eg, hands), making it more difficult to obtain full human skeleton data. Nevertheless, our proposed model, with uncertainty, is still able to hallucinate some skeleton features even without full human skeleton data.

Additionally, videos typically contain background sounds that are irrelevant to the actions observed, such as music, TV, or the noise from coffee or washing machines. In Table~\ref{epic-kitchens}, we see that the inclusion of audio features boosts AR by more than 3\% across all backbones. The largest performance gain (around 5\%) is observed when using the AssembleNet++ backbone. Our experimental results demonstrate that these irrelevant sounds do not confuse the model, highlighting the robustness of our approach to noisy and unconstrained audio sources.

\subsection{Discussion on Modalities}

We observe that using additional modalities helps improve AR. The performance gain from the use of ODF/SDF is approximately 2\%, 1.5\%, 4\%, and 3\% on HMDB-51, YUP++, MPII, and Charades, respectively. On fine-grained AR tasks, the performance gain from using ODF/SDF averages around 3\%. This suggests that object detection and saliency information contribute significantly to AR, particularly in fine-grained AR, where the surrounding objects of performers provide important cues to actions.

The inclusion of skeleton features further improves performance by approximately 1--2\% on average for HMDB-51, MPII and Charades. On EPIC-KITCHENS-55 and Toyota Smarthone, the performance gain from skeleton features is more substantial, exceeding 4\% and 3--4\%, respectively. Since YUP++ is a natural scene classification dataset and does not require skeleton information, we do not hallucinate skeleton features for this dataset.

We also find that the use of sound features further boosts AR performance. Adding sound information increases performance by approximately 3.5\% on average for EPIC-KITCHENS-55 and Toyota Smarthome.
This is expected, as visual and audio modalities are often highly correlated. For fine-grained AR, audio information is especially important for action-related tasks.
On Charades, the use of the VTN backbone yields the largest performance gain (around 3\%) for hallucinated sound information, as VTN is particularly well-suited for handling audio features.
On HMDB-51 and MPII, the performance gain is smaller (around 0.5--1\%) due to the lack of reliable sound information. 
For natural scene classification on YUP++, the sound features improve performance by 1--2\%.

Our analysis highlights that AR benefits from additional modalities such as skeleton and sound, in addition to the commonly used RGB and optical flow videos. Our proposed model can synthesize multiple modalities through a simple hallucination step, boost AR performance without introducing extra computational cost during the test stage.

\subsection{\lei{Discussion on the Large-scale Datasets}}

\begin{table}[t]%htbp % left bottom right top
% \setlength{\tabcolsep}{0.12em}
% \renewcommand{\arraystretch}{0.70}
%\fontsize{9}{9}\selectfont
\centering
\resizebox{0.8\linewidth}{!}{\begin{tabular}{ l c c c }
\toprule
&  K400 & K600 &  SSv2\\
\hline
VideoMAE V2~\cite{wang2023videomae} &  87.2 &  88.8 & 77.0 \\
+ BoW/FV & 87.5 & 88.9 & 77.0 \\
+ BoW/FV + OFF & \textbf{87.6} & \textbf{89.1} & 77.3 \\
+ BoW/FV + OFF + ODF/SDF & 87.5 & \textbf{89.1} & \textbf{77.4} \\
\hline
InternVideo2$_{s1}$ \cite{wang2024internvideo2} & 91.3 & 91.4 & 77.1 \\
+ BoW/FV & \textbf{91.8} & 91.6 & 77.0 \\
+ BoW/FV + OFF & \textbf{91.8} & \textbf{91.7} & 77.2 \\
+ BoW/FV + OFF + ODF/SDF & 91.6 & 91.5 & \textbf{77.3} \\
\bottomrule
\end{tabular}% }
}
\caption{Experimental results on the large-scale Kinetics-400 (K400), Kinetics-600 (K600), and Something-Something V2 (SSv2) datasets. For both VideoMAE V2 (ViT-g) and InternVideo2$_{s1}$ (1B), finetuned model weights specific to each dataset are used. Ground-truth descriptors for BoW/FV, OFF, and ODF/SDF used in feature hallucination are generated using 80,000 training samples from Mini-Kinetics-200~\cite{Xie_2018_ECCV}, balancing computational cost and feature storage constraints.}
\label{tab:large-scale}
\end{table}

\lei{We also evaluate our model on three widely-used large-scale action recognition datasets: Kinetics-400, Kinetics-600, and Something-Something V2. The experimental results are summarized in Table~\ref{tab:large-scale}.}

\lei{To address computational costs and feature storage constraints, we use 80,000 training samples from Mini-Kinetics-200~\cite{Xie_2018_ECCV} to generate ground-truth descriptors for BoW, FV, OFF, ODF and SDF, which are used for feature hallucination. For this process, we use our uncertainty loss and set the sketching dimension to $d=128$ for the power-normalized ground-truth descriptors. These descriptors are hallucinated from the pooled spatiotemporal token embeddings of pretrained VideoMAE V2 (ViT-g) and InternVideo2$_{s1}$-1B encoders.
After training the hallucination streams, we freeze their weights and proceed to fine-tune the HAF and PredNet components for each dataset. This approach allows HAF and PredNet to adapt to the hallucinated features of different datasets, enhancing both computational and storage efficiency.}

\lei{Notably, despite the hallucinated streams being trained on feature descriptors derived from a subset of Kinetics-400 (Mini-Kinetics-200), they still enhance action recognition performance on large-scale datasets. Interestingly, the improvements brought by BoW/FV and BoW/FV+OFF are particularly significant, yielding gains of over 0.4\%, 0.3\%, and 0.1\% on Kinetics-400, Kinetics-600, and Something-Something V2, respectively. Adding ODF and SDF contributes only marginal improvements. This is likely because both VideoMAE V2 and InternVideo2 are self-supervised learning frameworks that effectively capture deep semantic features through spatiotemporal masking and video frame reconstruction. However, handcrafted descriptors like those encoded via BoW and FV still play a valuable role in boosting performance, even when derived from a subset of Kinetics-400 and pretrained hallucinated streams.}

\lei{On Something-Something V2, we observe that BoW/FV does not improve performance for either VideoMAE V2 or InternVideo2. However, incorporating OFF significantly enhances performance, underscoring the challenging nature of Something-Something V2, which requires robust temporal reasoning. Motion-related information, such as that captured by OFF, proves crucial for these improvements.}

\lei{In contrast, on Kinetics-400 and Kinetics-600, adding OFF results in negligible performance gains. This indicates that motion information is far less relevant for the Kinetics datasets, as also demonstrated in recent studies~\cite{patrick2021keeping,girdhar2023omnimae}. These findings highlight the differing demands of these datasets, with temporal motion cues playing a pivotal role in Something-Something V2 but being less critical for the Kinetics datasets. This highlights the need for video understanding researchers to collect and curate datasets where motion, temporal information, and reasoning are crucial, fostering advancements that better serve the video understanding research community.}

\subsection{Discussion on Action Recognition Backbones}

We observe that the top performance of our model depends on the backbone used. Based on our comparisons, the AssembleNet++ backbone performs the other three backbones (I3D, VTN and FASTER). Models with AssembleNet++ achieve a performance boost of approximately 13\%, 5\%, and 10\% over I3D and FASTER backbones on Charades, EPIC-KITCHENS-55, and Toyota Smarthome, respectively. 

The VTN backbone (with ResNet) performs slightly worse than the other three backbones (AssembleNet, I3D and FASTER), as it is a lightweight model designed for AR on hardware with limited computational power (\eg, mobile devices). The FASTER framework is also a lightweight model that avoids redundant computation between neighboring clips. However, it uses an efficient model (R(2+1)D-50) for subtle motions, performing better than VTN by approximately 8\%, 3\%, 8\%, 2\%, and 4\% on HMDB-51, YUP++, MPII, EPIC-KITCHENS-55, and Toyota Smarthome, respectively. On charades, VTN performs slightly better than FASTER (by about 1\%). This can be attributed to: (i) the lack of available skeleton data and unreliable pose estimators on this dataset (due to pose complexity and mobile camera movements), (ii) noise in the sound data, and (iii) VTN's ability to handle time series data (\eg, audio and skeletons) more effectively.

We also observe that, for the hallucination task, the AssembleNet++ backbone generally outperforms the FASTER and I3D backbones, which perform equally well and both outperform VTN. This is expected because: (i) AssembleNet++ is optimized for hallucination tasks, (ii) both FASTER and I3D backbones use (2+1)D or 3D ConvNets, which are better suited for hallucinating spatio-temporal features (\eg, OFF, GSF), and (iii) VTN only uses 2D CNNs for video frame embedding, which is less efficient compared to the use of (2+1)D and 3D CNNs.

Although the VTN backbone performs worse than the others, its performance remains competitive compared to most existing state-of-the-art methods, especially on fine-grained AR tasks such as Charades, EPIC-KITCHENS-55, and Toyota Smarthome (Table~\ref{tab:charades},~\ref{epic-kitchens} and~\ref{toyota-smarthome}).

Recent self-supervised pretraining frameworks, such as VideoMAE V2 and InternVideo2 (Table~\ref{tab:large-scale}), have emerged as powerful video learners, effectively capturing self-supervisory features that were traditionally the domain of handcrafted methods. These video foundation models, pretrained on large-scale visual and motion datasets, are designed to extensively use spatiotemporal information, excelling in various video processing tasks.
Interestingly, despite the remarkable capabilities of these models, we observe that handcrafted descriptors still provide a meaningful performance boost. This can be attributed to their unique design and specialized focus, which complement the broader but more generalized feature representations learned by these video foundation models.

Our approach is `orthogonal' to these developments, which focus on extensive mining for combinations of neural blocks/dataflows to obtain an `optimal' pipeline. 
We achieve similar results with a simpler approach based on self-supervised learning. Our pipeline is more lightweight by comparison, as it does not require computations of  optical flow, detections, or segmentation masks at test time.

\subsection{Computational Costs and Efficiency}

\begin{table}[tbp]%htbp % left bottom right top
%\hspace{-0.9cm}
% \vspace{-0.3cm}
\setlength{\tabcolsep}{0.12em}
% \renewcommand{\arraystretch}{0.70}
%\fontsize{9}{9}\selectfont
\centering
\resizebox{\linewidth}{!}{\begin{tabular}{ l c c c c c }
\toprule
 & \fsninee {\em DET1:} & \fsninee {\em DET2:} & \fsninee {\em DET3:} & \fsninee {\em DET4:} & \fsninee ODF  \\
 & \fsninee {\em Inception} & \fsninee {\em Inception} & \fsninee {\em ResNet101} & \fsninee {\em NASNet} & \fsninee total  \\
 & \fsninee {\em V2} & \fsninee {\em ResNet V2} & \fsninee {\em  AVA} &  & \fsninee (+SVD)  \\
\hline
{\em sec. per frame}   & 0.07		& 0.38  & 0.10 & 0.91 & 1.46 (+0.09) \\
\hline
{\em s.p.c.} HMDB-51   & 6.5		& 35.3  & 9.3 & 84.5 		& 135.6 (+0.5)\\
{\em s.p.c.} YUP++     & 9.7		& 52.7  & 13.9 & 126.2  & 202.5 (+0.8)\\
{\em s.p.c.} MPII      & 12.4		& 67.1  & 17.7 & 160.8  & 258.0 (+1.3)\\
{\em s.p.c.} Charades  & 21.0		& 114.2  & 30.0 & 273.5 & 438.7 (+2.6)\\
{\em s.p.c.} EPIC-Kitchens  & 	20.3	& 110.4  & 29.0 & 264.3 & 424.0 (+2.6)\\
{\em s.p.c.} Toyota Smarthome  &  16.9 & 92.0  & 24.2 & 220.2 & 353.3 (+2.1) \\
% \hline
\bottomrule
\end{tabular}
}% }
\caption{Statistics for the object detectors used in our experiments. The table provides timings such as seconds per frame (denoted as \emph{sec. per frame}) and seconds per clip (denoted as \emph{s.p.c.}) for detectors used by ODF. Additionally, the total time incurred by a combined detector (\emph{ODF total}) is shown. We also report the time taken for the full Singular Value Decomposition (SVD) and all other ODF operations, assuming approximately 5 detections per frame.
}
%\vspace{-0.3cm}
\label{tab:stat1}
\end{table}

\begin{table}[tbp]%htbp % left bottom right top
\setlength{\tabcolsep}{0.5em}
% \renewcommand{\arraystretch}{0.70}
%\fontsize{9}{9}\selectfont
%MNL~\cite{Zhang_2018_CVPR} and ACLNet~\cite{Zhang_2019_CVPR} 
\centering
\resizebox{\linewidth}{!}{\begin{tabular}{ l c c c c }
\toprule
 & \fsninee {\em SAL1:} & \fsninee {\em SAL2:} & \fsninee SDF           & \fsninee ODF+SDF  \\
 & \fsninee {\em MNL} & \fsninee {\em ACLNet}  & \fsninee {\em total}   & \fsninee total  \\
 &                   &                         & \fsninee (+Eq.~\eqref{eq:sal})  &\fsninee (+Eq.~\eqref{eq:sal}+SVD)  \\
\hline
{\em sec. per frame}   & 0.60		& 0.30  & 0.90 (+0.003) & 2.36 (+0.1)  \\
\hline
{\em s.p.c.} HMDB-51   & 55.7		& 27.9  &  83.6 (+0.3) & 219.2 (+0.8) \\
{\em s.p.c.} YUP++     & 83.2		& 41.6  & 124.8 (+0.4) & 327.3 (+1.2) \\
{\em s.p.c.} MPII      & 106.0	& 53.0  & 159.0 (+0.5) & 417.0 (+1.8) \\
{\em s.p.c.} Charades  & 180.3	& 90.1  & 270.4 (+0.9) & 709.1 (+3.5) \\
{\em s.p.c.} EPIC-Kitchens & 174.3 & 87.1 &  261.4 (+0.9) & 685.4 (+2.9)\\
{\em s.p.c.} Toyota Smarthome & 145.2 & 72.6 & 217.8 (+0.7) & 571.1 (+2.4) \\
% \hline
\bottomrule
\end{tabular}
%\vspace{0.02cm}
}% }
\caption{Statistics for the saliency detectors used in our experiments. The table presents timings such as seconds per frame (\emph{sec. per frame}) and seconds per clip (\emph{s.p.c.}) for detectors used by SDF. The total time incurred by the combined detector (\emph{SDF total}) is also provided. Additionally, we report the time taken for the descriptor in Eq.~\eqref{eq:sal} and all other SDF operations. Finally, the combined time for both ODF and SDF operations (\emph{SDF+ODF total}) is included.
}
\vspace{-0.3cm}
\label{tab:stat2}
\end{table}

Table \ref{tab:stat1} shows the timing for object detectors used by ODF descriptors during training. The detections from all four object detectors we use take approximately $1.47$ seconds per frame. Therefore, obtaining four ODF descriptors per clip (a uniquely annotated sequence for training or classification) takes between 136 and 441 seconds. Table \ref{tab:stat2} presents the timing for saliency detectors used in our SDF descriptors during training.  The detections for both saliency  detectors take around $0.9$ seconds per frame, with obtaining both SDF descriptors per clip taking between 84 and 271 seconds. 

It is important to note that the majority of the computational cost arises from the detectors rather than from the ODF and SDF descriptors, whose computational cost is minimal. Moreover, the idea of learning these computationally expensive representations during training proves highly valuable. While the total computation time per training clip ranges from 220 to 712 seconds, these representations are obtained virtually for free (in milliseconds) during testing, thanks to the DET1, $\cdots$, DET4 and SAL1/SAL2 units, as shown in Figure~\ref{fig:pipe}. Assuming that 25\% of clips in charades are used for testing, this results in a savings of 137 days of computation on a single GPU (or the equivalent of 1 day's savings on 137 GPUs). Given the 6\% improvement on Charades over the baseline (without ODF and SDF descriptors), coupled with these substantial computational savings, we believe these statistics highlight the value of our approach. 

\subsection{Limitations and Challenges}

\begin{table}[t]
\setlength{\tabcolsep}{0.3em}
\centering
\resizebox{\linewidth}{!}{\begin{tabular}{ l r r r r r r r}
\toprule
&  Video & IncV2 & IncResV2 & Res101 &  NASNet & MNL & ACLNet\\
\hline
HMDB-51 & 2.2 GB & 64.5 GB & 64.7 GB & 69.3 GB & 69.9 GB & 6.7 GB & 2.4 GB\\
YUP++ & 788.6 MB & 12.9 GB & 13.5 GB & 4.2 GB & 14.6 GB & 1.6 GB & 687.7 MB\\
MPII & 8.7 GB & 65.7 GB & 83.8 GB & 48.2 GB & 97.3 GB & 11.3 GB & 2.5 GB\\
Charades & 59.0 GB & 453.6 GB & 473.1 GB & 155.8 GB & 490.2 GB & 210.8 GB & 76.6 GB \\
\bottomrule
\end{tabular}% }
}
\caption{\lei{Storage statistics for original videos and extracted features. We present the storage sizes of raw object detection features extracted using Inception V2 (IncV2), Inception ResNet V2 (IncResV2), ResNet101 (Res101), and NASNet, as well as raw saliency detection features obtained from MNL and ACLNet. All extracted features are stored in HDF5 format, while the original videos are in formats such as AVI.}}
\label{tab:storage-stats}
\end{table}

\begin{table}[tbp]%htbp % left bottom right top
% \vspace{-0.3cm}
\setlength{\tabcolsep}{0.5em}
% \renewcommand{\arraystretch}{0.70}
%\fontsize{9}{9}\selectfont
\centering
\resizebox{\linewidth}{!}{\begin{tabular}{ l c c c c c }
\toprule
 & {\em no. of} & {\em av. frame} & {\em no. of} & {\em no. of} & {\em no. of} \\
 & {\em frames} & {\em count} & {\em  videos} & {\em  clips} & {\em classes} \\
\hline
HMDB-51  & 628635		& 92.91  & 6766 & 6766 & 51\\
YUP++		 & 166463		& 138.72 & 1200 & 1200 & 20 \\
MPII		 & 662394		& 176.73 & 44   & 3748 & 60 \\
Charades & 19978821 & 300.51 & 9848 & 66500 & 157\\
EPIC-Kitchens & $\sim$ 11.5M & 290.43 & 432 & 39596 & 149\\
Toyota Smarthome & 3.9M & 242.01 & 16115 & 16115 & 31\\
Kinetics-400 & 54M & 180 & 300K & 300K & 400\\
Kinetics-600 & 90M & 180 & 500K & 500K & 600\\
Something-Something V2 & 22M & 100 & 220K & 220K & 174\\
% \hline
\bottomrule
\end{tabular}
%\vspace{0.02cm}
}
\caption{Statistics of the datasets used in our experiments.
}
\vspace{-0.3cm}
\label{tab:datasets}
\end{table}

While our framework performs well on several benchmarks, it does face practical limitations, particularly in terms of computational costs, scalability, and sensitivity to parameter choices.

Computational complexity is a major challenge, especially during the training stage when extracting ground-truth feature descriptors, such as ODF and SDF (see Table~\ref{tab:stat1} and Table~\ref{tab:stat2}). These feature extraction processes can be computationally expensive and result in high storage consumption, particularly for large-scale datasets. \lei{For example, the original Charades dataset is 59.0 GB, whereas the extracted object detection and saliency detection features require a total of 1.54 TB and 287.4 GB of storage, respectively, which are approximately 26.7 and 4.87 times the size of the original videos (see Table \ref{tab:storage-stats}).} Table \ref{tab:datasets} presents basic statistics for the datasets used in our experiments. Notebly, Kinetics-400, Kinetics-600 and Something-Something V2 are among the largest datasets in our study, with approximately 54M, 90M, and 22M frames, respectively.

The framework's sensitivity to parameter choices also plays a crucial role in its effectiveness. For instance, the sketch size ($d'$) is key in determining both computational efficiency and the quality of feature representation. A small $d'$ can introduce noise, leading to poor approximations of the original feature set, which may negatively impact the model's performance.

Additionally, the framework faces challenges when applied to recent self-supervised pretraining frameworks such as VideoMAE V2 and InternVideo2. These models can use self-supervised learning techniques and are trained end-to-end, eliminating the need for manually extracted handcrafted features during training. However, these handcrafted features, although not required, still provide significant improvements to these models. The heavy computational cost and storage constraints of extracting the full set of handcrafted features make it infeasible to fully exploit their potential (see the total number of frames in Table~\ref{tab:datasets}), which in turn complicates training models with the complete set of semantically rich features. This issue limits the ability to optimize hallucination stream weights, making training more challenging.

Finally, robustness to noisy data is another potential limitation. In real-world scenarios, noisy or poorly extracted ground truth features can degrade performance, particularly in training the hallucination streams. This may lead to poor-quality hallucinated features during testing, further affecting the model's overall effectiveness.

\end{sloppypar}

\section{Conclusion}
\label{sec:concl}

\begin{sloppypar}

\lei{In this work, we introduced a novel multimodal action recognition framework that enhances recognition accuracy by integrating diverse auxiliary features while reducing reliance on computationally expensive handcrafted descriptors at inference. To guide the model toward action-relevant regions, we proposed two domain-specific descriptors: Object Detection Features (ODF), which capture contextual cues from multiple object detectors, and Saliency Detection Features (SDF), which emphasize spatial and intensity patterns critical for action understanding. To handle incomplete multimodal data, we developed a self-supervised hallucination mechanism that synthesizes missing cues at test time, enriching feature representations without increasing computational overhead. Furthermore, we incorporated aleatoric uncertainty modeling and a robust loss function to mitigate feature noise, improving the robustness of our model in fine-grained action recognition tasks. Our framework remains compatible with state-of-the-art architectures, including I3D, AssembleNet, Video Transformer Network, FASTER, and recent models such as VideoMAE V2 and InternVideo2. Extensive experiments on Kinetics-400, Kinetics-600, and Something-Something V2 confirm that our method achieves state-of-the-art performance, demonstrating its effectiveness in capturing fine-grained action dynamics and advancing multimodal action recognition.}

% We have introduced two simple yet effective object and saliency descriptors that perform self-supervision in an action recognition (AR) hallucination-based network. Our approach demonstrates that modeling higher-order statistical moments can lead to compact representations that self-supervise our AR pipeline.
% Additionally, We hallucinate skeleton and sound features to further enhance AR performance.
% Our findings align with recent multi-task learning research, which suggests that related tasks can co-supervise the main task. 
% We are the first to hallucinate object and saliency detection descriptors, as well as skeleton and sound features, resulting in clear improvements in accuracy and state-of-the-art performance on AR tasks such as Charades, EPIC-KITCHENS-55, and Toyota Smarthome. Furthermore, our method achieves strong results on large-scale datasets, including Kinetics-400, Kinetics-600, and Something-Something V2. 
% More importantly, we demonstrate that hallucinating object and saliency detections, along with skeleton and audio features, offers significant benefits even for state-of-the-art AR backbones such as AssembleNet/AssembleNet++, Video Transformer Network (VTN), lightweight FASTER, and the recent VideoMAE V2 and InternVideo2 models.

\end{sloppypar}

{
\noindent
\textbf{Acknowledgements.}
The authors thank CSIRO Scientific Computing for help. This work was also supported by the National Computational Merit Allocation Scheme 2024 (NCMAS 2024), with computational resources provided by NCI Australia, an NCRIS-enabled capability supported by the Australian Government. We sincerely thank the anonymous reviewers for their invaluable insights and constructive feedback, which have greatly contributed to improving our work.

% \noindent Code: \url{https://github.com/}.

\noindent\textbf{Data availability statement}: All datasets used and studied in this paper are publicly available.
}

\appendix
\begin{sloppypar}

\newcommand{\fsnine}[0]{\fontsize{9}{9}\selectfont}
\newcommand{\fsninee}[0]{\fontsize{9}{9}\selectfont}

% \input{ijcv24-revision/backgr}

% \begin{minipage}[c]{\linewidth}
\begin{figure*}[t]%htbp % left bottom right top
%\hspace{0.5cm}%
\centering%%%%
\vspace{-0.3cm}
%
% \comment{
% \begin{subfigure}[b]{0.245\linewidth}
% \centering\includegraphics[trim=0 0 0 0, clip=true,width=0.95\linewidth]{exp1-1.pdf}
% %\vspace{-0.2cm}
% \caption{\label{fig:str11} BoW FC (train)}
% %\captionof{subfigure}{aaa\label{fig:stra}}
% %\vspace{0.2cm}
% \end{subfigure}
% %
% \begin{subfigure}[b]{0.245\linewidth}
% \centering\includegraphics[trim=0 0 0 0, clip=true,width=0.95\linewidth]{exp1-2.pdf}
% %\vspace{-0.2cm}
% \caption{\label{fig:str12} BoW FC (test)}
% %\captionof{subfigure}{bbb\label{fig:strb}}
% %\vspace{0.2cm}
% \end{subfigure}
% %
% \begin{subfigure}[b]{0.245\linewidth}
% \centering\includegraphics[trim=0 0 0 0, clip=true,width=0.95\linewidth]{exp1-3.pdf}
% %\vspace{-0.2cm}
% \caption{\label{fig:str13} FV1 FC (train)}
% %\captionof{subfigure}{cccc\label{fig:strc}}
% %\vspace{0.2cm}
% \end{subfigure}
% %
% \begin{subfigure}[b]{0.245\linewidth}
% \centering\includegraphics[trim=0 0 0 0, clip=true,width=0.95\linewidth]{exp1-5.pdf}
% %\vspace{-0.2cm}
% \caption{\label{fig:str15} FV2 FC (train)}
% %\captionof{subfigure}{ddd\label{fig:strd}}
% %\vspace{0.2cm}
% \end{subfigure}
% }
%
%
%
%
%
%
\begin{subfigure}[b]{0.245\linewidth}
\centering\includegraphics[trim=0 0 0 0, clip=true,width=0.95\linewidth]{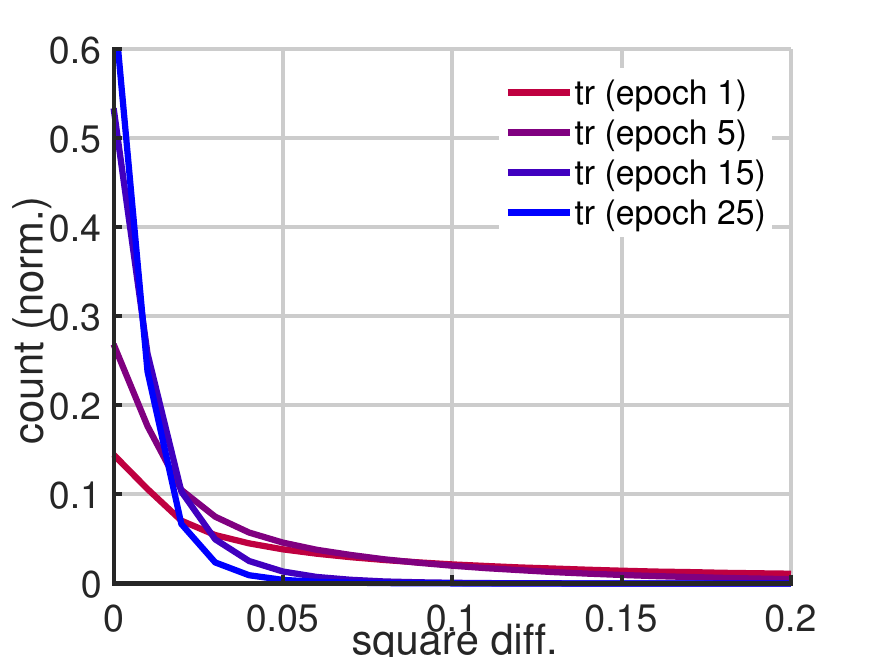}
%\vspace{-0.2cm}
\caption{\label{fig:str21} BoW FC (train)}
%\captionof{subfigure}{aaa\label{fig:stra}}
%\vspace{0.2cm}
\end{subfigure}
\begin{subfigure}[b]{0.245\linewidth}
\centering\includegraphics[trim=0 0 0 0, clip=true,width=0.95\linewidth]{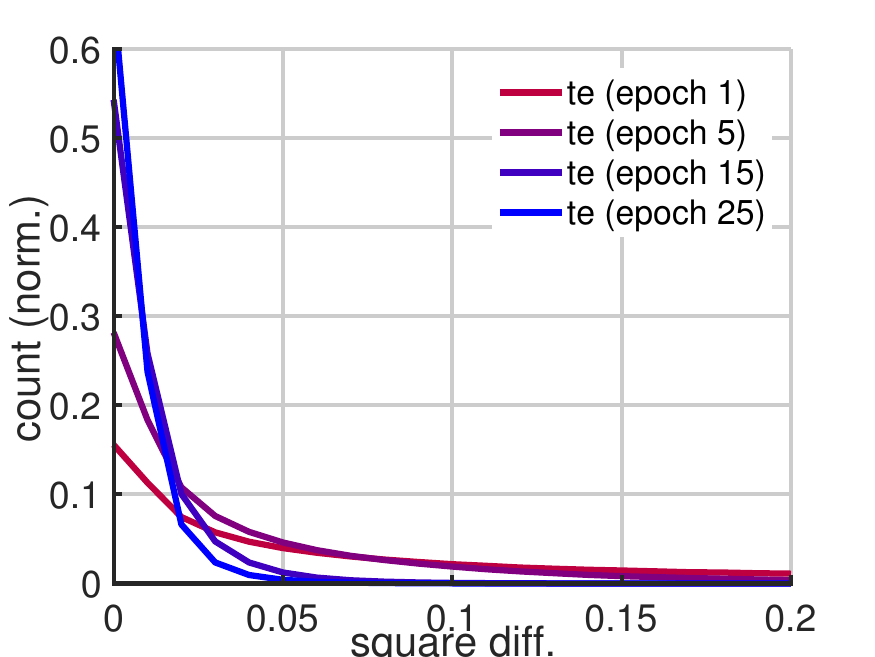}
%\vspace{-0.2cm}
\caption{\label{fig:str22} BoW FC (test)}
%\captionof{subfigure}{bbb\label{fig:strb}}
%\vspace{0.2cm}
\end{subfigure}
\begin{subfigure}[b]{0.245\linewidth}
\centering\includegraphics[trim=0 0 0 0, clip=true,width=0.95\linewidth]{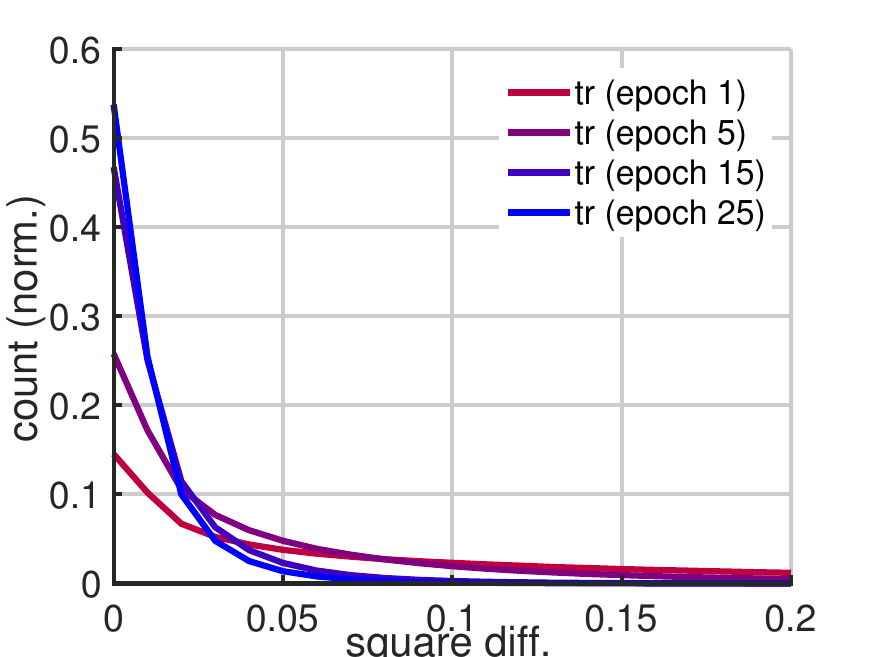}
%\vspace{-0.2cm}
\caption{\label{fig:str23} FV1 FC (train)}
%\captionof{subfigure}{cccc\label{fig:strc}}
%\vspace{0.2cm}
\end{subfigure}
\begin{subfigure}[b]{0.245\linewidth}
\centering\includegraphics[trim=0 0 0 0, clip=true,width=0.95\linewidth]{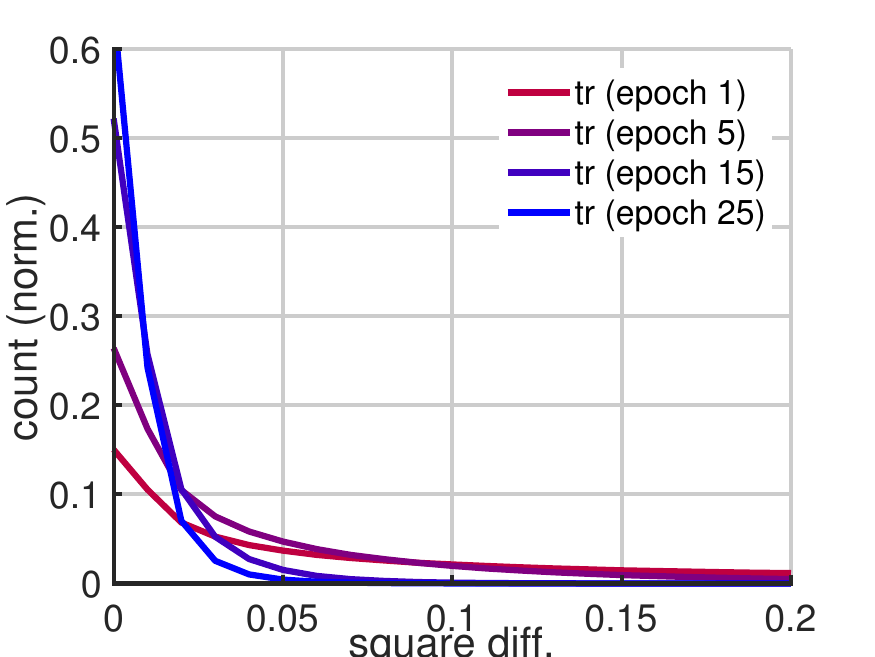}
%\vspace{-0.2cm}
\caption{\label{fig:str25} FV2 FC (train)}
%\captionof{subfigure}{ddd\label{fig:strd}}
%\vspace{0.2cm}
\end{subfigure}
\begin{subfigure}[b]{0.245\linewidth}
\centering\includegraphics[trim=0 0 0 0, clip=true,width=0.95\linewidth]{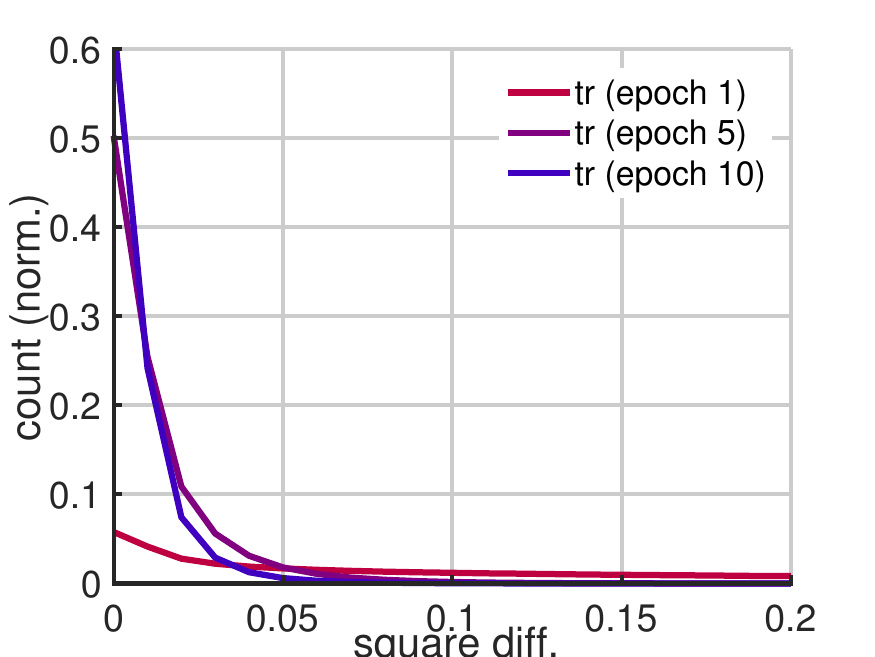}
%\vspace{-0.2cm}
\caption{\label{fig:str31} BoW Conv (train)}
%\captionof{subfigure}{aaa\label{fig:stra}}
%\vspace{0.2cm}
\end{subfigure}
\begin{subfigure}[b]{0.245\linewidth}
\centering\includegraphics[trim=0 0 0 0, clip=true,width=0.95\linewidth]{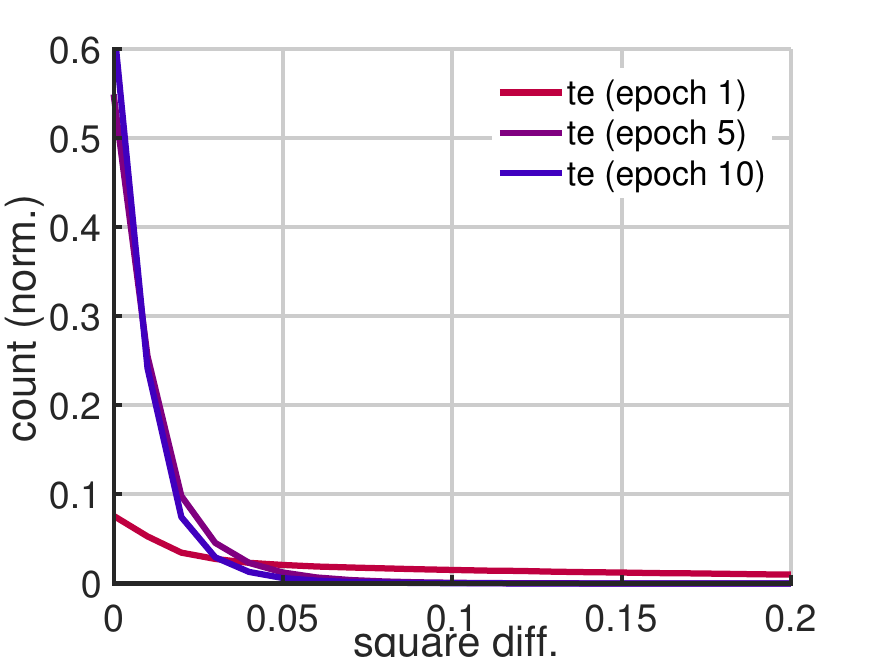}
%\vspace{-0.2cm}
\caption{\label{fig:str32} BoW Conv (test)}
%\captionof{subfigure}{bbb\label{fig:strb}}
%\vspace{0.2cm}
\end{subfigure}
\begin{subfigure}[b]{0.245\linewidth}
\centering\includegraphics[trim=0 0 0 0, clip=true,width=0.95\linewidth]{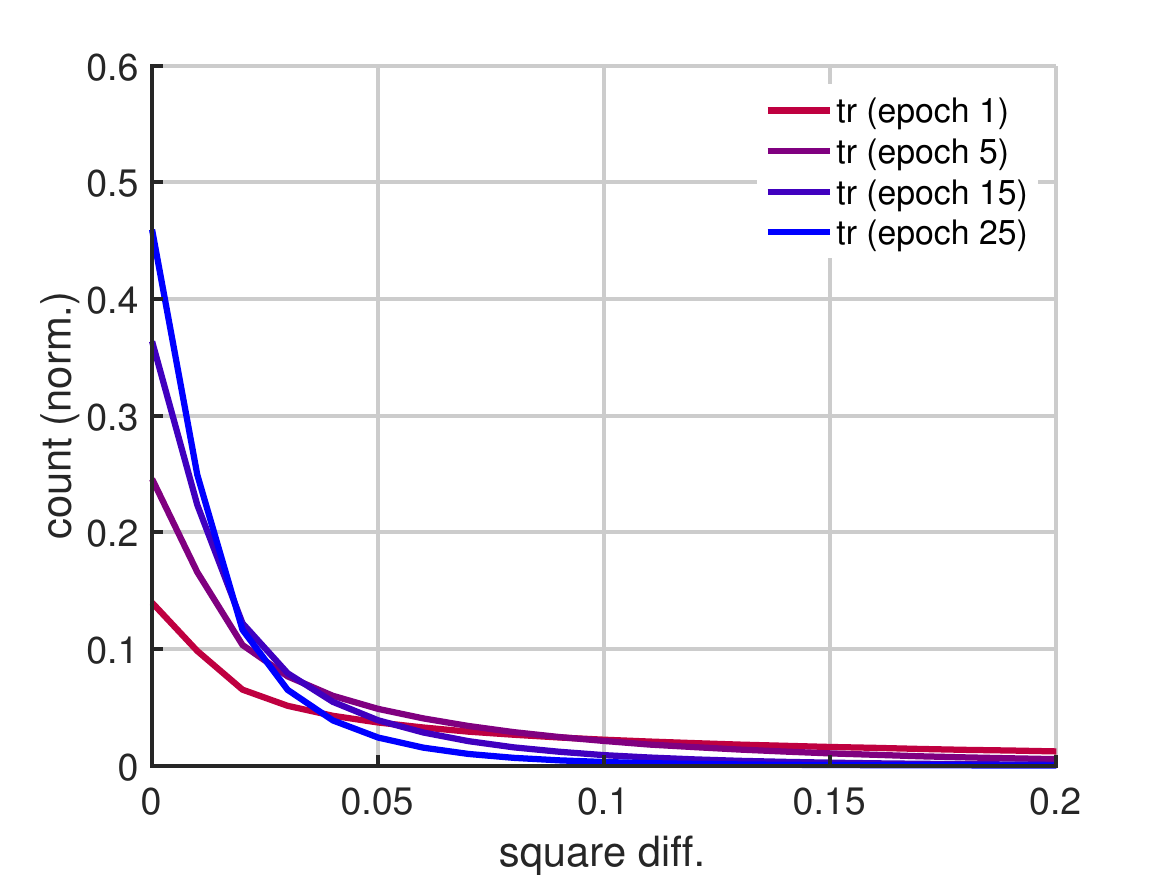}
%\vspace{-0.2cm}
\caption{\label{fig:str33} FV1 FC (train), -SK/PN}
%\captionof{subfigure}{cccc\label{fig:strc}}
%\vspace{0.2cm}
\end{subfigure}
\begin{subfigure}[b]{0.245\linewidth}
\centering\includegraphics[trim=0 0 0 0, clip=true,width=0.95\linewidth]{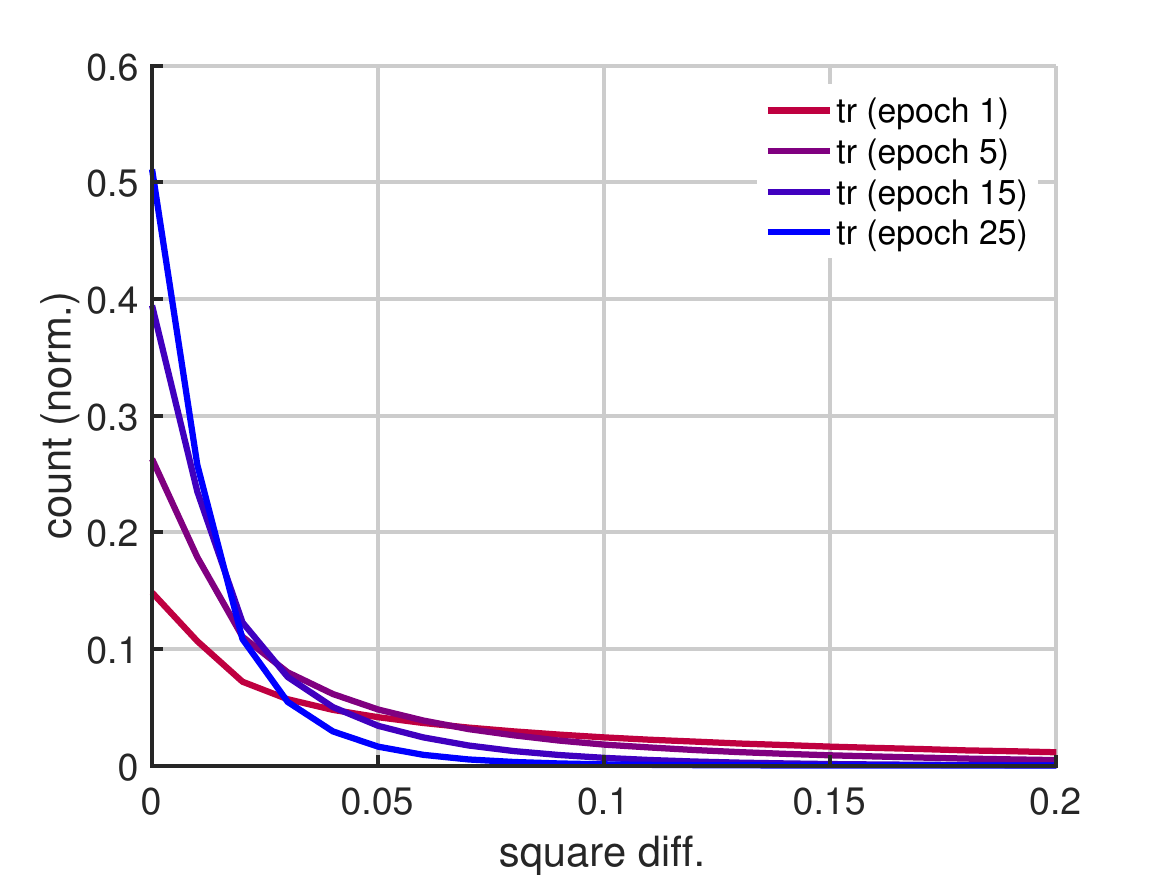}
%\vspace{-0.2cm}
\caption{\label{fig:str35} FV2 FC (train), -SK/PN}
%\captionof{subfigure}{ddd\label{fig:strd}}
%\vspace{0.2cm}
\end{subfigure}
\vspace{-0.3cm}
\caption{Evaluation of the squared difference between hallucinated and ground-truth representations on HMDB-51 (split 1). Experiments in the top row use ({\em FC}) streams with sketching and PN. The two leftmost plots in the bottom row use ({\em Conv}) streams, while the two rightmost plots in the bottom row examine ({\em FC}) streams without sketching or PN ({\em -SK/PN}).}
\vspace{0.4cm}
%\captionof{figure}{sfsd}
\label{fig:mse}
\end{figure*}
% \end{minipage}

\section{Hallucination Quality}
\label{sec:mse}

Below, we analyze the quality of hallucinated BoW and FV streams compared to their corresponding ground-truth feature vectors. 

Figure \ref{fig:mse} presents histograms of the squared differences between hallucinated features and ground-truth ones. Specifically, we plot histograms of $\{(\tilde{\psi}_{(bow),mn}-\psi_{(bow),mn})^2, m\!\in\!\idx{1000}, n\!\in\!\mathcal{N}\}$, where $m$ iterates over 1000 features and $n$ spans all videos in the dataset. For clarity, counts for training and testing splits are normalized by 1000 (the number of features) and the number of videos in each split, respectively. The histograms use bins of size $0.01$, creating smooth, continuous-like plots.

Figure \ref{fig:str21} demonstrates that during training, our BoW hallucination unit based on fully connected ({\em FC}) layers closely approximates the ground-truth BoW descriptors. Histograms at epochs 1, 5, 15, and 25 are shown with colors transitioning from red to blue. Early epochs exhibit a modest peak near the first bin, but as training progresses, this peak intensifies while subsequent bins diminish. This pattern reflects the reduction in approximation error over time.

Similarly, Figure \ref{fig:str22} illustrates that hallucinated BoW descriptors also closely approximate ground-truth descriptors in the testing split. Comparisons between testing and training histograms for BoW, as well as first- and second-order FV descriptors, reveal only minor differences. A ratio analysis of testing to training bins shows variations between $0.8\times$ and $1.25\times$. For clarity, we omit plots of FV testing split comparisons, as they align closely with training results.

Figures \ref{fig:str23} and \ref{fig:str25} show that the first- and second-order FV terms ({\em FV1} and {\em FV2}) are also well-learned by our hallucinating units. The results are displayed for the training split, as the testing behavior closely mirrors the training performance.

Finally, Figures \ref{fig:str31}, \ref{fig:str32}, \ref{fig:str33}, and \ref{fig:str35} highlight similar learning trends for BoW training and testing splits, as well as for the first- and second-order FV terms (training split only), using our hallucination unit based on {\em FC} layers without sketching or power normalization ({\em -SK/PN}).

\section{Visualization using UMAP}

\begin{figure*}[tbp]%htbp % left bottom right top
%\hspace{0.5cm}%
\centering%%%%
% \vspace{-0.3cm}
%
\begin{subfigure}[b]{0.48\linewidth}
\includegraphics[trim=0 0 0 0, clip=true,height=8cm]{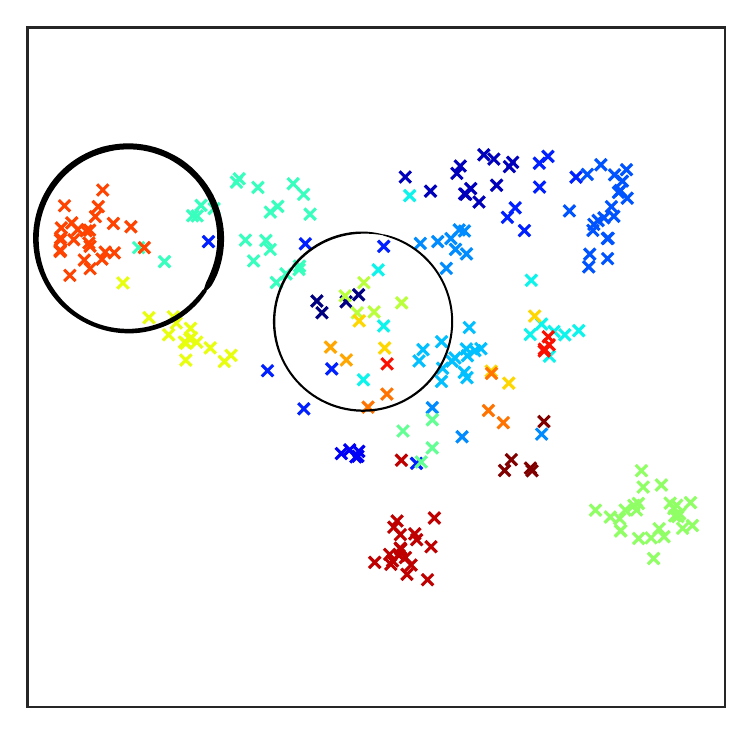}\vspace{-0.2cm}
\caption{\label{fig:yupva}}
\vspace{-0.2cm}
\end{subfigure}
\begin{subfigure}[b]{0.495\linewidth}
\includegraphics[trim=0 0 0 0, clip=true,height=8cm]{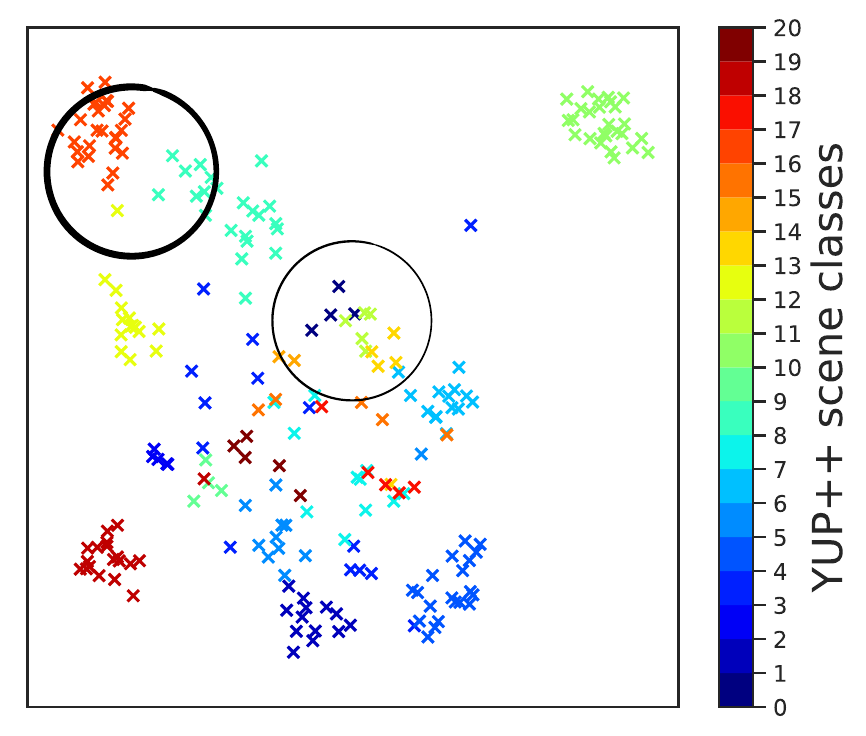}\vspace{-0.2cm}
\caption{\label{fig:yupvb}}
\vspace{-0.2cm}
\end{subfigure}
\caption{Visualization of the feature space (extracted from PredNet) for DEEP-HAL in Fig. \ref{fig:yupva} and DEEP-HAL+ODF in Fig. \ref{fig:yupvb} on the YUP++ dataset. For comparison, regions with notable differences are circled to highlight significant changes.
}
%\vspace{-0.3cm}
\label{fig:yupill}
\end{figure*}

\begin{figure*}[tbp]%htbp % left bottom right top
%\hspace{0.5cm}%
\centering%%%%
\vspace{0.1cm}
\begin{subfigure}[b]{0.48\linewidth}
\includegraphics[trim=0 0 0 0, clip=true,height=8cm]{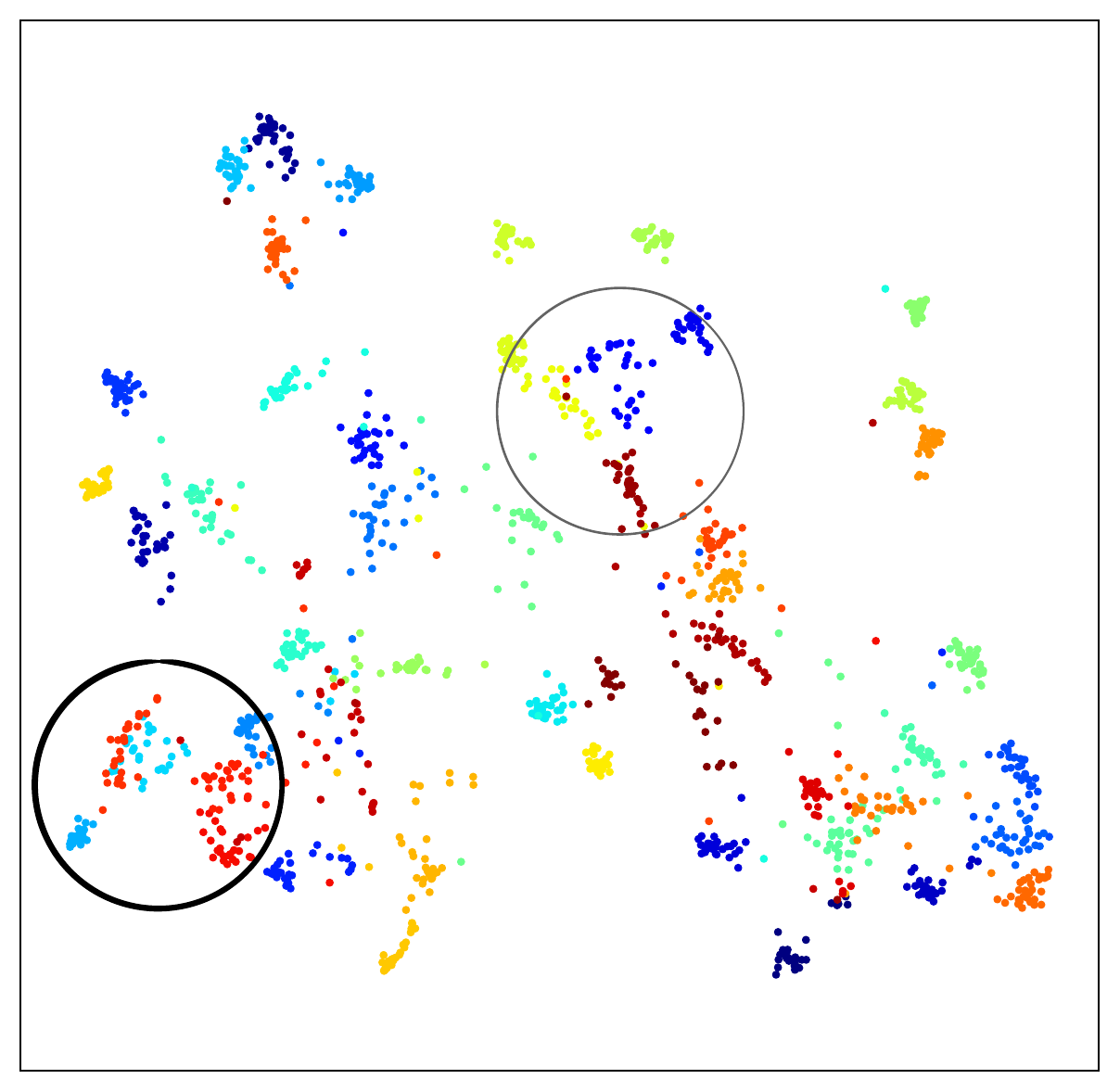}\vspace{-0.2cm}
\caption{\label{fig:hmdb51va}}
\vspace{-0.2cm}
\end{subfigure}
\begin{subfigure}[b]{0.495\linewidth}
\includegraphics[trim=0 0 0 0, clip=true,height=8cm]{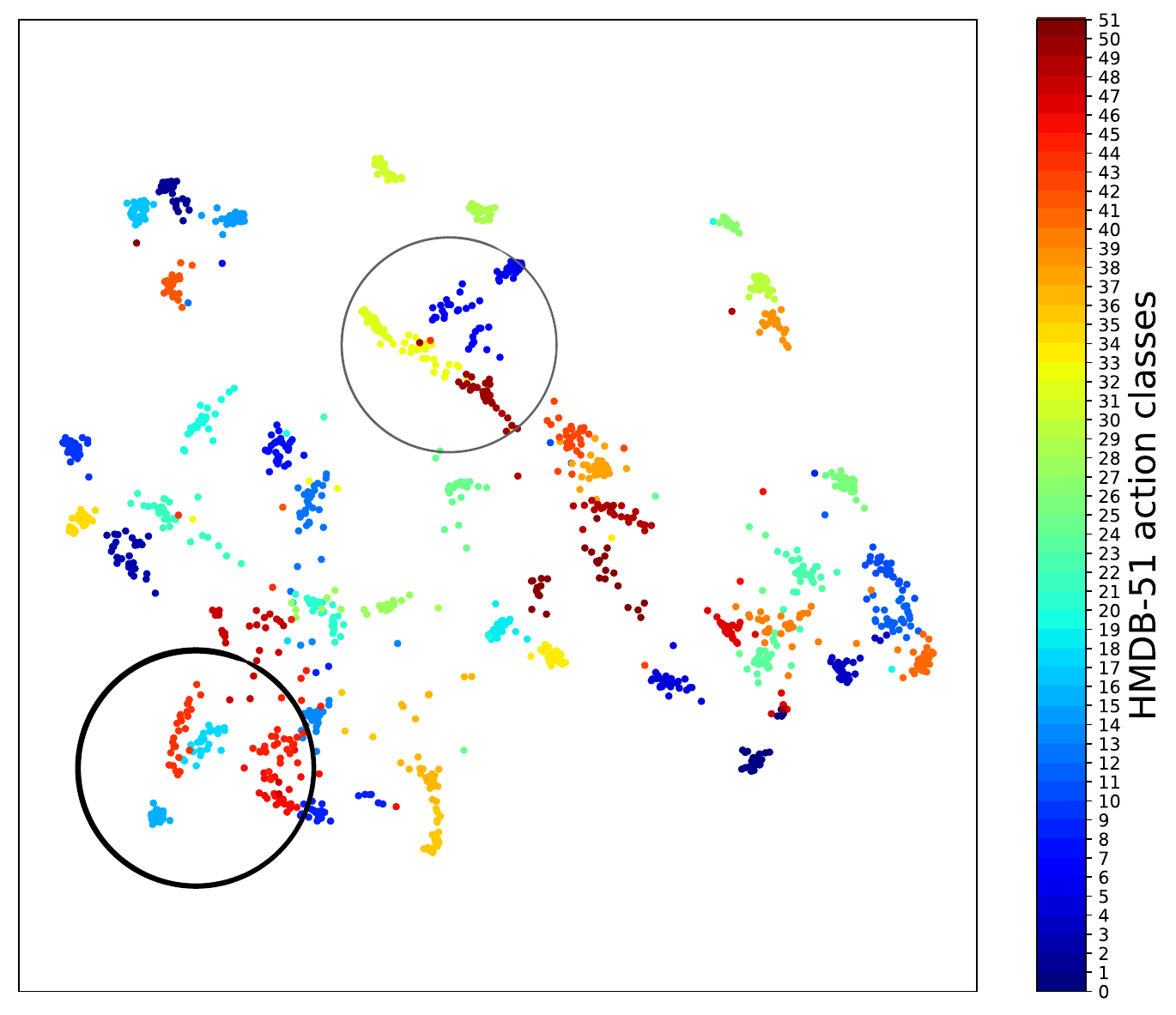}\vspace{-0.2cm}
\caption{\label{fig:hmdb51vb}}
\vspace{-0.2cm}
\end{subfigure}
\caption{Visualization of the feature space from PredNet for DEEP-HAL (Fig. \ref{fig:hmdb51va}) and DEEP-HAL+ODF (Fig. \ref{fig:hmdb51vb}) on the HMDB-51 dataset. Regions with notable differences are highlighted for comparison.
}
\vspace{-0.3cm}
\label{fig:hmdb51ill}
\end{figure*}

Figure \ref{fig:yupill} presents a UMAP \cite{umap} visualization of the YUP++ dataset. In Fig. \ref{fig:yupva}, the top-left corner shows samples from three classes, represented in red, green, and blue. These classes exhibit partial overlap in this representation. In contrast, Fig. \ref{fig:yupvb} depicts the same region, but the samples from the red, green, and blue classes are now more distinctly separated, indicating improved class-wise clustering.

Figure \ref{fig:hmdb51ill} illustrates a UMAP visualization of the HMDB-51 dataset. In Fig. \ref{fig:hmdb51va}, the bottom-left corner contains samples from two overlapping classes, shown in red and blue. However, in Fig. \ref{fig:hmdb51vb}, the samples from these two classes are better separated, and their respective clusters appear more clearly delineated, demonstrating improved class-wise organization in this visualization.

\section{Higher Resolution Frames on MPII}
\label{sec:mpii_bb}

\begin{table}[tbp]%htbp % left bottom right top
% \parbox{.99\linewidth}{
\setlength{\tabcolsep}{0.3em}
% \renewcommand{\arraystretch}{0.70}
%\fontsize{9}{9}\selectfont
% \hspace{-0.3cm}
\centering
\resizebox{\linewidth}{!}{\begin{tabular}{ c  c  c  c  c  c  c  c  c }
 & {\em sp1} & {\em sp2} & {\em sp3} & {\em sp4} & {\em sp5} & {\em sp6} & {\em sp7} & mAP \\
\hline
\kern-0.5em {\fontsize{8}{9}\selectfont HAF*+BoW halluc.}      		 & $78.8$ & $75.0$ & $84.1$ & $76.0$ & $77.0$ & $78.3$ & $75.2$ & $77.8$\\
\kern-0.5em {\fontsize{7.5}{9}\selectfont HAF*+BoW hal.+MSK/PN}   & $80.1$ & $79.2$ & $84.8$ & $83.9$ & $80.9$ & $78.5$ & $75.5$ & $80.4$\\
\hline
\kern-0.5em {\fontsize{8}{9}\selectfont HAF$^\bullet$+BoW halluc.}      		     & $78.8$ & $78.3$ & $84.2$ & $77.4$ & $77.1$ & $78.3$ & $75.2$ & $78.5$\\
\kern-0.5em {\fontsize{7.5}{9}\selectfont HAF$^\bullet$+BoW hal.+MSK/PN}   & $80.8$ & $80.9$ & $85.0$ & $83.9$ & $82.0$ & $79.8$ & $79.6$ & $\mathbf{81.7}$\\
\hline
\end{tabular}
\vspace{0.02cm}%\\
}
\caption{Evaluations on MPII. The ({\em HAF*+BoW halluc.}) represents our pipeline using the BoW stream, where ({\em *}) indicates human-centric pre-processing with a 256-pixel height resolution. The ({\em HAF*+BoW hal.+MSK/PN}) extends this pipeline by incorporating multiple sketches per BoW followed by Power Normalization ({\em PN}). Similarly, ({\em$^\bullet$}) denotes human-centric pre-processing with an increased 512-pixel height resolution.}
\vspace{-0.2cm}
\label{tab:mpiif2}
\end{table}

For the human-centric pre-processing applied to MPII, denoted by ({\em*}), we observe that the bounding boxes used for extracting human subjects are of low resolution. To address this, we first resize the RGB frames to 512 pixels in height (instead of 256 pixels), compute the corresponding optical flow, and then extract the human subjects. This adjustment effectively increases the resolution by a factor of $2\!\times$.

In Table \ref{tab:mpiif2}, results for ({\em HAF*+BoW halluc.}), our pipeline incorporating the BoW stream, and  ({\em HAF*+BoW hal.+MSK/PN}), which includes multiple sketches and PN, are shown for the standard 256 pixels height resolution. These results, denoted by ({\em*}), are taken from~\cite{Wang_2019_ICCV}.

The ({\em HAF$^\bullet$+BoW halluc.}), which also includes the BoW stream, and ({\em HAF$^\bullet$+BoW hal.+MSK/PN}) pipelines are analogous but computed with the increased 512-pixel height resolution, denoted by ({\em$^\bullet$}). As shown in the table, increasing the resolution by $2\!\times$ before human detection, extracting higher-resolution subjects, and then scaling them to a 256-pixel height for yields a 1.3\% improvement in accuracy.

\section{Data Pre-processing}
\label{sec:preproc}

For all video datasets, we apply a data augmentation strategy that includes random cropping of videos and left-right flips on both RGB and optical flow frames. During testing, we use center cropping and avoid flipping.

For the MPII dataset, which involves human-centric pre-processing, we first apply a human detector. Next, we randomly crop around the bounding box containing the human subject. This crop is included in the final sequence. We also allow scaling, zooming in, and left-right flipping. For longer videos, we sample sequences to create a 64-frame clip. For shorter videos (fewer than 64 frames), we repeat the sequence multiple times to match the expected input length. Finally, we scale the pixel values of both RGB and optical flow frames to the range between -1 and 1.
\end{sloppypar}
{\small
\bibliographystyle{spmpsci}
\bibliography{reference}

\begin{thebibliography}{100}
\providecommand{\url}[1]{{#1}}
\providecommand{\urlprefix}{URL }
\expandafter\ifx\csname urlstyle\endcsname\relax
  \providecommand{\doi}[1]{DOI~\discretionary{}{}{}#1}\else
  \providecommand{\doi}{DOI~\discretionary{}{}{}\begingroup
  \urlstyle{rm}\Url}\fi

\bibitem{4627014}
Abdullah, L.N., Noah, S.A.M.: Integrating audio visual data for human action
  detection.
\newblock In: 2008 Fifth International Conference on Computer Graphics, Imaging
  and Visualisation, pp. 242--246 (2008).
\newblock \doi{10.1109/CGIV.2008.65}

\bibitem{9420299}
Ahmad, T., Jin, L., Zhang, X., Lai, S., Tang, G., Lin, L.: Graph convolutional
  neural network for human action recognition: A comprehensive survey.
\newblock IEEE Transactions on Artificial Intelligence \textbf{2}(2), 128--145
  (2021).
\newblock \doi{10.1109/TAI.2021.3076974}

\bibitem{8834505}
Akbari, A., Jafari, R.: A deep learning assisted method for measuring
  uncertainty in activity recognition with wearable sensors.
\newblock In: 2019 IEEE EMBS International Conference on Biomedical Health
  Informatics (BHI), pp. 1--5 (2019).
\newblock \doi{10.1109/BHI.2019.8834505}

\bibitem{alwassel_2020_xdc}
Alwassel, H., Mahajan, D., Korbar, B., Torresani, L., Ghanem, B., Tran, D.:
  Self-supervised learning by cross-modal audio-video clustering.
\newblock In: Advances in Neural Information Processing Systems (NeurIPS)
  (2020)

\bibitem{Arnab_2021_ICCV}
Arnab, A., Dehghani, M., Heigold, G., Sun, C., Lu\v{c}i\'c, M., Schmid, C.:
  Vivit: A video vision transformer.
\newblock In: Proceedings of the IEEE/CVF International Conference on Computer
  Vision (ICCV), pp. 6836--6846 (2021)

\bibitem{soundnet}
Aytar, Y., Vondrick, C., Torralba, A.: Soundnet: Learning sound representations
  from unlabeled video.
\newblock In: Proceedings of the 30th International Conference on Neural
  Information Processing Systems, NIPS'16, p. 892–900. Curran Associates
  Inc., Red Hook, NY, USA (2016)

\bibitem{Baradel_2018_ECCV}
Baradel, F., Neverova, N., Wolf, C., Mille, J., Mori, G.: Object level visual
  reasoning in videos.
\newblock In: ECCV, pp. 1--16. Springer Science+Business Media, Munich, Germany
  (2018)

\bibitem{SalObjBenchmark_Tip2015}
Borji, A., Cheng, M.M., Jiang, H., Li, J.: Salient object detection: A
  benchmark.
\newblock TIP \textbf{24}(12), 5706--5722 (2015).
\newblock \doi{10.1109/TIP.2015.2487833}

\bibitem{seg_flow}
{Braux-Zin}, J., {Dupont}, R., {Bartoli}, A.: A general dense image matching
  framework combining direct and feature-based costs.
\newblock In: ICCV, pp. 185--192. IEEE, Sydney, NSW, Australia (2013)

\bibitem{brox_largedisp}
Brox, T., Malik, J.: Large displacement optical flow: Descriptor matching in
  variational motion estimation.
\newblock TPAMI \textbf{33}(3), 500--513 (2011).
\newblock \doi{10.1109/TPAMI.2010.143}.
\newblock \urlprefix\url{http://dx.doi.org/10.1109/TPAMI.2010.143}

\bibitem{bulat2021spacetime}
Bulat, A., Perez-Rua, J.M., Sudhakaran, S., Martinez, B., Tzimiropoulos, G.:
  Space-time mixing attention for video transformer.
\newblock In: A.~Beygelzimer, Y.~Dauphin, P.~Liang, J.W. Vaughan (eds.)
  Advances in Neural Information Processing Systems (2021).
\newblock \urlprefix\url{https://openreview.net/forum?id=QgX15Mdi1E_}

\bibitem{DBLP:journals/corr/BurdaGS15}
Burda, Y., Grosse, R.B., Salakhutdinov, R.: Importance weighted autoencoders.
\newblock In: Y.~Bengio, Y.~LeCun (eds.) 4th International Conference on
  Learning Representations, {ICLR} 2016, San Juan, Puerto Rico, May 2-4, 2016,
  Conference Track Proceedings (2016).
\newblock \urlprefix\url{http://arxiv.org/abs/1509.00519}

\bibitem{Cao_2017_CVPR}
Cao, Z., Simon, T., Wei, S.E., Sheikh, Y.: Realtime multi-person 2d pose
  estimation using part affinity fields.
\newblock In: The IEEE Conference on Computer Vision and Pattern Recognition
  (CVPR) (2017)

\bibitem{carreira2018short}
Carreira, J., Noland, E., Banki-Horvath, A., Hillier, C., Zisserman, A.: A
  short note about kinetics-600.
\newblock arXiv preprint arXiv:1808.01340  (2018)

\bibitem{i3d_net}
Carreira, J., Zisserman, A.: {Quo Vadis, Action Recognition? A New Model and
  the Kinetics Dataset}.
\newblock In: CVPR, pp. 1--10. IEEE, Honolulu, HI, USA (2018)

\bibitem{sstip}
Chakraborty, B., Holte, M.B., Moeslund, T.B., Gonz{\`{a}}lez, J.: Selective
  spatio-temporal interest points.
\newblock CVIU \textbf{116}(3), 396--410 (2012)

\bibitem{chen2020homm}
Chen, C., Fu, Z., Chen, Z., Jin, S., Cheng, Z., Jin, X., Hua, X.S.: Homm:
  Higher-order moment matching for unsupervised domain adaptation.
\newblock In: Proceedings of the AAAI conference on artificial intelligence,
  pp. 3422--3429 (2020)

\bibitem{chen2024motion}
Chen, Q., Wang, L., Koniusz, P., Gedeon, T.: Motion meets attention: Video
  motion prompts.
\newblock In: The 16th Asian Conference on Machine Learning (Conference Track)
  (2024)

\bibitem{Chen_Li_Yang_Li_Liu_2021}
Chen, Z., Li, S., Yang, B., Li, Q., Liu, H.: Multi-scale spatial temporal graph
  convolutional network for skeleton-based action recognition.
\newblock Proceedings of the AAAI Conference on Artificial Intelligence
  \textbf{35}(2), 1113--1122 (2021).
\newblock
  \urlprefix\url{https://ojs.aaai.org/index.php/AAAI/article/view/16197}

\bibitem{Cheng_2020_CVPR}
Cheng, K., Zhang, Y., He, X., Chen, W., Cheng, J., Lu, H.: Skeleton-based
  action recognition with shift graph convolutional network.
\newblock In: Proceedings of the IEEE/CVF Conference on Computer Vision and
  Pattern Recognition (CVPR) (2020)

\bibitem{9157077}
Cheng, K., Zhang, Y., He, X., Chen, W., Cheng, J., Lu, H.: Skeleton-based
  action recognition with shift graph convolutional network.
\newblock In: 2020 IEEE/CVF Conference on Computer Vision and Pattern
  Recognition (CVPR), pp. 180--189 (2020).
\newblock \doi{10.1109/CVPR42600.2020.00026}

\bibitem{anoop_generalized}
Cherian, A., Fernando, B., Harandi, M., Gould, S.: Generalized rank pooling for
  action recognition.
\newblock In: CVPR, pp. 3222--3231. IEEE, Honolulu, HI, USA (2017)

\bibitem{hok}
Cherian, A., Koniusz, P., Gould, S.: Higher-order pooling of {CNN} features via
  kernel linearization for action recognition.
\newblock In: WACV, pp. 130--138. IEEE, Santa Rosa, CA, USA (2017).
\newblock \doi{10.1109/WACV.2017.22}

\bibitem{anoop_rankpool_nonlin}
Cherian, A., Sra, S., Gould, S., Hartley, R.: Non-linear temporal subspace
  representations for activity recognition.
\newblock In: CVPR, pp. 2197--2206. IEEE, Salt Lake City, UT, USA (2018).
\newblock \doi{10.1109/CVPR.2018.00234}

\bibitem{Choi_2019_ICCV}
Choi, J., Chun, D., Kim, H., Lee, H.J.: Gaussian yolov3: An accurate and fast
  object detector using localization uncertainty for autonomous driving.
\newblock In: The IEEE International Conference on Computer Vision (ICCV)
  (2019)

\bibitem{Chollet_2017_CVPR}
Chollet, F.: Xception: Deep learning with depthwise separable convolutions.
\newblock In: Proceedings of the IEEE Conference on Computer Vision and Pattern
  Recognition (CVPR) (2017)

\bibitem{potion}
Choutas, V., Weinzaepfel, P., Revaud, J., Schmid, C.: {PoTion}: {P}ose motion
  representation for action recognition.
\newblock In: {CVPR}, pp. 7024--7033. IEEE, Salt Lake City, UT, USA (2018)

\bibitem{cormode_sketch}
Cormode, G., Hadjieleftheriou, M.: Finding frequent items in data streams.
\newblock Proc. VLDB Endow. \textbf{1}(2), 1530--1541 (2008).
\newblock \doi{10.14778/1454159.1454225}.
\newblock \urlprefix\url{http://dx.doi.org/10.14778/1454159.1454225}

\bibitem{csurka04_bovw}
Csurka, G., Dance, C.R., Fan, L., Willamowski, J., Bray, C.: Visual
  categorization with bags of keypoints.
\newblock In: ECCV Workshop, pp. 1--22. Springer Science+Business Media,
  Prague, Czech Republic (2004)

\bibitem{hof}
Dalal, N., Triggs, B., Schmid, C.: {Human Detection Using Oriented Histogram of
  Flow and Appearance}.
\newblock In: ECCV, pp. 428--441. Springer Science+Business Media, Graz,
  Austria (2006)

\bibitem{Damen_2018_ECCV}
Damen, D., Doughty, H., Farinella, G.M., Fidler, S., Furnari, A., Kazakos, E.,
  Moltisanti, D., Munro, J., Perrett, T., Price, W., Wray, M.: Scaling
  egocentric vision: The epic-kitchens dataset.
\newblock In: ECCV, pp. 1--17. Springer Science+Business Media, Munich, Germany
  (2018)

\bibitem{Das_2019_ICCV}
Das, S., Dai, R., Koperski, M., Minciullo, L., Garattoni, L., Bremond, F.,
  Francesca, G.: Toyota smarthome: Real-world activities of daily living.
\newblock In: Proceedings of the IEEE/CVF International Conference on Computer
  Vision (ICCV) (2019)

\bibitem{das2020vpn}
Das, S., Sharma, S., Dai, R., Bremond, F., Thonnat, M.: Vpn: Learning
  video-pose embedding for activities of daily living (2020)

\bibitem{5206848}
Deng, J., Dong, W., Socher, R., Li, L.J., Li, K., Fei-Fei, L.: Imagenet: A
  large-scale hierarchical image database.
\newblock In: 2009 IEEE Conference on Computer Vision and Pattern Recognition,
  pp. 248--255 (2009).
\newblock \doi{10.1109/CVPR.2009.5206848}

\bibitem{ding2025learnable}
Ding, D., Wang, L., Zhu, L., Gedeon, T., Koniusz, P.: Learnable expansion of
  graph operators for multi-modal feature fusion.
\newblock In: The Thirteenth International Conference on Learning
  Representations (2025).
\newblock \urlprefix\url{https://openreview.net/forum?id=SMZqIOSdlN}

\bibitem{10.1145/3701716.3717744}
Ding, X., Wang, L.: Do language models understand time?
\newblock WWW '25 Companion. Association for Computing Machinery, New York, NY,
  USA (2025).
\newblock \doi{10.1145/3701716.3717744}.
\newblock \urlprefix\url{https://doi.org/10.1145/3701716.3717744}

\bibitem{dingjourney}
Ding, X., Wang, L.: The journey of action recognition.
\newblock In: Companion Proceedings of the ACM Web Conference 2025, WWW '25
  Companion. Association for Computing Machinery, New York, NY, USA (2025).
\newblock \doi{10.1145/3701716.3717746}.
\newblock \urlprefix\url{https://doi.org/10.1145/3701716.3717746}

\bibitem{cuboid}
Doll{\'a}r, P., Rabaud, V., Cottrell, G., Belongie, S.: Behavior recognition
  via sparse spatio-temporal features.
\newblock In: ICCCN, pp. 65--72. IEEE, Honolulu, HI, USA (2005).
\newblock \urlprefix\url{http://dl.acm.org/citation.cfm?id=1259587.1259830}

\bibitem{cnn_lstm_ar}
Donahue, J., Hendricks, L.A., Guadarrama, S., Rohrbach, M., Venugopalan, S.,
  Darrell, T., Saenko, K.: Long-term recurrent convolutional networks for
  visual recognition and description.
\newblock In: CVPR, pp. 2625--2634. IEEE, Boston, MA, USA (2015)

\bibitem{8578672}
Dorta, G., Vicente, S., Agapito, L., Campbell, N.D.F., Simpson, I.: Structured
  uncertainty prediction networks.
\newblock In: 2018 IEEE/CVF Conference on Computer Vision and Pattern
  Recognition, pp. 5477--5485 (2018).
\newblock \doi{10.1109/CVPR.2018.00574}

\bibitem{dosovitskiy2021an}
Dosovitskiy, A., Beyer, L., Kolesnikov, A., Weissenborn, D., Zhai, X.,
  Unterthiner, T., Dehghani, M., Minderer, M., Heigold, G., Gelly, S.,
  Uszkoreit, J., Houlsby, N.: An image is worth 16x16 words: Transformers for
  image recognition at scale.
\newblock In: International Conference on Learning Representations (2021).
\newblock \urlprefix\url{https://openreview.net/forum?id=YicbFdNTTy}

\bibitem{Pengfei_ICCV19}
Fang, P., Zhou, J., Kumar~Roy, S., Petersson, L., Harandi, M.: Bilinear
  attention networks for person retrieval.
\newblock In: ICCV, pp. 8030--8039. IEEE, Seoul, Korea (2019)

\bibitem{slowfast}
Feichtenhofer, C., Fan, H., Malik, J., He, K.: Slowfast networks for video
  recognition.
\newblock In: ICCV, pp. 6202--6211. IEEE, Seoul, Korea (2019)

\bibitem{spat_temp_resnet}
Feichtenhofer, C., Pinz, A., Wildes, R.P.: Spatiotemporal residual networks for
  video action recognition.
\newblock In: NIPS, pp. 3468--3476. MIT Press, Barcelona, Spain (2016)

\bibitem{yuppp}
Feichtenhofer, C., Pinz, A., Wildes, R.P.: Temporal residual networks for
  dynamic scene recognition.
\newblock In: CVPR, pp. 4728--4737. IEEE, Honolulu, HI, USA (2017)

\bibitem{basura_rankpool}
Fernando, B., Gavves, E., M., J.O., Ghodrati, A., Tuytelaars, T.: Modeling
  video evolution for action recognition.
\newblock In: {CVPR}, pp. 5378--5387. IEEE, Boston, MA, USA (2015)

\bibitem{basura_rankpool2}
Fernando, B., Gould, S.: Learning end-to-end video classification with
  rank-pooling.
\newblock In: ICML, vol.~48, pp. 1187--1196. ACM, New York City, NY, USA (2016)

\bibitem{hog2d}
Freeman, W.T., Roth, M.: Orientation histograms for hand gesture recognition.
\newblock Tech. Rep. TR94-03, MERL - Mitsubishi Electric Research Laboratories,
  Cambridge, MA 02139 (1994).
\newblock \urlprefix\url{http://www.merl.com/publications/TR94-03/}

\bibitem{Gao_2020_CVPR}
Gao, R., Oh, T.H., Grauman, K., Torresani, L.: Listen to look: Action
  recognition by previewing audio.
\newblock In: IEEE/CVF Conference on Computer Vision and Pattern Recognition
  (CVPR) (2020)

\bibitem{soft_ass}
van Gemert, J.C., Veenman, C.J., Smeulders, A.W.M., Geusebroek, J.M.: Visual
  word ambiguity.
\newblock TPAMI \textbf{32}(7), 1271--1283 (2010).
\newblock \doi{10.1109/TPAMI.2009.132}.
\newblock \urlprefix\url{http://dx.doi.org/10.1109/TPAMI.2009.132}

\bibitem{Ghadiyaram_2019_CVPR}
Ghadiyaram, D., Tran, D., Mahajan, D.: Large-scale weakly-supervised
  pre-training for video action recognition.
\newblock In: CVPR, pp. 12046--12055. IEEE, Long Beach,California,USA (2019)

\bibitem{girdhar2023omnimae}
Girdhar, R., El-Nouby, A., Singh, M., Alwala, K.V., Joulin, A., Misra, I.:
  Omnimae: Single model masked pretraining on images and videos.
\newblock In: Proceedings of the IEEE/CVF conference on computer vision and
  pattern recognition, pp. 10406--10417 (2023)

\bibitem{fast_rcnn}
Girshick, R.: Fast r-cnn.
\newblock In: ICCV, pp. 1440--1448. IEEE, Santiago, Chile (2015)

\bibitem{rcnn}
Girshick, R., Donahue, J., Darrell, T., Malik, J.: Region-based convolutional
  networks for accurate object detection and segmentation.
\newblock TPAMI \textbf{38}(1), 142--158 (2016)

\bibitem{goyal2017something}
Goyal, R., Ebrahimi~Kahou, S., Michalski, V., Materzynska, J., Westphal, S.,
  Kim, H., Haenel, V., Fruend, I., Yianilos, P., Mueller-Freitag, M., et~al.:
  The" something something" video database for learning and evaluating visual
  common sense.
\newblock In: Proceedings of the IEEE international conference on computer
  vision, pp. 5842--5850 (2017)

\bibitem{Gu_2018_CVPR}
Gu, C., Sun, C., Ross, D.A., Vondrick, C., Pantofaru, C., Li, Y.,
  Vijayanarasimhan, S., Toderici, G., Ricco, S., Sukthankar, R., Schmid, C.,
  Malik, J.: Ava: A video dataset of spatio-temporally localized atomic visual
  actions.
\newblock In: CVPR, pp. 6047--6056. IEEE, Salt Lake City, UT, USA (2018)

\bibitem{Hadji_2018_ECCV}
Hadji, I., Wildes, R.P.: A new large scale dynamic texture dataset with
  application to {ConvNet} understanding.
\newblock In: ECCV. Springer Science+Business Media, Munich, Germany (2018)

\bibitem{mask_rnn}
He, K., Gkioxari, G., Doll{\'a}r, P., Girshick, R.B.: Mask r-cnn.
\newblock In: ICCV, pp. 2980--2988. IEEE, Venice, Italy (2017)

\bibitem{He_2016_CVPR}
He, K., Zhang, X., Ren, S., Sun, J.: Deep residual learning for image
  recognition.
\newblock In: CVPR, pp. 1--12. IEEE, Las Vegas, NV, USA (2016)

\bibitem{7780459}
He, K., Zhang, X., Ren, S., Sun, J.: Deep residual learning for image
  recognition.
\newblock In: 2016 IEEE Conference on Computer Vision and Pattern Recognition
  (CVPR), pp. 770--778 (2016).
\newblock \doi{10.1109/CVPR.2016.90}

\bibitem{klloss}
He, Y., Zhu, C., Wang, J., Savvides, M., Zhang, X.: Bounding box regression
  with uncertainty for accurate object detection.
\newblock In: The IEEE Conference on Computer Vision and Pattern Recognition
  (CVPR) (2019)

\bibitem{flow_def2}
Horn, B.K.P., Schunck, B.G.: Determining optical flow.
\newblock Artificial Intelligence \textbf{17}, 185--203 (1981)

\bibitem{ChengCVPR17}
Hou, Q., Cheng, M.M., Hu, X., Borji, A., Tu, Z., Torr, P.H.S.: Deeply
  supervised salient object detection with short connections.
\newblock In: Proc. IEEE Conf. Comp. Vis. Patt. Recogn., pp. 3203--3212. IEEE,
  Honolulu, HI, USA (2017)

\bibitem{9008835}
Howard, A., Sandler, M., Chen, B., Wang, W., Chen, L., Tan, M., Chu, G.,
  Vasudevan, V., Zhu, Y., Pang, R., Adam, H., Le, Q.: Searching for
  mobilenetv3.
\newblock In: 2019 IEEE/CVF International Conference on Computer Vision (ICCV),
  pp. 1314--1324. IEEE Computer Society, Los Alamitos, CA, USA (2019).
\newblock \doi{10.1109/ICCV.2019.00140}.
\newblock
  \urlprefix\url{https://doi.ieeecomputersociety.org/10.1109/ICCV.2019.00140}

\bibitem{Huang_Huang_Ouyang_Wang_2020}
Huang, L., Huang, Y., Ouyang, W., Wang, L.: Part-level graph convolutional
  network for skeleton-based action recognition.
\newblock Proceedings of the AAAI Conference on Artificial Intelligence
  \textbf{34}(07), 11045--11052 (2020).
\newblock \doi{10.1609/aaai.v34i07.6759}.
\newblock \urlprefix\url{https://ojs.aaai.org/index.php/AAAI/article/view/6759}

\bibitem{uncertainty4}
H{\"{u}}llermeier, E., Waegeman, W.: Aleatoric and epistemic uncertainty in
  machine learning: an introduction to concepts and methods.
\newblock Mach. Learn. \textbf{110}(3), 457--506 (2021).
\newblock \doi{10.1007/s10994-021-05946-3}.
\newblock \urlprefix\url{https://doi.org/10.1007/s10994-021-05946-3}

\bibitem{aistats-uncert}
Huo, Z., Pakbin, A., Chen, X., Hurley, N.C., Yuan, Y., Qian, X., Wang, Z.,
  Huang, S., Mortazavi, B.: Uncertainty quantification for deep context-aware
  mobile activity recognition and unknown context discovery.
\newblock In: AISTATS, pp. 3894--3904 (2020).
\newblock \urlprefix\url{http://proceedings.mlr.press/v108/huo20a.html}

\bibitem{squeezenet}
Iandola, F.N., Moskewicz, M.W., Ashraf, K., Han, S., Dally, W.J., Keutzer, K.:
  Squeezenet: Alexnet-level accuracy with 50x fewer parameters and
  {\textless}1mb model size.
\newblock CoRR \textbf{abs/1602.07360} (2016).
\newblock \urlprefix\url{http://arxiv.org/abs/1602.07360}

\bibitem{uncertainty3}
Indrayan, A.: Medical biostatistics, 2nd ed. edn.
\newblock Chapman \& Hall/CRC,, Boca Raton : (c2008.).
\newblock
  \urlprefix\url{http://www.loc.gov/catdir/toc/ecip0723/2007030353.html}

\bibitem{jebara_prodkers}
Jebara, T., Kondor, R., Howard, A.: Probability product kernels.
\newblock JMLR \textbf{5}, 819--844 (2004)

\bibitem{jegou_bursts}
J\'egou, H., Douze, M., Schmid, C.: {On the Burstiness of Visual Elements}.
\newblock In: CVPR, pp. 1169--1176. IEEE, Long Beach,alifornia, USA (2009)

\bibitem{cnn3d_ar}
Ji, S., Xu, W., Yang, M., Yu, K.: {3D} convolutional neural networks for human
  action recognition.
\newblock TPAMI \textbf{35}, 221--231 (2010)

\bibitem{cnn_basic_ar}
Karpathy, A., Toderici, G., Shetty, S., Leung, T., Sukthankar, R., Fei-Fei, L.:
  Large-scale video classification with convolutional neural networks.
\newblock In: CVPR, pp. 1725--1732. IEEE, Columbus, OH, USA (2014).
\newblock \doi{10.1109/CVPR.2014.223}.
\newblock \urlprefix\url{https://doi.org/10.1109/CVPR.2014.223}

\bibitem{kay2017kinetics}
Kay, W., Carreira, J., Simonyan, K., Zhang, B., Hillier, C., Vijayanarasimhan,
  S., Viola, F., Green, T., Back, T., Natsev, P., et~al.: The kinetics human
  action video dataset.
\newblock arXiv preprint arXiv:1705.06950  (2017)

\bibitem{uncertainty5}
Kendall, A., Gal, Y.: What uncertainties do we need in bayesian deep learning
  for computer vision?
\newblock In: I.~Guyon, U.V. Luxburg, S.~Bengio, H.~Wallach, R.~Fergus,
  S.~Vishwanathan, R.~Garnett (eds.) Advances in Neural Information Processing
  Systems, vol.~30. Curran Associates, Inc. (2017).
\newblock
  \urlprefix\url{https://proceedings.neurips.cc/paper/2017/file/2650d6089a6d640c5e85b2b88265dc2b-Paper.pdf}

\bibitem{Kendall_2018_CVPR}
Kendall, A., Gal, Y., Cipolla, R.: Multi-task learning using uncertainty to
  weigh losses for scene geometry and semantics.
\newblock In: Proceedings of the IEEE Conference on Computer Vision and Pattern
  Recognition (CVPR) (2018)

\bibitem{DBLP:journals/corr/KingmaW13}
Kingma, D.P., Welling, M.: Auto-encoding variational bayes.
\newblock In: Y.~Bengio, Y.~LeCun (eds.) 2nd International Conference on
  Learning Representations, {ICLR} 2014, Banff, AB, Canada, April 14-16, 2014,
  Conference Track Proceedings (2014).
\newblock \urlprefix\url{http://arxiv.org/abs/1312.6114}

\bibitem{kipf2017semi}
Kipf, T.N., Welling, M.: Semi-supervised classification with graph
  convolutional networks.
\newblock In: International Conference on Learning Representations (ICLR)
  (2017)

\bibitem{uncertainty2}
Kiureghian, A.D., Ditlevsen, O.: Aleatory or epistemic? does it matter?
\newblock Structural Safety \textbf{31}(2), 105--112 (2009).
\newblock \doi{https://doi.org/10.1016/j.strusafe.2008.06.020}.
\newblock
  \urlprefix\url{https://www.sciencedirect.com/science/article/pii/S0167473008000556}.
\newblock Risk Acceptance and Risk Communication

\bibitem{3D-HOG}
Kl{\"a}ser, A., Marszalek, M., Schmid, C.: {A Spatio-Temporal Descriptor Based
  on 3D-Gradients}.
\newblock In: BMCV, pp. 1--10. BMVA, Leeds, UK (2008)

\bibitem{kondratyuk2021movinets}
Kondratyuk, D., Yuan, L., Li, Y., Zhang, L., Tan, M., Brown, M., Gong, B.:
  Movinets: Mobile video networks for efficient video recognition.
\newblock In: Proceedings of the IEEE/CVF conference on computer vision and
  pattern recognition, pp. 16020--16030 (2021)

\bibitem{me_tensor_eccv16}
Koniusz, P., Cherian, A., Porikli, F.: Tensor representations via kernel
  linearization for action recognition from {3D} skeletons.
\newblock In: ECCV, pp. 1--14. Springer Science+Business Media, Amsterdam, The
  Netherlands (2016)

\bibitem{me_SAO}
Koniusz, P., Mikolajczyk, K.: {Soft Assignment of Visual Words as Linear
  Coordinate Coding and Optimisation of its Reconstruction Error}.
\newblock In: ICIP, pp. 2461--2464. IEEE, Brussels, Belgium (2011)

\bibitem{kon_tpami2020b}
Koniusz, P., Wang, L., Cherian, A.: Tensor representations for action
  recognition.
\newblock In: IEEE Transactions on Pattern Analysis and Machine Intelligence.
  IEEE (2020)

\bibitem{me_tensor_tech_rep}
Koniusz, P., Yan, F., Gosselin, P.H., Mikolajczyk, K.: {Higher-order Occurrence
  Pooling on Mid- and Low-level Features: Visual Concept Detection}.
\newblock Technical Report \textbf{1}(1), 1--20 (2013)

\bibitem{me_tensor}
Koniusz, P., Yan, F., Gosselin, P.H., Mikolajczyk, K.: Higher-order occurrence
  pooling for bags-of-words: {V}isual concept detection.
\newblock TPAMI \textbf{39}(2), 313--326 (2017)

\bibitem{me_ATN}
Koniusz, P., Yan, F., Mikolajczyk, K.: {Comparison of Mid-Level Feature Coding
  Approaches And Pooling Strategies in Visual Concept Detection}.
\newblock CVIU \textbf{117}, 479--492 (2012).
\newblock \doi{10.1016/j.cviu.2012.10.010}

\bibitem{kon_tpami2020a}
Koniusz, P., Zhang, H.: Power normalizations in fine-grained image, few-shot
  image and graph classification.
\newblock In: IEEE Transactions on Pattern Analysis and Machine Intelligence.
  IEEE (2020)

\bibitem{me_deeper}
Koniusz, P., Zhang, H., Porikli, F.: A deeper look at power normalizations.
\newblock In: CVPR, pp. 5774--5783. IEEE, Salt Lake City, UT, USA (2018)

\bibitem{10.1007/978-3-030-58565-5_45}
Korban, M., Li, X.: Ddgcn: A dynamic directed graph convolutional network for
  action recognition.
\newblock In: A.~Vedaldi, H.~Bischof, T.~Brox, J.M. Frahm (eds.) Computer
  Vision -- ECCV 2020, pp. 761--776. Springer International Publishing, Cham
  (2020)

\bibitem{10.1145/3341105.3373906}
Kozlov, A., Andronov, V., Gritsenko, Y.: Lightweight Network Architecture for
  Real-Time Action Recognition, p. 2074–2080.
\newblock Association for Computing Machinery, New York, NY, USA (2020).
\newblock \urlprefix\url{https://doi.org/10.1145/3341105.3373906}

\bibitem{kuehne2011hmdb}
Kuehne, H., Jhuang, H., Garrote, E., Poggio, T., Serre, T.: {HMDB}: {A} large
  video database for human motion recognition.
\newblock In: ICCV, pp. 2556--2563. IEEE, Barcelona, Spain (2011)

\bibitem{acmmm20_KumarKSXS20}
Kumar, D., Kumar, C., Seah, C., Xia, S., Shao, M.: Finding achilles' heel:
  Adversarial attack on multi-modal action recognition.
\newblock In: C.W. Chen, R.~Cucchiara, X.~Hua, G.~Qi, E.~Ricci, Z.~Zhang,
  R.~Zimmermann (eds.) MM, pp. 3829--3837. {ACM}, Seattle, United States
  (2020).
\newblock \doi{10.1145/3394171.3413531}.
\newblock \urlprefix\url{https://doi.org/10.1145/3394171.3413531}

\bibitem{harris3d}
Laptev, I.: On space-time interest points.
\newblock IJCV \textbf{64}(2-3), 107--123 (2005).
\newblock \doi{10.1007/s11263-005-1838-7}.
\newblock \urlprefix\url{http://dx.doi.org/10.1007/s11263-005-1838-7}

\bibitem{mv-stip}
Li, C., Su, B., Wang, J., Zhang, Q.: Human action recognition using
  multi-velocity {STIPs} and motion energy orientation histogram.
\newblock J. Inf. Sci. Eng. \textbf{30}, 295--312 (2014)

\bibitem{acmmm20_2020A}
Li, J., Wei, P., Zhang, Y., Zheng, N.: A slow-i-fast-p architecture for
  compressed video action recognition.
\newblock In: MM, pp. 2039--2047. ACM, Seattle, United States (2020)

\bibitem{Li_2019_CVPR}
Li, M., Chen, S., Chen, X., Zhang, Y., Wang, Y., Tian, Q.: Actional-structural
  graph convolutional networks for skeleton-based action recognition.
\newblock In: The IEEE Conference on Computer Vision and Pattern Recognition
  (CVPR) (2019)

\bibitem{Lin_eccv2014_coco}
Lin, T.Y., Maire, M., Belongie, S., Hays, J., Perona, P., Ramanan, D.,
  Doll{\'a}r, P., Zitnick, C.L.: Microsoft coco: Common objects in context.
\newblock In: D.~Fleet, T.~Pajdla, B.~Schiele, T.~Tuytelaars (eds.) Computer
  Vision -- ECCV 2014, pp. 740--755. Springer International Publishing, Cham
  (2014)

\bibitem{liu_sadefense}
Lingqiao, L., Wang, L., Liu, X.: {In Defence of Soft-assignment Coding}.
\newblock In: ICCV, pp. 2486--2493. IEEE, Barcelona, Spain (2011)

\bibitem{acmmm19_LiuGQWL19}
Liu, Z., Gao, G., Qin, A.K., Wu, T., Liu, C.H.: Action recognition with
  bootstrapping based long-range temporal context attention.
\newblock In: L.~Amsaleg, B.~Huet, M.A. Larson, G.~Gravier, H.~Hung, C.~Ngo,
  W.T. Ooi (eds.) MM, pp. 583--591. {ACM}, Nice, France (2019).
\newblock \doi{10.1145/3343031.3350916}.
\newblock \urlprefix\url{https://doi.org/10.1145/3343031.3350916}

\bibitem{liu2021video}
Liu, Z., Ning, J., Cao, Y., Wei, Y., Zhang, Z., Lin, S., Hu, H.: Video swin
  transformer.
\newblock arXiv preprint arXiv:2106.13230  (2021)

\bibitem{DBLP:journals/corr/abs-2112-06183}
Lu, C., Koniusz, P.: Few-shot keypoint detection with uncertainty learning for
  unseen species.
\newblock CoRR \textbf{abs/2112.06183} (2021).
\newblock \urlprefix\url{https://arxiv.org/abs/2112.06183}

\bibitem{Ma_2018_ECCV}
Ma, N., Zhang, X., Zheng, H.T., Sun, J.: Shufflenet v2: Practical guidelines
  for efficient cnn architecture design.
\newblock In: Proceedings of the European Conference on Computer Vision (ECCV)
  (2018)

\bibitem{ckn}
Mairal, J., Koniusz, P., Harchaoui, Z., Schmid, C.: Convolutional kernel
  networks.
\newblock In: NIPS, pp. 1--9. MIT Press, Montreal, Quebec, Canada (2014)

\bibitem{uncertainty1}
Matthies, H.G.: Quantifying uncertainty: Modern computational representation of
  probability and applications.
\newblock In: A.~Ibrahimbegovic, I.~Kozar (eds.) Extreme Man-Made and Natural
  Hazards in Dynamics of Structures, pp. 105--135. Springer Netherlands,
  Dordrecht (2007)

\bibitem{umap}
McInnes, L., Healy, J., Saul, N., Grossberger, L.: Umap: Uniform manifold
  approximation and projection.
\newblock The Journal of Open Source Software \textbf{3}(29), 861 (2018)

\bibitem{9607406}
Neimark, D., Bar, O., Zohar, M., Asselmann, D.: Video transformer network.
\newblock In: 2021 IEEE/CVF International Conference on Computer Vision
  Workshops (ICCVW), pp. 3156--3165 (2021).
\newblock \doi{10.1109/ICCVW54120.2021.00355}

\bibitem{multisensory2018}
Owens, A., Efros, A.A.: Audio-visual scene analysis with self-supervised
  multisensory features.
\newblock arXiv preprint arXiv:1804.03641  (2018)

\bibitem{UHAR_BMVC2021}
Paoletti, G., Cavazza, J., Beyan, C., Del~Bue, A.: {Unsupervised Human Action
  Recognition with Skeletal Graph Laplacian and Self-Supervised Viewpoints
  Invariance}.
\newblock In: The 32nd British Machine Vision Conference (BMVC) (2021)

\bibitem{brox_accurate}
Papenberg, N., Bruhn, A., Brox, T., Didas, S., Weickert, J.: Highly accurate
  optic flow computation with theoretically justified warping.
\newblock IJCV \textbf{67}, 141--158 (2006)

\bibitem{patrick2021keeping}
Patrick, M., Campbell, D., Asano, Y., Misra, I., Metze, F., Feichtenhofer, C.,
  Vedaldi, A., Henriques, J.F.: Keeping your eye on the ball: Trajectory
  attention in video transformers.
\newblock Advances in neural information processing systems \textbf{34},
  12493--12506 (2021)

\bibitem{perronnin_fisher}
Perronnin, F., Dance, C.: Fisher kernels on visual vocabularies for image
  categorization.
\newblock In: CVPR, vol.~0, pp. 1--8. IEEE, Minneapolis, Minnesota, USA (2007)

\bibitem{perronnin_fisherimpr}
Perronnin, F., S\'anchez, J., Mensink, T.: {Improving the Fisher Kernel for
  Large-Scale Image Classification}.
\newblock In: ECCV, pp. 143--156. Springer Science+Business Media, Heraklion,
  Crete (2010)

\bibitem{pham_sketch}
Pham, N., Pagh, R.: Fast and scalable polynomial kernels via explicit feature
  maps.
\newblock In: ACM SIGKDD, pp. 239--247. ACM, Chicago, USA (2013).
\newblock \doi{10.1145/2487575.2487591}.
\newblock \urlprefix\url{http://doi.acm.org/10.1145/2487575.2487591}

\bibitem{Piergiovanni_2019_ICCV}
Piergiovanni, A., Angelova, A., Toshev, A., Ryoo, M.S.: Evolving space-time
  neural architectures for videos.
\newblock In: ICCV, pp. 1793--1802. IEEE, Seoul, Korea (2019)

\bibitem{piergiovanni2023rethinking}
Piergiovanni, A., Kuo, W., Angelova, A.: Rethinking video vits: Sparse video
  tubes for joint image and video learning.
\newblock In: Proceedings of the IEEE/CVF Conference on Computer Vision and
  Pattern Recognition, pp. 2214--2224 (2023)

\bibitem{Piergiovanni_2021_CVPR}
Piergiovanni, A., Ryoo, M.S.: Recognizing actions in videos from unseen
  viewpoints.
\newblock In: Proceedings of the IEEE/CVF Conference on Computer Vision and
  Pattern Recognition (CVPR), pp. 4124--4132 (2021)

\bibitem{9895208}
Qin, Z., Liu, Y., Ji, P., Kim, D., Wang, L., McKay, R.I., Anwar, S., Gedeon,
  T.: Fusing higher-order features in graph neural networks for skeleton-based
  action recognition.
\newblock IEEE Transactions on Neural Networks and Learning Systems
  \textbf{35}(4), 4783--4797 (2024).
\newblock \doi{10.1109/TNNLS.2022.3201518}

\bibitem{yolo}
Redmon, J., Divvala, S., Girshick, R., Farhadi, A.: You only look once:
  Unified, real-time object detection.
\newblock In: CVPR, pp. 779--788. IEEE, Boston, MA, USA (2015)

\bibitem{faster-rcnn}
Ren, S., He, K., Girshick, R., Sun, J.: Faster {R-CNN}: {T}owards real-time
  object detection with region proposal networks.
\newblock In: NIPS, pp. 91--99. MIT Press, Montreal, Canada (2015)

\bibitem{epic_flow}
Revaud, J., Weinzaepfel, P., Harchaoui, Z., Schmid, C.: {EpicFlow:
  Edge-Preserving Interpolation of Correspondences for Optical Flow}.
\newblock In: CVPR, pp. 1164--1172. IEEE, Boston, MA, USA (2015)

\bibitem{rohrbach2012database}
Rohrbach, M., Amin, S., Andriluka, M., Schiele, B.: A database for fine grained
  activity detection of cooking activities.
\newblock In: CVPR, pp. 1194--1201. IEEE, Providence, Rhode Island (2012)

\bibitem{ILSVRC15}
Russakovsky, O., Deng, J., Su, H., Krause, J., Satheesh, S., Ma, S., Huang, Z.,
  Karpathy, A., Khosla, A., Bernstein, M., Berg, A.C., Fei-Fei, L.: {ImageNet}
  large scale visual recognition challenge.
\newblock IJCV \textbf{115}(3), 211--252 (2015).
\newblock \doi{10.1007/s11263-015-0816-y}

\bibitem{assemblenet_plus}
Ryoo, M.S., Piergiovanni, A., Kangaspunta, J., Angelova, A.: Assemblenet++:
  Assembling modality representations via attention connections.
\newblock In: ECCV, pp. 1--19. Springer Science+Business Media, Glasgow, UK
  (2020)

\bibitem{assemblenet}
Ryoo, M.S., Piergiovanni, A., Tan, M., Angelova, A.: Assemblenet: Searching for
  multi-stream neural connectivity in video architectures.
\newblock In: ICLR, pp. 1--15. ICLR, Addis Ababa, Ethiopia (2020)

\bibitem{Sandler_2018_CVPR}
Sandler, M., Howard, A., Zhu, M., Zhmoginov, A., Chen, L.C.: Mobilenetv2:
  Inverted residuals and linear bottlenecks.
\newblock In: The IEEE Conference on Computer Vision and Pattern Recognition
  (CVPR) (2018)

\bibitem{sift_3d}
Scovanner, P., Ali, S., Shah, M.: {A 3-Dimentional SIFT Descriptor and its
  Application to Action Recognition}.
\newblock In: MM, pp. 357--356. ACM, Augsburg, Germany (2007)

\bibitem{10.1145/3412841.3441974}
Seo, Y.M., Choi, Y.S.: Graph Convolutional Networks for Skeleton-Based Action
  Recognition with LSTM Using Tool-Information, p. 986–993.
\newblock Association for Computing Machinery, New York, NY, USA (2021).
\newblock \urlprefix\url{https://doi.org/10.1145/3412841.3441974}

\bibitem{8953648}
Shi, L., Zhang, Y., Cheng, J., Lu, H.: Two-stream adaptive graph convolutional
  networks for skeleton-based action recognition.
\newblock In: 2019 IEEE/CVF Conference on Computer Vision and Pattern
  Recognition (CVPR), pp. 12018--12027 (2019).
\newblock \doi{10.1109/CVPR.2019.01230}

\bibitem{2sagcn2019cvpr}
Shi, L., Zhang, Y., Cheng, J., Lu, H.: Two-stream adaptive graph convolutional
  networks for skeleton-based action recognition.
\newblock In: CVPR (2019)

\bibitem{Shotton2011}
Shotton, J., Fitzgibbon, A., Cook, M., Sharp, T., Finocchio, M., Moore, R.,
  Kipman, A., Blake, A.: {Real-Time Human Pose Recognition in Parts from Single
  Depth Images}.
\newblock In: CVPR, pp. 1297--1304 (2011)

\bibitem{sigurdsson2016hollywood}
Sigurdsson, G.A., Varol, G., Wang, X., Farhadi, A., Laptev, I., Gupta, A.:
  Hollywood in homes: Crowdsourcing data collection for activity understanding.
\newblock In: ECCV, pp. 1--17. Springer Science+Business Media, Amsterdam, The
  Netherlands (2016)

\bibitem{two_stream}
Simonyan, K., Zisserman, A.: Two-stream convolutional networks for action
  recognition in videos.
\newblock In: NIPS, pp. 568--576. MIT Press, Montreal, Quebec, Canada (2014)

\bibitem{sivic_vq}
Sivic, J., Zisserman, A.: {Video Google}: {A} text retrieval approach to object
  matching in videos.
\newblock ICCV \textbf{2}, 1470--1477 (2003)

\bibitem{10655590}
Srivastava, S., Sharma, G.: Omnivec2 - a novel transformer based network for
  large scale multimodal and multitask learning.
\newblock In: 2024 IEEE/CVF Conference on Computer Vision and Pattern
  Recognition (CVPR), pp. 27402--27414. IEEE Computer Society, Los Alamitos,
  CA, USA (2024).
\newblock \doi{10.1109/CVPR52733.2024.02588}.
\newblock
  \urlprefix\url{https://doi.ieeecomputersociety.org/10.1109/CVPR52733.2024.02588}

\bibitem{6343802}
Stork, J.A., Spinello, L., Silva, J., Arras, K.O.: Audio-based human activity
  recognition using non-markovian ensemble voting.
\newblock In: 2012 IEEE RO-MAN: The 21st IEEE International Symposium on Robot
  and Human Interactive Communication, pp. 509--514 (2012).
\newblock \doi{10.1109/ROMAN.2012.6343802}

\bibitem{Subedar_2019_ICCV}
Subedar, M., Krishnan, R., Meyer, P.L., Tickoo, O., Huang, J.:
  Uncertainty-aware audiovisual activity recognition using deep bayesian
  variational inference.
\newblock In: Proceedings of the IEEE/CVF International Conference on Computer
  Vision (ICCV) (2019)

\bibitem{Szegedy_2017_AAAI}
Szegedy, C., Ioffe, S., Vanhoucke, V., Alemi, A.A.: Inception-v4,
  inception-resnet and the impact of residual connections on learning.
\newblock In: AAAI, pp. 4278--4284. AAAI Press, San Francisco,CA,USA (2017).
\newblock \urlprefix\url{http://dl.acm.org/citation.cfm?id=3298023.3298188}

\bibitem{Szegedy_2016_CVPR}
Szegedy, C., Vanhoucke, V., Ioffe, S., Shlens, J., Wojna, Z.: Rethinking the
  inception architecture for computer vision.
\newblock In: CVPR, pp. 2818--2826. IEEE, Las Vegas, NV, USA (2016)

\bibitem{pmlr-v97-tan19a}
Tan, M., Le, Q.: {E}fficient{N}et: Rethinking model scaling for convolutional
  neural networks.
\newblock In: K.~Chaudhuri, R.~Salakhutdinov (eds.) Proceedings of the 36th
  International Conference on Machine Learning, \emph{Proceedings of Machine
  Learning Research}, vol.~97, pp. 6105--6114. PMLR (2019).
\newblock \urlprefix\url{https://proceedings.mlr.press/v97/tan19a.html}

\bibitem{tencent_hall}
Tang, Y., Ma, L., Zhou, L.: Hallucinating optical flow features for video
  classification.
\newblock In: IJCAI, pp. 926--932. IJCAI, Macao, China (2019)

\bibitem{tomar2006converting}
Tomar, S.: Converting video formats with ffmpeg.
\newblock Linux Journal \textbf{2006}(146), 10 (2006)

\bibitem{tong2022videomae}
Tong, Z., Song, Y., Wang, J., Wang, L.: Videomae: Masked autoencoders are
  data-efficient learners for self-supervised video pre-training.
\newblock Advances in neural information processing systems \textbf{35},
  10078--10093 (2022)

\bibitem{spattemp_filters}
Tran, D., Bourdev, L., Fergus, R., Torresani, L., Paluri, M.: {Learning
  Spatiotemporal Features with {3D} Convolutional Networks}.
\newblock In: ICCV, pp. 4489--4497. IEEE, Santiago, Chile (2015)

\bibitem{tran2018closer}
Tran, D., Wang, H., Torresani, L., Ray, J., LeCun, Y., Paluri, M.: A closer
  look at spatiotemporal convolutions for action recognition.
\newblock In: Proceedings of the IEEE conference on Computer Vision and Pattern
  Recognition, pp. 6450--6459 (2018)

\bibitem{hof2}
Uijlings, J.R., Duta, I.C., Rostamzadeh, N., Sebe, N.: {Realtime Video
  Classification using Dense HOF/HOG}.
\newblock In: ICMR, pp. 145--152. ACM, New York, NY, USA (2014)

\bibitem{long_term_ar}
{Varol}, G., {Laptev}, I., {Schmid}, C.: Long-term temporal convolutions for
  action recognition.
\newblock TPAMI \textbf{40}(6), 1510--1517 (2018)

\bibitem{dense_traj}
Wang, H., Kl{\"a}ser, A., Schmid, C., Cheng-Lin, L.: {Action Recognition by
  Dense Trajectories}.
\newblock In: CVPR, pp. 3169--3176. IEEE, Colorado Springs, CO, USA (2011)

\bibitem{dense_mot_boundary}
Wang, H., Kl{\"a}ser, A., Schmid, C., Liu, C.L.: {Dense Trajectories and Motion
  Boundary Descriptors for Action Recognition}.
\newblock IJCV \textbf{103}, 60--79 (2013)

\bibitem{improved_traj}
Wang, H., Schmid, C.: {Action Recognition with Improved Trajectories}.
\newblock In: ICCV, pp. 3551--3558. IEEE, Sydney, Australia (2013)

\bibitem{anoop_advers}
Wang, J., Cherian, A.: Learning discriminative video representations using
  adversarial perturbations.
\newblock In: ECCV, pp. 716--733. Springer Science+Business Media, Munich,
  Germany (2018).
\newblock \doi{10.1007/978-3-030-01225-0\_42}.
\newblock \urlprefix\url{https://doi.org/10.1007/978-3-030-01225-0\_42}

\bibitem{lei_thesis_2017}
Wang, L.: Analysis and evaluation of {K}inect-based action recognition
  algorithms.
\newblock Master's thesis, School of the Computer Science and Software
  Engineering, The University of Western Australia (2017)

\bibitem{wang2023robust}
Wang, L.: Robust human action modelling.
\newblock Ph.D. thesis, The Australian National University (Australia) (2023)

\bibitem{wang2023videomae}
Wang, L., Huang, B., Zhao, Z., Tong, Z., He, Y., Wang, Y., Wang, Y., Qiao, Y.:
  Videomae v2: Scaling video masked autoencoders with dual masking.
\newblock In: Proceedings of the IEEE/CVF Conference on Computer Vision and
  Pattern Recognition, pp. 14549--14560 (2023)

\bibitem{lei_tip_2019}
Wang, L., Huynh, D.Q., Koniusz, P.: A comparative review of recent kinect-based
  action recognition algorithms.
\newblock TIP \textbf{29}(1), 15--28 (2019).
\newblock \doi{10.1109/TIP.2019.2925285}

\bibitem{lei_icip_2019}
Wang, L., Huynh, D.Q., Mansour, M.R.: Loss switching fusion with similarity
  search for video classification.
\newblock In: IEEE ICIP, pp. 974--978 (2019).
\newblock \doi{10.1109/ICIP.2019.8803051}

\bibitem{lei_mm21}
Wang, L., Koniusz, P.: Self-Supervising Action Recognition by Statistical
  Moment and Subspace Descriptors, p. 4324–4333.
\newblock Association for Computing Machinery, New York, NY, USA (2021).
\newblock \urlprefix\url{https://doi.org/10.1145/3474085.3475572}

\bibitem{wang2022temporal}
Wang, L., Koniusz, P.: Temporal-viewpoint transportation plan for skeletal
  few-shot action recognition.
\newblock In: Proceedings of the Asian Conference on Computer Vision, pp.
  4176--4193 (2022)

\bibitem{wang2022uncertainty}
Wang, L., Koniusz, P.: Uncertainty-dtw for time series and sequences.
\newblock In: European Conference on Computer Vision, pp. 176--195. Springer
  (2022)

\bibitem{wang20233mformer}
Wang, L., Koniusz, P.: 3mformer: Multi-order multi-mode transformer for
  skeletal action recognition.
\newblock In: Proceedings of the IEEE/CVF Conference on Computer Vision and
  Pattern Recognition, pp. 5620--5631 (2023)

\bibitem{wang2024flow}
Wang, L., Koniusz, P.: Flow dynamics correction for action recognition.
\newblock In: ICASSP 2024-2024 IEEE International Conference on Acoustics,
  Speech and Signal Processing (ICASSP), pp. 3795--3799. IEEE (2024)

\bibitem{Wang_2019_ICCV}
Wang, L., Koniusz, P., Huynh, D.Q.: Hallucinating {IDT} descriptors and {I3D}
  optical flow features for action recognition with cnns.
\newblock In: ICCV, pp. 8697--8707. IEEE, Seoul, Korea (2019)

\bibitem{wang20213d}
Wang, L., Liu, J., Koniusz, P.: 3d skeleton-based few-shot action recognition
  with jeanie is not so na\" ive.
\newblock arXiv preprint arXiv:2112.12668  (2021)

\bibitem{wang2024meet}
Wang, L., Liu, J., Zheng, L., Gedeon, T., Koniusz, P.: Meet jeanie: a
  similarity measure for 3d skeleton sequences via temporal-viewpoint
  alignment.
\newblock International Journal of Computer Vision pp. 1--32 (2024)

\bibitem{wang2024high}
Wang, L., Sun, K., Koniusz, P.: High-order tensor pooling with attention for
  action recognition.
\newblock In: ICASSP 2024-2024 IEEE International Conference on Acoustics,
  Speech and Signal Processing (ICASSP), pp. 3885--3889. IEEE (2024)

\bibitem{RFCN}
Wang, L., Wang, L., Lu, H., Zhang, P., Ruan, X.: Saliency detection with
  recurrent fully convolutional networks.
\newblock In: ECCV, pp. 825--841. Springer Science+Business Media, Amsterdam,
  The Netherlands (2016).
\newblock \doi{10.1007/978-3-319-46493-0_50}

\bibitem{wangtaylor}
Wang, L., Yuan, X., Gedeon, T., Zheng, L.: Taylor videos for action
  recognition.
\newblock In: Forty-first International Conference on Machine Learning (2024)

\bibitem{DBLP:journals/corr/abs-2109-08472}
Wang, M., Xing, J., Liu, Y.: Actionclip: {A} new paradigm for video action
  recognition.
\newblock CoRR \textbf{abs/2109.08472} (2021).
\newblock \urlprefix\url{https://arxiv.org/abs/2109.08472}

\bibitem{10.1145/3389189.3389196}
Wang, W., Seraj, F., Havinga, P.J.M.: A sound-based crowd activity recognition
  with neural network based regression models.
\newblock In: Proceedings of the 13th ACM International Conference on PErvasive
  Technologies Related to Assistive Environments, PETRA '20. Association for
  Computing Machinery, New York, NY, USA (2020).
\newblock \doi{10.1145/3389189.3389196}.
\newblock \urlprefix\url{https://doi.org/10.1145/3389189.3389196}

\bibitem{wang2024internvideo2}
Wang, Y., Li, K., Li, X., Yu, J., He, Y., Chen, G., Pei, B., Zheng, R., Xu, J.,
  Wang, Z., et~al.: Internvideo2: Scaling video foundation models for
  multimodal video understanding.
\newblock ECCV  (2024)

\bibitem{weinberger_sketch}
Weinberger, K., Dasgupta, A., Langford, J., Smola, A., Attenberg, J.: Feature
  hashing for large scale multitask learning.
\newblock In: ICML, pp. 1113--1120. ACM, Montreal, Canada (2009).
\newblock \doi{10.1145/1553374.1553516}.
\newblock \urlprefix\url{http://doi.acm.org/10.1145/1553374.1553516}

\bibitem{deep_flow}
Weinzaepfel, P., Revaud, J., Harchaoui, Z., Schmid, C.: {DeepFlow: Large
  displacement optical flow with deep matching}.
\newblock In: ICCV, pp. 1--8. IEEE, Sydney, NSW, Australia (2013).
\newblock \urlprefix\url{http://hal.inria.fr/hal-00873592}

\bibitem{hes-stip}
Willems, G., Tuytelaars, T., Gool, L.V.: An efficient dense and scale-invariant
  spatio-temporal interest point detector.
\newblock In: ECCV, pp. 650--663. Springer Science+Business Media, Marseille,
  France (2008).
\newblock \doi{10.1007/978-3-540-88688-4_48}.
\newblock \urlprefix\url{https://doi.org/10.1007/978-3-540-88688-4_48}

\bibitem{Wu_2019_CVPR}
Wu, C.Y., Feichtenhofer, C., Fan, H., He, K., Krahenbuhl, P., Girshick, R.:
  Long-term feature banks for detailed video understanding.
\newblock In: CVPR, pp. 284--293. IEEE, Long Beach, California, USA (2019)

\bibitem{5692578}
Wu, Q., Wang, Z., Deng, F., Feng, D.D.: Realistic human action recognition with
  audio context.
\newblock In: 2010 International Conference on Digital Image Computing:
  Techniques and Applications, pp. 288--293 (2010).
\newblock \doi{10.1109/DICTA.2010.57}

\bibitem{Xie_2018_ECCV}
Xie, S., Sun, C., Huang, J., Tu, Z., Murphy, K.: Rethinking spatiotemporal
  feature learning: Speed-accuracy trade-offs in video classification.
\newblock In: Proceedings of the European Conference on Computer Vision (ECCV)
  (2018)

\bibitem{Yan_2019_CVPR}
Yan, A., Wang, Y., Li, Z., Qiao, Y.: {PA3D}: Pose-action {3D} machine for video
  recognition.
\newblock In: CVPR, pp. 7922--7931. IEEE, Long Beach, California, USA (2019)

\bibitem{yan2022multiview}
Yan, S., Xiong, X., Arnab, A., Lu, Z., Zhang, M., Sun, C., Schmid, C.:
  Multiview transformers for video recognition.
\newblock In: Proceedings of the IEEE/CVF conference on computer vision and
  pattern recognition, pp. 3333--3343 (2022)

\bibitem{stgcn2018aaai}
Yan, S., Xiong, Y., Lin, D.: {Spatial Temporal Graph Convolutional Networks for
  Skeleton-Based Action Recognition}.
\newblock In: AAAI (2018)

\bibitem{yang2021unik}
Yang, D., Wang, Y., Dantcheva, A., Garattoni, L., Francesca, G., Bremond, F.:
  Unik: A unified framework for real-world skeleton-based action recognition.
\newblock BMVC  (2021)

\bibitem{yao2023side4video}
Yao, H., Wu, W., Li, Z.: Side4video: Spatial-temporal side network for
  memory-efficient image-to-video transfer learning.
\newblock arXiv preprint arXiv:2311.15769  (2023)

\bibitem{LTP}
Yeffet, L., Wolf, L.: Local trinary patterns for human action recognition.
\newblock In: ICCV, pp. 492--497. IEEE, Seoul, Korea (2009)

\bibitem{acmmm19_ZhangZCG19}
Zhang, C., Zou, Y., Chen, G., Gan, L.: {PAN:} persistent appearance network
  with an efficient motion cue for fast action recognition.
\newblock In: L.~Amsaleg, B.~Huet, M.A. Larson, G.~Gravier, H.~Hung, C.~Ngo,
  W.T. Ooi (eds.) MM, pp. 500--509. {ACM}, Nice, France (2019).
\newblock \doi{10.1145/3343031.3350876}.
\newblock \urlprefix\url{https://doi.org/10.1145/3343031.3350876}

\bibitem{Zhang_2019_CVPR}
Zhang, H., Zhang, J., Koniusz, P.: Few-shot learning via saliency-guided
  hallucination of samples.
\newblock In: CVPR, pp. 2770--2779. IEEE, Long Beach California (2019)

\bibitem{Zhang_2018_CVPR}
Zhang, J., Zhang, T., Dai, Y., Harandi, M., Hartley, R.: Deep unsupervised
  saliency detection: A multiple noisy labeling perspective.
\newblock In: CVPR, pp. 1--10. IEEE, Salt Lake City,UT,USA (2018)

\bibitem{9156373}
Zhang, X., Xu, C., Tao, D.: Context aware graph convolution for skeleton-based
  action recognition.
\newblock In: 2020 IEEE/CVF Conference on Computer Vision and Pattern
  Recognition (CVPR), pp. 14321--14330 (2020).
\newblock \doi{10.1109/CVPR42600.2020.01434}

\bibitem{Zhang_2021_ICCV}
Zhang, Y., Li, X., Liu, C., Shuai, B., Zhu, Y., Brattoli, B., Chen, H., Marsic,
  I., Tighe, J.: Vidtr: Video transformer without convolutions.
\newblock In: Proceedings of the IEEE/CVF International Conference on Computer
  Vision (ICCV), pp. 13577--13587 (2021)

\bibitem{faster}
Zhu, L., Sevilla{-}Lara, L., Tran, D., Feiszli, M., Yang, Y., Wang, H.:
  {FASTER} recurrent networks for video classification.
\newblock CoRR \textbf{abs/1906.04226} (2019).
\newblock \urlprefix\url{http://arxiv.org/abs/1906.04226}

\bibitem{zhuadvancing}
Zhu, L., Wang, L., Raj, A., Gedeon, T., Chen, C.: Advancing video anomaly
  detection: A concise review and a new dataset.
\newblock In: The Thirty-eight Conference on Neural Information Processing
  Systems Datasets and Benchmarks Track (2024)

\bibitem{Background-Detection:CVPR-2014}
Zhu, W., Liang, S., Wei, Y., Sun, J.: Saliency optimization from robust
  background detection.
\newblock In: CVPR, pp. 2814--2821. IEEE, Columbus, OH, USA (2014).
\newblock \doi{10.1109/CVPR.2014.360}

\bibitem{Zoph_2018_CVPR}
Zoph, B., Vasudevan, V., Shlens, J., Le, Q.V.: Learning transferable
  architectures for scalable image recognition.
\newblock In: CVPR, pp. 1--14. IEEE, Salt Lake City, UT, USA (2018)

\end{thebibliography}
}

\end{document}